\documentclass[sort&compress]{elsarticle}
\usepackage[utf8]{inputenc}

\usepackage{graphicx}
\usepackage{subcaption}
\usepackage{hyperref}
\usepackage{gensymb}
\usepackage{amsfonts}

\usepackage{enumitem}

\usepackage{booktabs}
\usepackage{algorithm}
\usepackage{algpseudocode}

\begin{document}

\begin{frontmatter}

\title{Multiobjective Coverage Path Planning: Enabling Automated Inspection of Complex, Real-World Structures}
\author[bir,uio]{K.O.~Ellefsen\corref{cor1}}
\ead{koellefsen@gmail.com}
\author[isi]{H.A.~Lepikson}
\author[bir,dfki]{J.C.~Albiez}

\address[bir]{Brazilian Institute of Robotics, SENAI CIMATEC, Salvador, Bahia, Brazil}

\address[isi]{SENAI Institute for Innovation in Automation, SENAI CIMATEC, Salvador, Bahia, Brazil}
\address[dfki]{Robotics Innovation Center, DFKI GmbH, Bremen, Germany}
\address[uio]{Department of Informatics, University of Oslo, Norway}

\begin{abstract}

An important open problem in robotic planning is the autonomous generation of 3D inspection paths -- that is, planning the best path to move a robot along in order to inspect a target structure. We recently suggested a new method for planning paths allowing the inspection of complex 3D 
structures, given a triangular mesh model of the structure. The method differs from previous 
approaches in its emphasis on generating and considering also plans that result in imperfect 
coverage of the inspection target. In many practical tasks, one would accept imperfections 
in coverage if this results in a substantially more energy efficient inspection path. 
The key idea is using a multiobjective evolutionary algorithm to optimize the energy usage and coverage of inspection plans simultaneously -- and the result is a set of plans exploring the different ways to balance the two objectives.
We here test our method on a set of inspection targets with large variation in size and complexity, and compare its performance with two state-of-the-art methods for complete coverage path planning. The results strengthen our confidence in the ability of our method to generate good inspection plans for different types of targets. The method's advantage is most clearly seen for real-world inspection targets, since traditional complete coverage methods have no good way of generating plans for structures with hidden parts. Multiobjective evolution, by optimizing energy usage and coverage together ensures a good balance between the two -- both when 100\% coverage is feasible, and when large parts of the object are hidden.

\end{abstract}

\begin{keyword}
Coverage Path Planning \sep Multiobjective Evolutionary Algorithm \sep Robotic Inspection



\end{keyword}

\end{frontmatter}
\section{Introduction}

Recent years' advances in robotic autonomy have resulted in an increased demand for the 
ability to autonomously plan and schedule complex missions. This paper addresses one 
challenging mission type, which is that of inspecting 3D structures. Currently, such 
inspection is typically planned and performed manually, but recent work has addressed 
ways to perform inspection planning autonomously, in domains ranging from ship hull 
inspection with an AUV (autonomous underwater vehicle)~\cite{Hover2012} to inspection of 
bridges with a wheeled robot~\cite{Lim2014} and the inspection of buildings and fields with 
UAVs (unmanned aerial vehicles)~\cite{Bircher2015}.

In a recent paper~\cite{Ellefsen2016a}, we proposed an inspection path planning algorithm for an AUV designed for inspection of subsea oilfield infrastructure. The demands presented by this type of inspection mission highlight limitations to previous inspection path planners, in particular with regards to their ability to handle complex structures with occluded or hidden parts.

Previous work on planning inspection paths has typically made the assumption that the inspection 
mission has to achieve 100\% \emph{sensor coverage} of a given structure. In our application, as in many 
practical inspection planning problems, a 100\% covering plan is not necessarily desirable: If the 
energy difference between a well-covering plan and a perfectly covering plan is too large, 
we may prefer the former. In fact, 100\% coverage may not even be \emph{feasible} in situations 
where structures have occluded or hidden parts -- further emphasizing the need for 
\emph{partial coverage} inspection plans. This led us to suggest a radically different inspection path planner, which considers coverage of the inspection target an \emph{objective} rather than a \emph{constraint}. Initial results suggested that this \emph{multiobjective inspection path planner} could flexibly handle structures with hidden elements~\cite{Ellefsen2016a}.

In this paper, we expand on our preliminary results~\cite{Ellefsen2016a} by comparing the multiobjective inspection path planner to state-of-the-art methods for coverage path planning. Since these methods originally assume that a 100\% coverage is desired, we implemented \emph{generalized} versions, where we relax the constraint on complete coverage. This allows us to compare the methods on structures with hidden parts, and to get an idea of the diversity in the plans each method can produce.

We compare the methods on structures with very different levels of complexity, to better understand the strengths and weaknesses of each one. The results from this comparison strengthen our confidence in the flexibility and robustness of our multiobjective inspection path planning algorithm. It generates high-quality plans across a large variety of shapes and coverage levels,  and consistently produces plans with better or similar energy-efficiency compared to competing approaches.

\section{Application Target}

\begin{figure}
 \centering
 \includegraphics[width=0.5\textwidth]{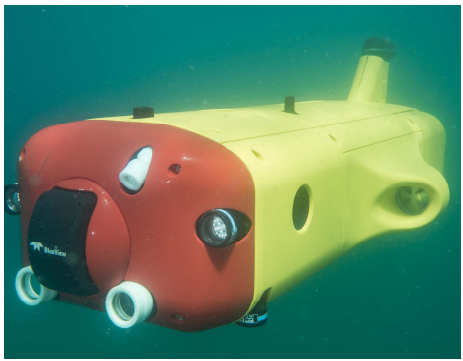} 
 \caption{\textbf{The FlatFish AUV is the target platform for the inspection plans.} The figure shows it during ocean trials.}
 \label{fig:flatfish}
\end{figure}

The work presented here is done in co-operation with the project FlatFish, which aims to develop a subsea-resident AUV for inspection of offshore infrastructure. The FlatFish AUV, currently being developed in a cooperation between the Brazilian Institute of Robotics, Shell and DFKI, Germany, is shown in Figure~\ref{fig:flatfish}. We refer the reader to~\cite{albiez2015} for details on the AUV, and~\cite{7761203} for information about the algorithm used to gather data and reconstruct inspection targets.

Underwater structures like oil and gas production systems or the foundations of buildings,
piers and off-shore wind facilities, have to be regularly inspected to evaluate the state
of the structure and to plan future interventions for repair and maintenance. Being able to 
inspect the subsea part of an asset regularly and/or on demand plays a key 
role for subsea asset integrity assurance (SIA) and integrity management. The key 
information gathered from inspections encompasses the general structural integrity and 
close visual data (also known as close visual inspection, CVI). These
inspections are currently performed by remotely operated vehicles (ROVs) or, if water depth
and availability allow it, by divers. Due to the fact that inspections of this kind require a
special support vessel, they are time-consuming, expensive, need to be planned a long time
in advance and rely on good weather conditions and seasonal constraints (e.g. not during
hurricane season or winter storms).

Current research in subsea robotics focuses strongly on using autonomous underwater 
vehicles (AUVs) for inspections~\cite{McLeod2011,German2012,Soltwedel2013}. The idea of this approach is to have a subsea resident AUV with a 
docking station in the offshore asset and let the robot perform 
inspections on the infrastructure whenever it is needed. Besides the obvious technical 
challenges like robust navigation or subsea docking, creating an efficient plan for structure 
inspection is very important. 

The inspection plan must cover all interesting parts of 
the structure with the AUVs sensors and must be as energy efficient as possible since the 
AUV runs on internal battery power. The 3D model of all the inspection targets within the 
field is known, either from the time of deployment or from previous inspection runs. The 
goal for the inspection planner therefore is to find the optimal trajectory around a known 
target.

\section{Background}
\label{sec:background}

\subsection{Coverage Path Planning}

Planning a path that inspects a three-dimensional object is an example of a coverage path planning (CPP) problem: The task of finding a path covering an area of interest, while avoiding any obstacles~\cite{Galceran2013}. In general, the CPP problem is NP-Hard, meaning an optimal solution (in terms of path length or energy usage) is unrealistic for anything but toy problems. Therefore, most CPP solvers seek a good, but not optimal, solution. 

CPP methods can be classified as \emph{continuous} or \emph{discrete}, depending on the way the robot covers structures~\cite{Englot2012a}. The former assumes \emph{continuous sensing or deposition} -- implying, for inspection missions, that information is gathered as the robot moves along the path. Discrete CPP, on the other hand, makes the assumption that information is gathered at \emph{specific locations}, for instance in the case of a robot that needs to stabilize for a moment in a position before gathering data.

Most previous work in coverage path planning (see~\cite{Galceran2013} for a recent review) considers how to make a \emph{robot's body} cover a 2-dimensional area (as needed in for instance vacuum cleaning and harvesting robots). On the other hand, robotic inspection missions like the one considered in this paper require us to ``cover'' a 3-dimensional object with the robot's \emph{field of view}. Advanced methods for 3D coverage path planning have received attention only in recent years, and the most relevant work in this direction is described below. 


\subsubsection{Complete coverage path planning for 3D targets}
\label{sec:ccpp_background}
Common to all previous methods on planning 3D inspection paths is to focus on the \emph{complete coverage} problem, meaning they aim to generate the best path covering 100\% of the inspection target. While time constraints are common in robotic planning problems~\cite{Khaluf2013}, these complete coverage solvers make the assumption that we would like to cover the entire target structure -- with no strict upper time limit.

Most previous methods~\cite{Trucco1997,Chen2004,Englot2011, Englot2012, EnglotHover2012, Hover2012, Hollinger2012a, Cashmore2013, Cashmore2014, Bircher2015} reduce the problem's complexity by making the simplifying assumption that we are dealing with \emph{discrete inspection}: Inspection where data is gathered when the robot is standing still -- on \emph{vertices} of the path plan, rather than \emph{edges}~\cite{Englot2012a}. This allows them to divide the planning problem in two: 1) planning the location of viewpoints and 2) planning the order of visiting them. How these two steps are achieved vary greatly between the approaches, but common to them all is that they first find a set of viewpoints that together give a complete coverage of the inspection target, and thereafter find a tour that visits these viewpoints efficiently.

Methods for discrete coverage path planning have demonstrated good performance, even handling obstacles and occluding elements gracefully~\cite{Englot2012}. However, their simplifying assumption that we can optimize viewpoints and the paths between them separately may be a problem whenever there are \emph{dependencies} between the two goals of viewpoint planning and path 
planning~\cite{Papadopoulos2013}. This is the case, for instance, in a cluttered environment where some viewpoints are not reachable from all other viewpoints, and when dealing with \emph{continuous inspection}. To handle such dependencies, planners need to optimize viewpoints and paths \emph{together}. Methods for \emph{continuous} coverage path planning have been suggested using a genetic algorithm~\cite{Lim2014}, sampling-based motion planning~\cite{Papadopoulos2013} and sampling-based path planning~\cite{Englot2010}. These methods generate plans tailored to robots that perform inspections while moving.


An alternative to time-consuming optimization techniques is to generate a continuous coverage 
path rapidly through geometrical calculations~\cite{Bibuli2007}. Coverage plans for 3D structures can be generated by ``slicing'' the structure at evenly spaced depth levels, and circling around each slice at a fixed distance~\cite{Atkar2001, Cheng2008, Bibuli2007}. While such circling-based plans may not be optimal for certain structures (for instance, this method cannot optimize the robot's trajectory so that it can view occluded objects), this method has the advantages of being simple and having a very low computational complexity.

\subsubsection{Multiobjective coverage path planning}
\label{sec:background_multiobjective}

Several authors have suggested ways to use \emph{multiobjective} optimization in path planning -- most frequently to balance energy usage and risk~\cite{Mittal2007, Vadakkepat2000, Zhang2013a}. However, to our knowledge multiobjective optimization for \emph{planning inspection paths} has not been explored until our recently published paper~\cite{Ellefsen2016a}.

\begin{figure}
\centering
\includegraphics[width=\textwidth]{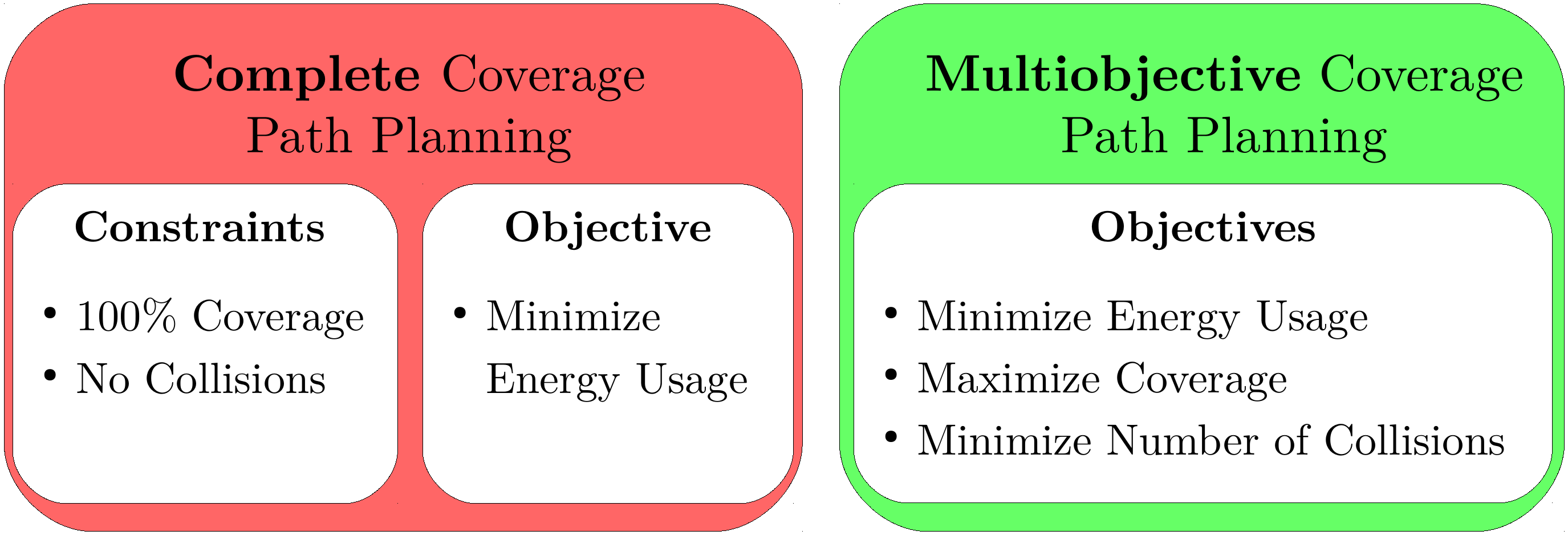}
\caption{\textbf{The key idea of our inspection path planning approach.} Unlike previous methods, we consider structure coverage as an objective instead of a constraint. The method also allows us to turn other typical constraints, such as avoiding collisions, into objectives -- allowing plans with collisions and/or low coverage to serve as ``stepping stones'' on the way to optimized, collision-free plans.}
\label{fig:concept}
\end{figure}

The traditional approach to planning inspection paths for 3D objects assumes the \emph{constraint} of requiring a complete coverage, in addition to other constraints related to the robot's potential movements, such as not allowing plans coming too close to obstacles. Our \emph{multiobjective coverage path planning} method instead treats maximizing inspection coverage and minimizing the number of collisions as additional factors to optimize (Figure~\ref{fig:concept}) -- in contrast to previous methods which typically only optimize plans' energy usage.

The key implications of this reformulation of the inspection planning problem are 1) That inspection plans can be optimized also for complex structures where 100\% coverage is \emph{not possible} and 2) That a large family of optimal inspection plans can be generated, ranging from short inspections of the most important structural features to long inspection missions covering most or all of the structure surface. Section~\ref{sec:multiobjective_inspection} contains more details about how our multiobjective inspection path planning is implemented.

\subsection{Multiobjective Evolutionary Optimization}
This section describes the methods we apply from the field of multiobjective evolutionary optimization (MOEA) in order to generate inspection path plans, and to compare the performance of different configurations of the optimizer.

\subsubsection{Applied evolutionary algorithms}
\label{sec:nsga_moead}
Multiobjective Evolutionary Optimization is a rapidly developing field, with a large selection of candidate algorithms (see~\cite{Deb2015} for a recent review). Our main focus here is on the algorithm  NSGA-II (Non-dominated Sorting Genetic Algorithm, version II)~\cite{Deb2002a}, since it is known to perform well on problems with few (2-3) objectives~\cite{Ishibuchi2015}, and since it is one of the most popular MOEAs -- and therefore available in many Evolutionary Algorithm frameworks. As an additional control, to increase the confidence in our results, we also carried out the main experiment with the more recent MOEA/D algorithm~\cite{Zhang2007}. MOEA/D belongs to a \emph{different class} of MOEA than NSGA-II, and this control therefore helps ascertain whether or not the results from NSGA-II can generalize to other MOEAs. Below follow brief descriptions of the two applied algorithms.

NSGA-II uses the the concept of \emph{Pareto dominance} when comparing and selecting solutions. Solution $A$ dominates solution $B$ if the following two requirements are fulfilled: 1) $A$ is not outperformed by $B$ on any objective, and 2) $A$ performs better than $B$ with regards to \emph{some} objective. NSGA-II also maintains a high level of \emph{diversity} in the population of solutions, by giving an increased preference to solutions that have objective values far from those of the remaining population. Together, the search for non-dominated and diversified solutions may produce a good approximation of the true Pareto front, which represents the set of solutions with \emph{optimal trade-offs} between the different objectives. In the case of inspection path planning, the Pareto front contains those solutions that strike the optimal balance between energy costs and inspection coverage -- those that ``get the most coverage for their spent energy''. See~\cite{Deb2002a} for a more detailed and formal description of NSGA-II.

MOEA/D also attempts to find an approximation of the Pareto front, but has a very different way of doing so: Instead of relying on Pareto dominance, MOEA/D \emph{decomposes} the multiobjective optimization problem into several different single-objective problems. Throughout evolution, the population is composed of the best solutions so far to each of these subproblems. Since similar subproblems will often have similar solutions, recombination mainly takes place between \emph{neighboring solutions}, that is, solutions solving similar subproblems. The neighborhood size is therefore an important parameter in this algorithm, affecting its ability to properly explore the search landscape. If the decomposition of the multiobjective problem has resulted in an even distribution of subproblems, evolution can explore many different ways to balance the objectives simultaneously, resulting in a good approximation of the Pareto front. See~\cite{Zhang2007} for more details on MOEA/D.

Common for both NSGA-II and MOEA/D is that we do not have to formalize our \emph{preferences} between the different objectives. Rather, we select the solution that best matches our preferences \emph{a posteriori}, after the final solutions have been generated~\cite{Knowles2005}. The user of the inspection planner can thus select the plan best matching the requirements of the current mission from a population of optimized candidate plans.


\subsubsection{Measuring optimization performance}

\begin{figure}
\centering
\includegraphics[width=0.8\textwidth]{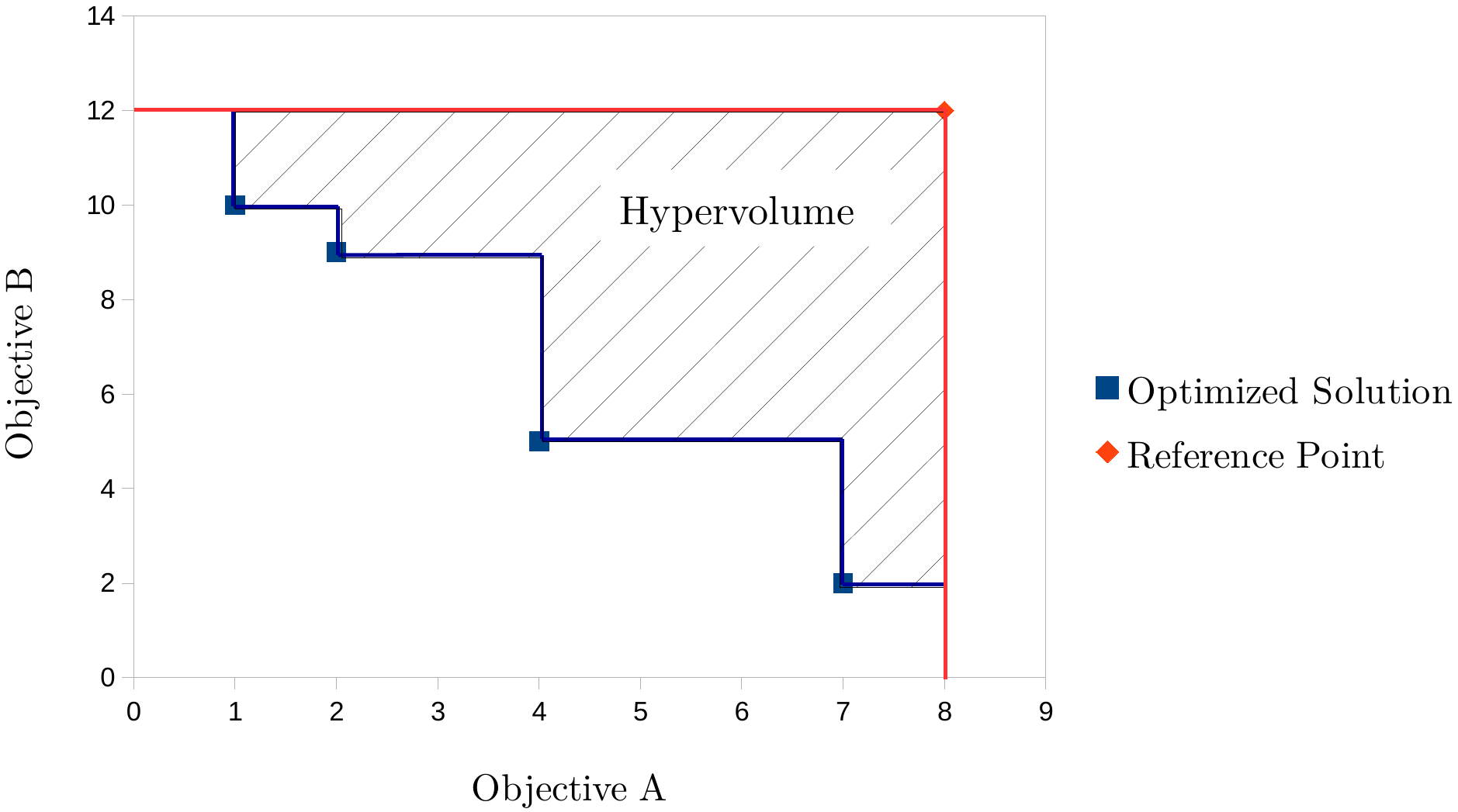}
\caption{\textbf{Calculating the hypervolume spanned by a set of optimized solutions.} This performance indicator measures the quality of a population of solutions resulting from multiobjective optimization as the hypervolume they span relative to a reference point. In the special case of having exactly two objectives, the hypervolume is an area.}
\label{fig:hypervolume}
\end{figure}

Since the objectives we wish to optimize, energy usage and coverage degree, are in conflict, we cannot measure the performance of optimization by looking at either in isolation: Optimizing coverage is easy if we have infinite energy resources, and optimizing energy is easy if we do not have to cover anything. Instead, we rely on the hypervolume indicator~\cite{Zitzler1998, Auger2009}, a popular measure of the performance of multiobjective evolutionary algorithms. It gives a population of solutions a value reflecting how closely this population matches the true Pareto front, by calculating the size of the space spanned by this population (Figure~\ref{fig:hypervolume}). The hypervolume is measured relative to a \emph{reference point}, which is by many authors recommended to be selected by taking the worst possible value for each objective, and increasing it slightly~\cite{Auger2009} -- a technique we have adopted in our measurements. We used the hypervolume indicator to measure the optimization progress over generations of evolution, as well as for comparing different experimental treatments.

Another technique we apply to compare different experimental treatments is to study their \emph{attainment surfaces} and difference in \emph{empirical attainment functions}~\cite{Fonseca1996, GrunertDaFonseca2001}. Attainment surfaces summarize the performance of several runs of a multiobjective optimization algorithm, by indicating which parts of the objective space were reached (or \emph{attained}) by that algorithm. Here, we plot \emph{best}, \emph{worst} and \emph{median} attainment surfaces. Together, the three indicate the entire area of the objective space reached by \emph{any} run of the algorithm, and its median performance. Finally, measuring \emph{difference} in attainment functions facilitate the comparison between different algorithms, or different settings of an algorithm, by indicating which areas of the objective space were attained by one algorithm, but not by another. We refer the reader to~\cite{Lopez-Ibanez2010b} for more details on how to measure and visualize attainment functions.

\section{Methods}
\label{sec:methods}
The main goal of this paper is validating our previously published multiobjective coverage path planning algorithm~\cite{Ellefsen2016a} by comparing it to other state-of-the-art methods. Since none of the previous methods for 3D coverage path planning are fully suitable for this comparison (they all generate \emph{complete coverage} plans), we implemented \emph{generalized} versions of two state-of-the-art inspection planning techniques. The generalizations are needed to enable the planners to handle the generation of plans with \emph{imperfect coverage} -- allowing a more relevant comparison with our method.

The two methods we compare our algorithm with are circularly sweeping plans, which is part of several different algorithms for continuous coverage path planning for 3D objects~\cite{Bibuli2007,Atkar2001, Cheng2008}, and sampling-based coverage path planning, which was explored in a series of papers by Englot and Hover, both for continuous~\cite{Englot2010} and discrete~\cite{Englot2011, Englot2012, EnglotHover2012} 3D coverage path planning. We consider these the two most relevant methods for this comparison, since they have both been suggested in recent papers on inspection path planning, they are able to generate continuous inspection plans, and it is possible to generalize them to generate plans without complete coverage.

The remainder of this chapter describes the three techniques and how we compared them. Since the details of the algorithms are already published elsewhere, we focus on the modifications we made to enable comparison of the methods -- in particular those enabling the previous methods to generate plans with \emph{incomplete coverage}.

\subsection{Comparing Inspection Path Planners}
\begin{figure}
\centering
\includegraphics[width=\textwidth]{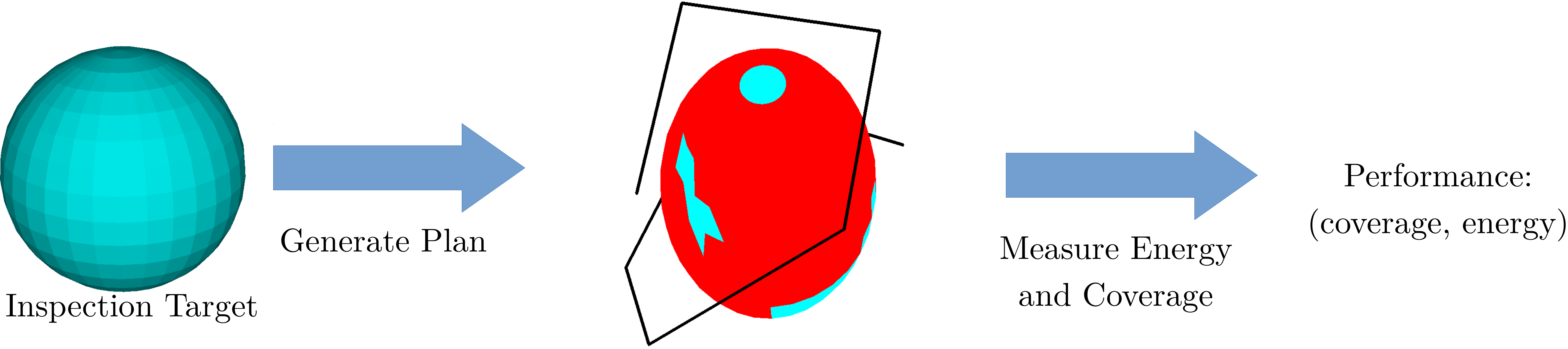}
\caption{\textbf{How we test the planners.} Each planner receives an inspection target, and generates an inspection plan for it. Energy and coverage of each plan is estimated, and the plan is plotted as shown in the middle. The black line is the plan path, and the inspection target is colored to show which parts were successfully inspected (red/dark) and not (cyan/light).}
\label{fig:main_structure}
\end{figure}

Figure~\ref{fig:main_structure} shows the procedure for comparing the three inspection path planners. They all accept 3D mesh models as inputs, and generate a path consisting of a sequence of waypoints and orientations. Orientations describe the heading of the robot when moving between consecutive waypoints. Each generated plan is tested for its  energy usage (Section~\ref{sec:energy_estimation}) and coverage of the inspection target (Section~\ref{sec:coverage_estimation}), before plans resulting from the three different methods are compared.

\subsubsection{Energy usage estimation}
\label{sec:energy_estimation}

The estimation of energy usage is necessary in two parts of this work: 1) when comparing the results of different planning techniques, and 2) when assigning fitness values to individuals in the evolutionary optimization of inspection plans. The energy usage of a plan is estimated by the following formula:

\begin{equation}
 \sum\limits_{\vec{e} \in Edges(plan)} (w_{trans}\cdot ||\vec{e}||) + 
 (w_{rot}\cdot(1-cos(\theta_{\vec{e}-1,\vec{e}})))
\end{equation}

where $w_{trans}$ and $w_{rot}$ are constants regulating how much we emphasize 
translations and rotations in the energy calculation, and $\theta_{\vec{e}-1,\vec{e}}$ is the 
angle between edge $i-1$ and edge $i$ (the first edge has $\theta=0$). 
For each edge in a plan, the energy usage is thus calculated as the sum of the translation 
energy spent along that edge, and the rotational energy required to move in the direction of 
the edge. The latter is estimated as being proportional to the \emph{difference in direction} 
between the current and previous edge. The weights applied in our experiment were $w_{trans} = 0.1$ and $w_{rot} = 1$, emphasizing the cost of direction changes in the underwater environment.

This energy estimate was applied due to its simplicity in implementation and low complexity in computation, facilitating the generation of plans and comparison of algorithms. In practice, when applying the plan optimization methods discussed here in real-world trials, it will probably be better to pay the extra cost of computing more accurate energy estimates, to get a plan as closely tailored as possible to a specific AUV's performance. Energy usage estimation is an external method in all the optimization techniques discussed herein, making it straightforward to exchange the calculation we apply with any other function taking an inspection path as input and returning a number representing an energy estimate. Obtaining a more exact energy estimate calibrated to a specific AUV could be done by following the method proposed in~\cite{DeCarolis2014}.

The sampling-based planner and the circling-based planner never generate plans where edges collide with the inspection target (collision here defined as any part of the edge being closer to the inspection target than a safety buffer of 1.5 $m$). The EA-based inspection planner can, however generate plans with collisions: We chose to \emph{penalize} rather than \emph{discard} evolved plans with collisions because good plans with a few collisions may serve as stepping-stones for the search towards good plans without collisions.

Any edges that come too close to the inspection target are penalized by increasing their energy usage. It is important that this penalty is large enough to avoid rewarding plans for taking shortcuts \emph{through} the inspection target. We chose to set the penalty to be equal to \emph{twice the length} of the longest side of the inspection target. This value reflects the fact that the collision could be avoided by post-processing the plan to traverse around the sides of the inspection target rather than pass through it. Any inspection information gathered while traversing a colliding edge is disregarded. In practice, we see this penalty leading to the final, evolved solutions avoiding collisions -- and for the ``seeded'' evolutionary runs (Section~\ref{sec:seeding}) we \emph{never} observed an optimized plan intersecting the inspection target. 

\subsubsection{Coverage estimation}
\label{sec:coverage_estimation}
Coverage estimation takes as input an inspection path or a single edge, a 3D model of the structure to be inspected and a list of each of the robot's cameras (including their relevant parameters, such as heading relative to the robot, opening angles and range) that will be used to inspect the structure, and estimates which geometric primitives will be observed by the path or edge.

Coverage estimation is used both \emph{to compare} the planning methods and \emph{internally} in the methods during plan generation. In the sampling-based planning method and the evolutionary plan optimization, it is necessary to estimate the set of primitives observed by each considered edge. This is done by simulating camera ``snapshots'' at several locations along the edge, with the robot orientation defined for that edge, and maintaining a set of all observed primitives.

When estimating the coverage of an entire plan, the algorithm iterates through \emph{each edge} in the plan, estimating all geometric primitives visible along the edge as described above. After iterating over all edges in the plan, we have a set estimating all geometric primitives seen while carrying out the plan. The total area of these primitives is calculated and compared with the area of all primitives in the 3D model of the inspection target with the following formula:

\begin{equation}
coverage\_score = 1.0-(covered\_area/total\_area)
\label{eq:coverage}
\end{equation}

$coverage\_score$ thus ranges from 0 (complete coverage) to 1 (no coverage). We chose 0 as the optimum, to frame coverage optimization as a minimization-problem, in line with the optimization of energy usage. Further details of how we calculate the coverage score (including pseudocode) are found in~\cite{Ellefsen2016a}.

\subsubsection{Common assumptions and parameters}
\label{sec:assumptions}

\begin{figure}
 \centering
 \includegraphics[width=0.5\textwidth]{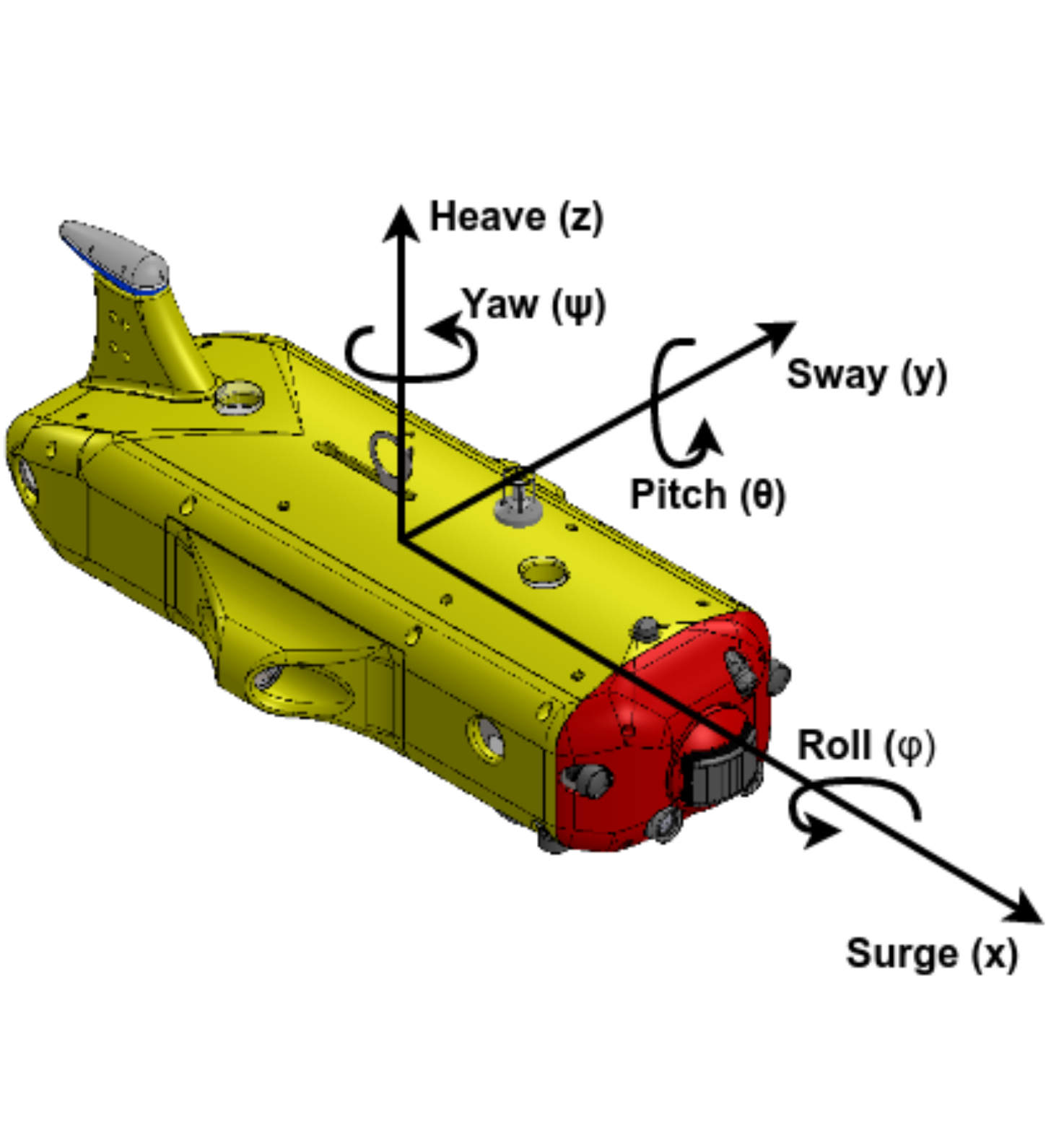} 
 \caption{\textbf{Degrees of freedom of the FlatFish AUV.} The AUV actively controls five of the six degrees of freedom, and roll is passively stabilized. Image from~\cite{7761251}.}
 \label{fig:flatfish_degrees_of_freedom}
\end{figure}

Assumptions used to evaluate coverage and energy usage are adapted to our vehicle and mission type -- especially with regards to the FlatFish AUV's camera setup and degrees of freedom~\cite{albiez2015}. The FlatFish AUV operates in five of the six Euclidean degrees of freedom actively (Figure~\ref{fig:flatfish_degrees_of_freedom}). The buoyancy  and weight distribution on FlatFish is designed to have an optimal metacentric height, which passively stabilizes the AUV on the roll axis. Pitch is controlled by the two diving thrusters, and controlled to be zero all the time when moving slowly to minimize thruster effort. The active degrees of freedom used by FlatFish during inspection are therefore heave, sway, surge and yaw.

For all the compared methods, it is assumed that the robot inspects \emph{while moving} (that is, on edges in the generated plan), and that the robot's orientation 1) always lies in the x-y plane (no rotation in pitch or roll) and 2) is perpendicular to the movement direction in the x-y plane (this results in the largest covered volume, since our robot's horizontally facing camera is in front). In the case of a move along the z-axis, the robot is assumed to point towards the center of the inspection target.

It is further assumed that the robot has \emph{two cameras} to utilize in inspection, one pointing straight forward, and the other straight down. The two cameras gather information simultaneously. The parameters of the simulated cameras are shown in Table~\ref{table:cam_parameters}.

A final assumption, which simplifies plan evaluation, is that all non-occluded structures in the field of view of the AUV are equally visible. We make this assumption since 1) The stand-off distance to the structure is always too small to cause any significant influence of attenuation of our on-board lights and lasers, and 2) The focus here is on off-line planning -- unpredictable changes in currents, visibility and accessibility will occur, but adaptation to these will need to be handled on-line. 



\begin{table}
\centering
\begin{tabular}{lr}
 \hline
 Parameter & Value \\
 \hline
 Vertical and Horizontal Pixels & 1024 \\
 Near Plane Distance & $0.1m$ \\
 Far Plane Distance & $10m$ \\
 Vertical Field of View & 46\degree \\
 Horizontal Field of View & 46\degree \\
 \hline
\end{tabular}
\caption{\textbf{Parameters of the simulated cameras.}}
\label{table:cam_parameters}
\end{table}


\subsection{Discretization}
\label{sec:discretizing viewpoints}
The MOEA-planner and the circling-paths planner both rely on the same discretization of the space around inspection targets into \emph{candidate waypoints}. In both cases, a sequence of these candidate waypoints constitutes the inspection plan.

\begin{figure}
\centering
\includegraphics[width=0.5\textwidth]{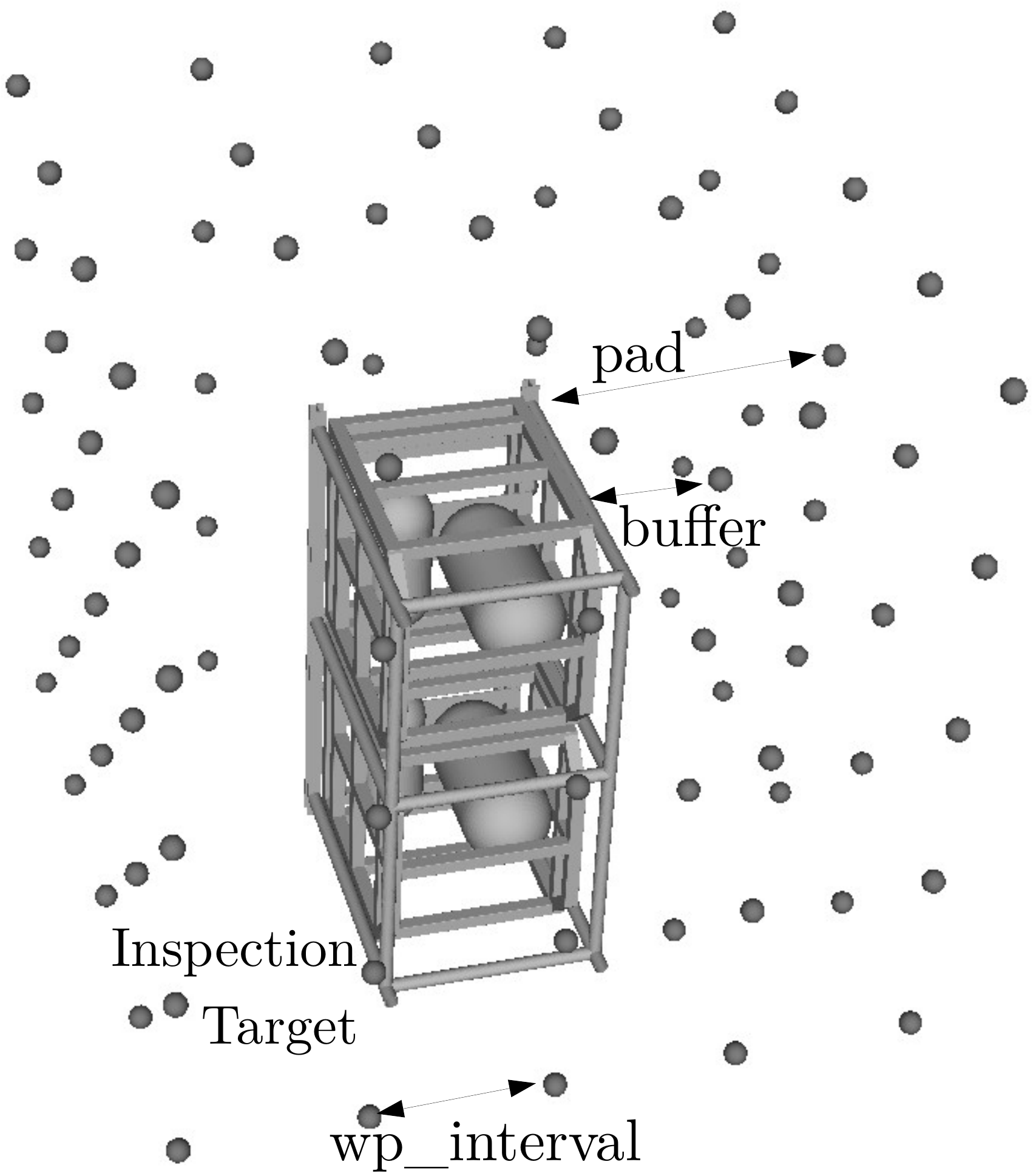}
\caption{\textbf{Generating candidate waypoints for a structure.} The figure illustrates the effect of the parameters $pad$, $buffer$ and $wp\_interval$. Note that, to avoid a cluttered illustration, this example has a far coarser discretization than that which was used during the experiments documented here.}
\label{fig:discretizing_into_waypoints}
\end{figure}

The generation of candidate waypoints takes three parameters: $pad$, which defines how much ``padding'' to add around the structure's bounding box (and thus, the maximum distance waypoints can have from the structure), $buffer$, which defines the minimum distance between the structure and candidate waypoints, and $wp\_interval$, which defines the spacing between candidate waypoints. The effect of these parameters is illustrated in Figure~\ref{fig:discretizing_into_waypoints}. Note that we do not pad the structure below it's minimum z-value, since our inspection targets are mainly floor-mounted structures. This constraint of never generating waypoints \emph{below a structure} was imposed on all three compared methods. 

The spacing between waypoints, was calculated as a function of the \emph{volume of the structure's bounding box} as follows:

\begin{equation}
	wp\_interval = \sqrt[3]{padded\_bb\_vol/volume\_scaling}
\end{equation}

where $padded\_bb\_vol$ is the volume of the bounding box around the structure plus padding (in other words, the total volume inside which we will add candidate waypoints), and $volume\_scaling$ is a constant factor that controls the proportion of the $padded\_bb\_vol$ that each candidate waypoint occupies. This calculation of $wp\_interval$ maintains the complexity of the search relatively independent of the structure volume. Table~\ref{table:structure_details} shows that it generates a similar number of candidate waypoints for our three structures, despite their great differences in volume.

The values we chose for the parameters $pad$, $buffer$ and $volume\_scaling$ are given in Table~\ref{table:discretization_parameters}. For any given application, these values need to be informed by the search complexity one is willing to accept, and the parameters of the inspection robot.

\begin{table}
\centering
\begin{tabular}{l r}
 \hline
Parameter & Value\\
 \hline
 $pad$   & $4.0m$\\
 $buffer$   & $2.0m$\\
 $volume\_scaling$   & $1000$\\
 \hline
\end{tabular}
\caption{\textbf{Parameters of the discretization into candidate waypoints.}}
\label{table:discretization_parameters}
\end{table}

\subsection{Multiobjective Inspection Path Optimization}
\label{sec:multiobjective_inspection}
We recently proposed a way to use multiobjective evolutionary computation to generate inspection path plans exhibiting good balance between energy and coverage~\cite{Ellefsen2016a}. Algorithm~\ref{alg:moea_overview} shows a high-level overview of the method.

\begin{algorithm}
\caption{\textproc{$MultiobjectiveInspectionPlanning$}($structure$, $seeds$)}
\label{alg:moea_overview}
\begin{algorithmic}[1]
  \State $wp\_candidates \gets CandidateWaypoints(structure)$
  \State $population \gets \emptyset$
  \While{$population < pop\_size$}
  	\If{$Rand() < p_{seeded}$}
  		\State $population \gets population + DrawElems(seeds,1)$
	\Else
  		\State $n \gets RandInt(min\_init\_size, max\_init\_size)$
  		\State $rand\_ind \gets DrawElems(wp\_candidates,n)$
		\State $population \gets population + rand\_ind$
	\EndIf
  \EndWhile
  \State $FitnessEvaluate(population)$
  \For{$gen$ $\in[0 : num\_generations]$}
  	\State $selected \gets TournamentSelect(population, pop\_size)$
  	\State $offspring \gets CrossoverAndMutate(selected, p_{crossover}, p_{mutation})$
  	\State $FitnessEvaluate(offspring)$
  	\State $population \gets NSGASelect(population+offspring, pop\_size)$
  \EndFor
  \State \Return $population$
\end{algorithmic}
\end{algorithm}

This algorithm applies the widely used multiobjective optimizer NSGA-II (Non-dominated Sorting Genetic Algorithm, version II)~\cite{Deb2002a} in order to evolve an initial population of plans into increasingly well performing inspection path plans. It was implemented using the Python-framework DEAP~\cite{Gagn2012} for distributed evolutionary algorithms. The values of key parameters in our experiments are given in Table~\ref{table:ea_params}. The table also indicates the maximum number of fitness evaluation, including evaluating initial solutions and all the generations of evolution. Note, however, that much fewer evaluations will be used in practice, since individuals that are not changed through mutation or crossover will not need to be reevaluated. Since our main goal is to compare multiobjective to single-objective inspection path planning, we did not carry out extensive tuning or sensitivity analysis when settling on these parameters. However, we encourage further studies to investigate how multiobjective inspection path plans vary with these and other parameters.

\begin{table}
\centering
\begin{tabular}{l r}
 \hline
Parameter & Value\\
 \hline
 $pop\_size$   & $40$\\
 $num\_generations$   & $400$\\
 $p_{mutation}$  & $0.1$\\
 $p_{crossover}$   & $0.1$\\
 $p_{seeded}$ Seeded Runs & 0.35\\
 $p_{seeded}$ Unseeded Runs & 0.0\\
 Max. Num. of Fitness Evaluations & $16,040$\\
 \hline
\end{tabular}
\caption{\textbf{Parameters of the evolutionary algorithm.}}
\label{table:ea_params}
\end{table}

The optimization starts by generating the set of candidate waypoints as described in Section~\ref{sec:discretizing viewpoints}, and assigning each waypoint a \emph{unique ID}. An inspection plan is represented as \emph{sequence of viewpoint IDs} of any length. By modifying this sequence through crossover and mutation, the evolutionary optimization selects which waypoints to include in an inspection plan and in which order they will be visited. Since this algorithm is already described in~\cite{Ellefsen2016a}, we will not re-iterate the details. However, we find it appropriate to discuss the way we \emph{initialize} plans, since we will explore the effects of two different forms of initialization in Section~\ref{sec:results}.

\subsubsection{Initializing the optimization}
\label{sec:seeding}

Individuals in the initial population of path plans are generated through random sampling, \emph{or} from a previously generated set of \emph{seed plans}, which are plans that are simple to generate, but still contain some information on the beneficial features of an inspection path. Initializing evolutionary algorithms with \emph{seed solutions} has previously demonstrated the potential to improve performance~\cite{Hernandez-Diaz2008, Ellefsen2011}. As seeds for the inspection path planning, we used the circling plans generated as described in Section~\ref{sec:circling_sweeps}, since these are good plans that are simple to generate. For each structure, we created a \emph{pool of seeds}, consisting of all circling plans for that structure.

A common problem in evolutionary algorithms is progress halting or slowing down due to a 
lack of \emph{diversity} in the population of candidate solutions~\cite{Mouret2012}. Therefore, to ensure a proper level of diversity in the initial population, we always generated a significant amount of its individuals \emph{randomly}. Initial solutions were sampled (with replacement) from the pool of seeds with probability $p_{seeded}$, and otherwise generated randomly. To investigate the effect of seeding on the evolutionary optimization, we regulated this probability as shown in Table~\ref{table:ea_params} -- performing 1) runs where all initial individuals are random (unseeded runs), and 2) runs where a significant proportion (on average 35\%) of initial individuals are seeded.



\subsubsection{MOEA/D}
\label{sec:moead_parameters}
The main comparison between multiobjective and traditional inspection path planners applied the NSGA-II algorithm, as described above. However, to increase our confidence that the findings are general to multiobjective optimization algorithms, and not due to specific features of the NSGA-II algorithm, we also generated inspection plans with the MOEA/D algorithm, which belongs to a different class of MOEA than NSGA-II.

For MOEA/D, we applied the same parameters as for NSGA-II when possible, and otherwise relied on default parameters from the applied MOEA/D implementation (from the Platypus framework for evolutionary computation\footnote{\url{http://platypus.readthedocs.io}}). These default values are presented in Table~\ref{table:moead_params}. Note that since a generation of NSGA-II and a generation of MOEA/D may involve very different amounts of computation, we specified the target number of evaluations here, rather than the number of generations, for a more accurate comparison. The target number of evaluations was set to the median number of evaluations from the initial NSGA-II runs on each inspection target. 20 independent runs of each algorithm (MOEA/D and NSGA-II) were performed, each run terminating in the same generation that the target number of evaluations was reached.

For the MOEA/D specific parameters, $\delta$ is the probability of selecting individuals for mating only from an individual's \emph{neighborhood}. Correspondingly, with probability (1-$\delta$), \emph{all individuals} are considered for the current round of mating. \emph{Scalarizing function}  indicates the function used to decompose the multiobjective optimization problem into N different single-objective problems. See~\cite{Zhang2007} for more details about the MOEA/D algorithm and its parameters.

\begin{table}
\centering

\begin{tabular}{l r}
 \hline
Parameter & Value\\
 \hline
 Target number & 3073 (sphere)\\
 of evaluations& 2966 (SSIV)\\
 & 3194 (manifold)\\
Neighborhood size & 10\\
 $\delta$ & 0.8\\
Scalarizing function & Tchebycheff\\
 \hline
 \end{tabular}
 
\caption{\textbf{The parameters for MOEA/D.} The parameters not listed here applied the same values as NSGA-II (Table~\ref{table:ea_params}). See main text for explanation of the parameters.}

\label{table:moead_params}
\end{table}

\subsection{Generalized Circling Sweeps}
\label{sec:circling_sweeps}

Ideas related to covering a 3D object by following several circling trajectories at a fixed distance from its sides are present in several algorithms~\cite{Bibuli2007,Atkar2001, Cheng2008}. While the details differ, the common idea is moving around the target structure at a fixed distance at a single depth level, before moving to the next depth (the distance between which depends on the field of view of the vehicle's sensors), making a new sweep around the edges of the target, and so on. By carefully choosing the spacing between such depth-levels, complete coverage of the sides of a 3D structure can be guaranteed, as long as no obstacles or occluding elements are present. 

To enable comparison with our method, we here generalize this technique to explore different balances between energy usage and coverage. The generalization consists in, instead of carefully selecting depth-levels, testing many different depth-deltas ($\Delta z$), measuring the coverage and energy usage of each resulting plan. In the extreme case of a maximum $\Delta z$, only a single circling around the target is performed, whereas on the other end of the spectrum, we find plans circling the target enough to cover its sides completely, not considering occluding elements.

\subsubsection{Method details}

To facilitate development and comparisons, the circling sweeps were, like the sampling-based plans and the multiobjective plans, generated as a sequence of waypoints. This also facilitated the use of such plans as seeds for evolved solutions (Section~\ref{sec:seeding}).

\begin{algorithm}
\caption{\textproc{$CircleEachLevel$}($structure,wp\_candidates$)}
\label{alg:gen_circles}
\begin{algorithmic}[1]
  \State $all\_level\_circles \gets \emptyset$
  \State $z\_levels \gets UniqueZLevels(wp\_candidates)$
  \For{$z \in z\_levels$}
  	\State $edge\_wps \gets GetNeighborWaypoints(structure,wp\_candidates)$
  	\State $traced\_edge \gets MooreNeighborTracing(edge\_wps)$
  	\State $pruned\_edge \gets PruneEdge(traced\_edge)$
  	\State $all\_level\_circles \gets all\_level\_circles + pruned\_edge$
  \EndFor
  
  \State \Return $all\_level\_circles$
\end{algorithmic}
\end{algorithm}

The algorithm has two parts: 1) Generating paths sweeping around each candidate depth level of the inspection target, and 2) Generating several \emph{complete} plans by combining different sweeps from part 1. The first part (Algorithm~\ref{alg:gen_circles}) is initiated with the relevant 3D model and the set of potential waypoints, generated as described in Section~\ref{sec:discretizing viewpoints}. Since this set of waypoints is generated by sampling in regular intervals along the x-, y- and z-axes, we can easily identify a small set of unique z-levels ($z\_vals$ in Algorithm~\ref{alg:gen_circles}). For each level, a plan is generated, tracing the edges of the structure at that level. This is done by first identifying all candidate waypoints at that level within a given distance from the inspection target ($GetNeighborWaypoints$), before tracing around the outside of this set of neighbors using the Moore Neighbor Tracing algorithm. Finally, the traced edge is simplified by removing all waypoints directly between other consecutive waypoints ($PruneEdge$) -- thus reducing the number of waypoints while keeping the plan identical. The resulting sequence of waypoints ($pruned\_edge$) performs a single loop around the structure.

When Algorithm~\ref{alg:gen_circles} terminates, one circling sub-plan has been generated for each depth level. The second part of the method is generating a population of complete edge tracing plans, circling around the target different numbers of times. This is done by iterating over a variable, $\Delta z$, from 1 to the total number of z-levels, and ensuring consecutive loops are $\Delta z$ levels apart in each final plan. In addition, the complete plans are centered around the target along the z-axis, ensuring the circling plans cover the structure in an energy-efficient way. An example of the effect of varying $\Delta z$ can be seen in Figure~\ref{fig:manifold-circling}, which presents circling plans for $\Delta z$ equal to 1 (left), 2 (middle) and 7 (right).


\subsection{Generalized Sampling-based Coverage Path}
\label{sec:sampling}
Englot and Hover published a series of papers centered around the idea of generating inspection plans by sampling in \emph{robot configuration space}~\cite{Englot2010, Englot2011, EnglotHover2012, Englot2012}. The dimensionality of this space varies depending on the number of degrees of freedom of our robot. Since we in our comparison have standardized the way of assigning robot orientations (Section~\ref{sec:assumptions}), a configuration is simply a position in 3D space. The paper on sampling-based inspection path planning of most relevance in this comparison is~\cite{Englot2010}, which is the only one that focuses on \emph{continuous inspection}, that is, inspections where data is gathered while the robot is moving.


Part 1 of the algorithm generates a dense graph of nodes (representing robot configurations) and edges (representing straight paths between waypoints), by sampling robot configurations pseudo-randomly and adding edges between them and their nearest neighbor configurations, until \emph{each geometric primitive} is observed by the robot along some edge. In part 2, a \emph{minimum-cost closed walk} along the graph which covers 100\% of the geometric primitives is approximated, by use of integer programming.

The main change in modifying this method to allow sub-100\% covering inspection plans was to allow the stopping criterion of the sampling in part 1 to be a user-defined fraction, $f$, instead of 100\% coverage, and to relax the constraint of complete coverage on the integer programming problem formulated in part 2.

\subsubsection{Graph generation}

Algorithm~\ref{alg:sampling_based_coverage} describes part 1 of the generalized sampling-based algorithm, which notably differs from Algorithm 1 in~\cite{Englot2010} in taking an argument $f$, which is the fraction of the structure the plan needs to cover. Identically to the original algorithm, the generalized version also takes as input a starting configuration ($q_0$), and an inspection target ($structure$).

\begin{algorithm}
\caption{\textproc{$BuildInspectionGraph$}($q_0,structure,f$)}
\begin{algorithmic}[1]
  \State $g.init(q_0)$
  \State $observed\_primitives \gets \emptyset$
  \While{$observed\_primitives.size()/structure.size() < f$}
  	\If{$observed\_primitives.size()/structure.size() < (f-\epsilon)$}
  		\State $AddToGraph(g)$
  	\Else
  		\State $AddMissingView(g)$
  	\EndIf
  \EndWhile
\end{algorithmic}
\label{alg:sampling_based_coverage}
\end{algorithm}

Algorithm~\ref{alg:sampling_based_coverage} builds a graph of candidate waypoints and edges between them, until at least a fraction $f$ of the total area of the inspection target can be observed by traversing the complete set of edges. Initially, waypoints are added by pseudo-random sampling, and edges by connecting sampled waypoints to their closest neighbors ($AddToGraph$ -- see details in ~\cite{Englot2010}). However, when only a small fraction $\epsilon$ of the structure remains to reach the coverage goal, a more focused search ($AddMissingView$) begins for the best robot configurations to cover these remaining primitives.

While its goal is the same, our implementation of $AddMissingView$ differs from that in~\cite{Englot2010} in one important aspect. The original $AddMissingView$ selects the first primitive that has not yet been observed by any edge in $g$, calculates a robot configuration that permits the robot's sensor to observe this primitive and builds an edge that includes this configuration. Then, the process is repeated for the next unobserved primitive.

In our case, generating an edge covering each missing primitive is not possible, since many primitives in our models are \emph{not visible} from anywhere around the structure. Therefore, we implemented a different way to gather the final $\epsilon$ fraction of primitives: We select a \emph{random} unobserved primitive, and sample a \emph{random} valid edge near it. If the edge gathers any new observation, and can be connected to the rest of the graph, it is added to the graph. Whether or not a new edge was added, a \emph{different} random unobserved primitive is chosen, and the process repeated. This ensures the process does not get stuck trying to find a viewpoint for a primitive that is impossible to observe due to occluding elements.

\subsubsection{Path-finding}

When Algorithm~\ref{alg:sampling_based_coverage} terminates, we have a graph with the necessary edges to cover a fraction $f$ of the inspection target's geometric primitives. The second part of the method finds the optimal way to traverse this graph, by formulating an integer programming problem with the requirement that \emph{any primitive covered by any of the sampled edges ($ObservedPrimitives$ in Algorithm~\ref{alg:sampling_based_coverage}) must be covered by the final plan}. This is a relaxation of the integer programming problem in~\cite{Englot2010}, which had the requirement that \emph{all primitives on the structure must be covered by the final plan}. Since the integer programming problem is otherwise identical to that formulated in~\cite{Englot2010}, it will not be repeated here.





\subsection{Test Models}
\label{sec:test_models}

\begin{figure}
    \centering 
     \begin{subfigure}[b]{0.24\textwidth}
        \includegraphics[width=\textwidth]{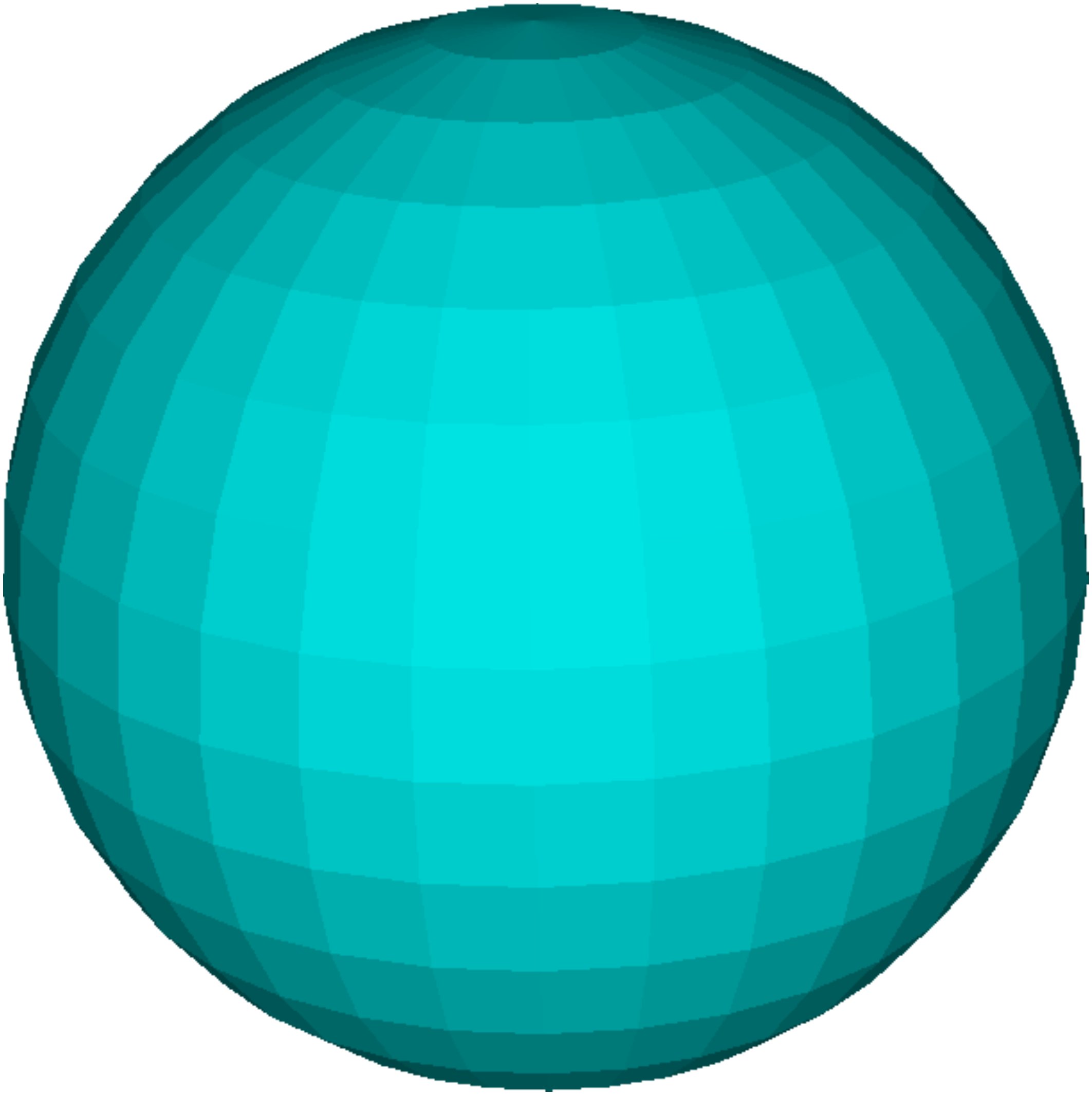}
        \caption{Sphere}
        \label{fig:sphere}
    \end{subfigure}
        ~ 
    \begin{subfigure}[b]{0.3\textwidth}
        \includegraphics[width=\textwidth]{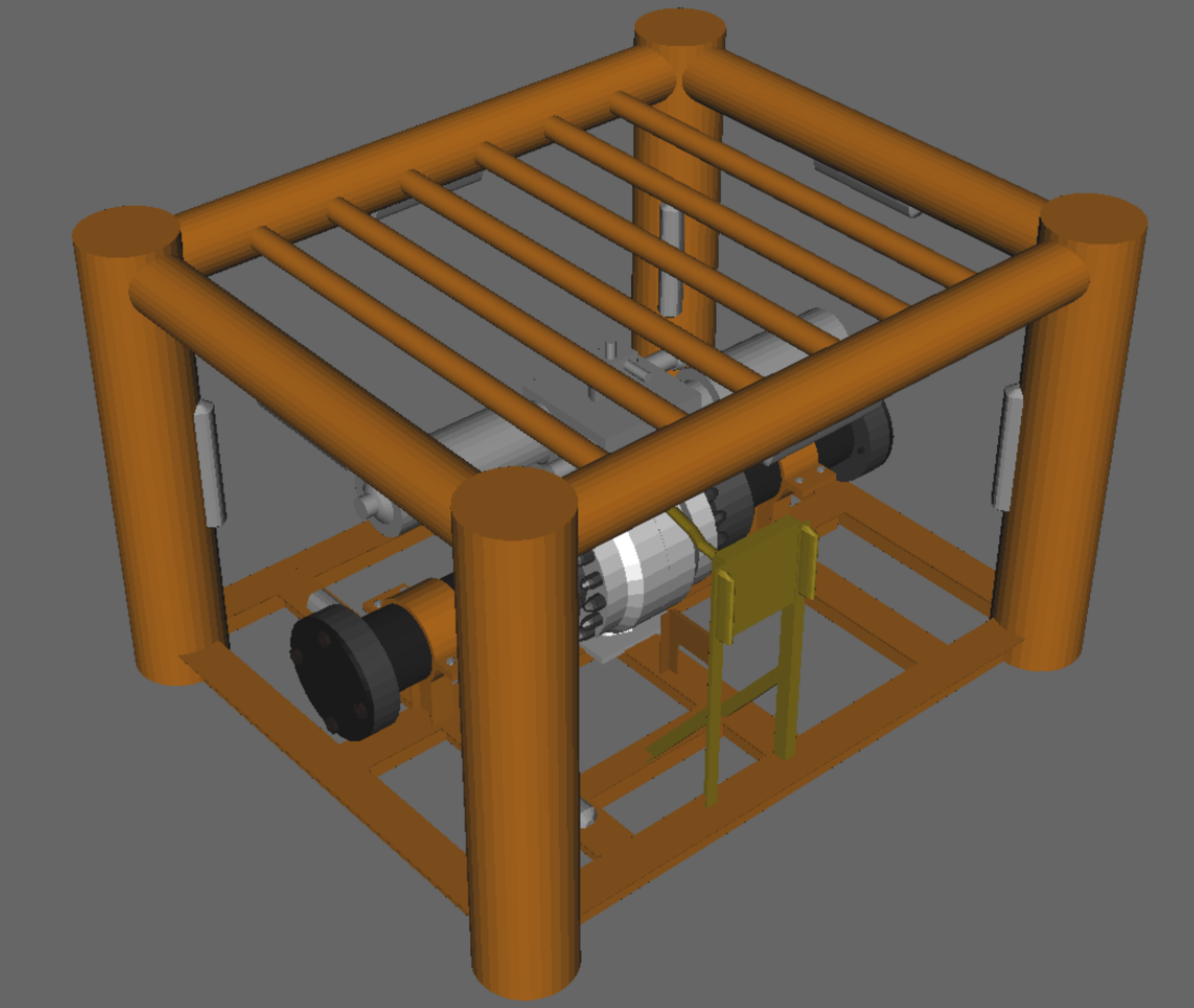}
        \caption{Subsea isolation valve}
        \label{fig:oil_pump}
    \end{subfigure}
        ~ 
    \begin{subfigure}[b]{0.4\textwidth}
        \includegraphics[width=\textwidth]{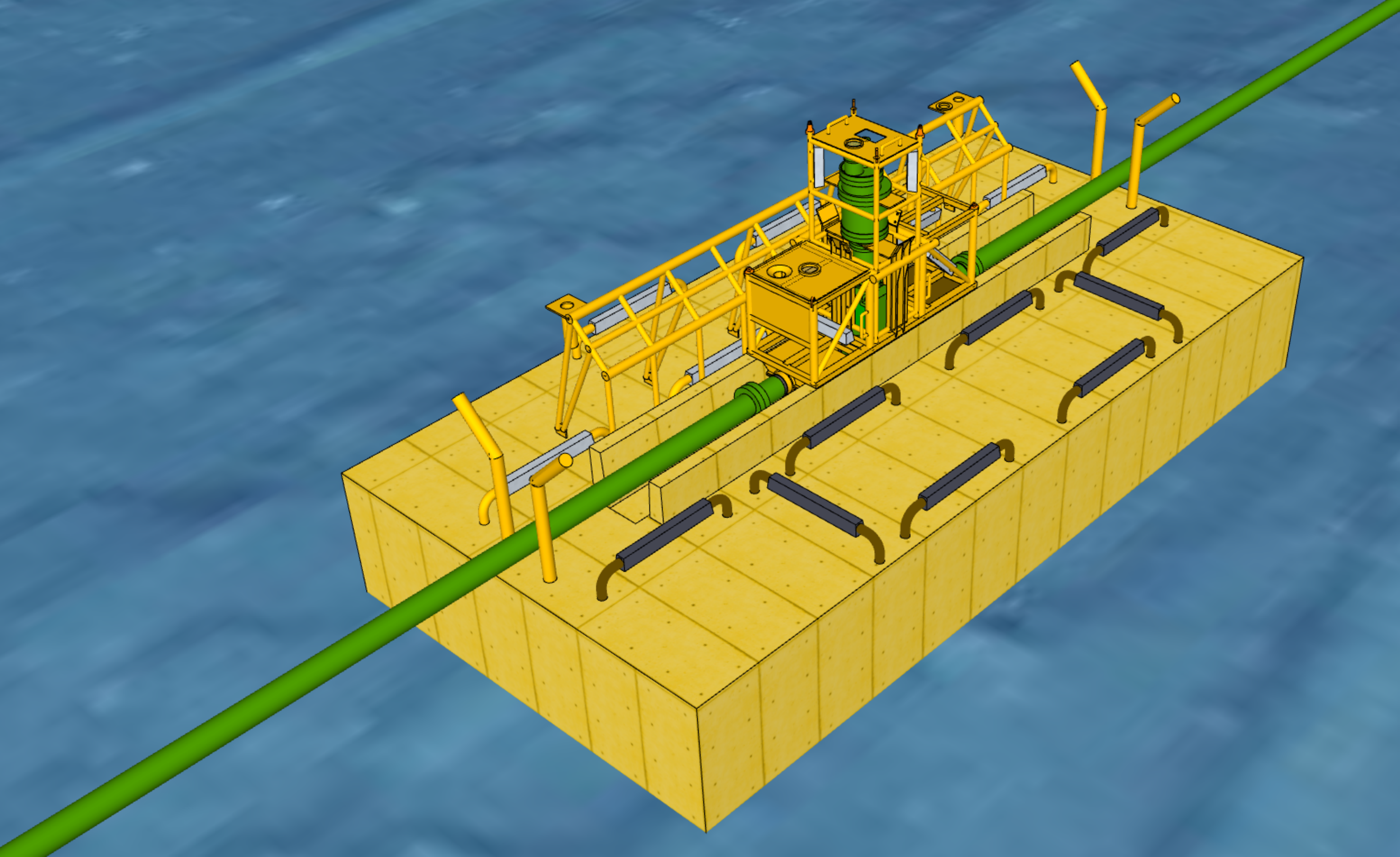}
        \caption{Large subsea manifold}
        \label{fig:manifold}
    \end{subfigure}
    \caption{\textbf{The inspection targets.} These were chosen since they represent three different levels of complexity. Complete coverage inspection is only possible for the sphere.}\label{fig:targets}
\end{figure}

Our multiobjective inspection planner was tested against the previous techniques on three different 3D models: 1) A simple sphere, 2) A relatively small subsea isolation valve (SSIV) and 3) A large, complicated subsea manifold (Figure~\ref{fig:targets}). These were chosen to test the algorithms on problems with a variety of sizes and complexity levels: The sphere allows 100\% coverage, and has previously been used to test complete coverage planning~\cite{Englot2010}, whereas complete coverage is impossible for the two other structures. Table~\ref{table:structure_details} shows further details about the 3D models.

\begin{table}
\centering
\begin{tabular}{l c c c r}
 \hline
Structure & Triangles & Dimensions & Candidate Waypoints & $f\_max$\\
 \hline
 Sphere   & 960 & Radius $10m$ & 1492 & 1.0 \\
 SSIV   & 26707 & $4m\times3.2m\times2.6m$ & 1604 & 0.78\\
 Manifold   & 88153 & $117m\times17m\times11m$ & 1489 & 0.84\\
 \hline
\end{tabular}
\caption{\textbf{Details about the inspection targets.} All are 3D mesh models, made up of a large number of triangles. Their sizes vary greatly, but a similar number of candidate waypoints are generated for all by automatically scaling the distance between neighbor waypoints. $f\_max$ is the maximum $f$-value given to the generalized sampling-based planner for each structure (Section~\ref{sec:comparison}).}
\label{table:structure_details}
\end{table}
    
\subsection{Experimental Setup}
\label{sec:comparison}
To have a robust basis for comparison of the methods, we performed 20 independent runs of the EA-planner and the sampling-based planner. The evolutionary algorithm is inherently random, both due to its random initial population and random mutation and crossover events. By initializing different runs with different seeds to the random number generator, we therefore easily achieved independently evolved populations.

The sampling-based planner, however, samples pseudo-randomly using a Sobol sequence, which produces the same sequence of robot configurations each time it is run on the same structure~\cite{Englot2010}. The method also relies on a starting configuration ($q_0$ in Algorithm~\ref{alg:sampling_based_coverage}). To mitigate effects of a choosing a particularly good or bad starting configuration, we performed 20 runs of this algorithm with different random starting positions (all required to be \emph{outside} the bounding box of the target, but no further than $10m$ away). The circling plan generator has no random elements, and was therefore run only once.

Due to the generalizations mentioned earlier in this chapter, a single run of each algorithm generates \emph{a population} of results with different balances between coverage and energy usage. In the case of the EA, this population is simply its \emph{evolutionary} population. In the case of the circling-based planner, the population consists of all the different plans with different values of $\Delta z$ (Section~\ref{sec:circling_sweeps}).

For the sampling-based planner, a varied population of results was generated by running the algorithm with 11 different values of the \emph{coverage threshold} $f$ (Algorithm~\ref{alg:sampling_based_coverage}), evenly distributed in the interval [$f\_min$, $f\_max$]. In other words, we generated 10 ``reduced coverage plans'' ($f < f\_max$) for every ``maximum coverage plan'' ($f = f\_max$). This was repeated 20 times, resulting in a total of 220 sampling-based plans generated for each inspection target. The coverage threshold represents the \emph{fraction of the inspection target's area} that we require the generated plan to cover, and regulating it naturally generates a population of plans with variation in coverage degree and energy usage. For all the structures, we chose a value of $f\_min=0.1$, meaning the ``least strict'' requirement sent to the sampling-based planner was to cover 10\% of each structure. 

Since we do not know the theoretical maximum coverage for our two complex structures (for the sphere, the maximum is 100\%), we found an upper limit for the coverage threshold by gradually increasing $f$, and observing where the sampling-based planner begins having trouble. Giving a too large $f$-value to the sampling-based planner will lead to a prohibitively large inspection graph $g$ (Algorithm~\ref{alg:sampling_based_coverage}) - resulting in the integer programming problem defined on the graph becoming too complex. The inspection paths generated for the largest $f$-values (e.g. the leftmost plan in Figure~\ref{fig:manifold-sampling}) indicate that the values we selected for $f\_max$ are near the maximal value the planner can handle.

\section{Results and Discussion}
\label{sec:results}

When comparing the performance of different single-objective optimization methods, one can rely on traditional statistical measurements on the single objective. Comparing multiobjective optimization methods is more complicated, but for multiobjective evolutionary optimization, methods such as comparing hypervolumes~\cite{Zitzler1998, Auger2009} or attainment surfaces~\cite{Fonseca1996, GrunertDaFonseca2001} are emerging as common techniques. 

\begin{figure}
\centering

    \begin{subfigure}[b]{0.55\textwidth}
\includegraphics[width=\textwidth]{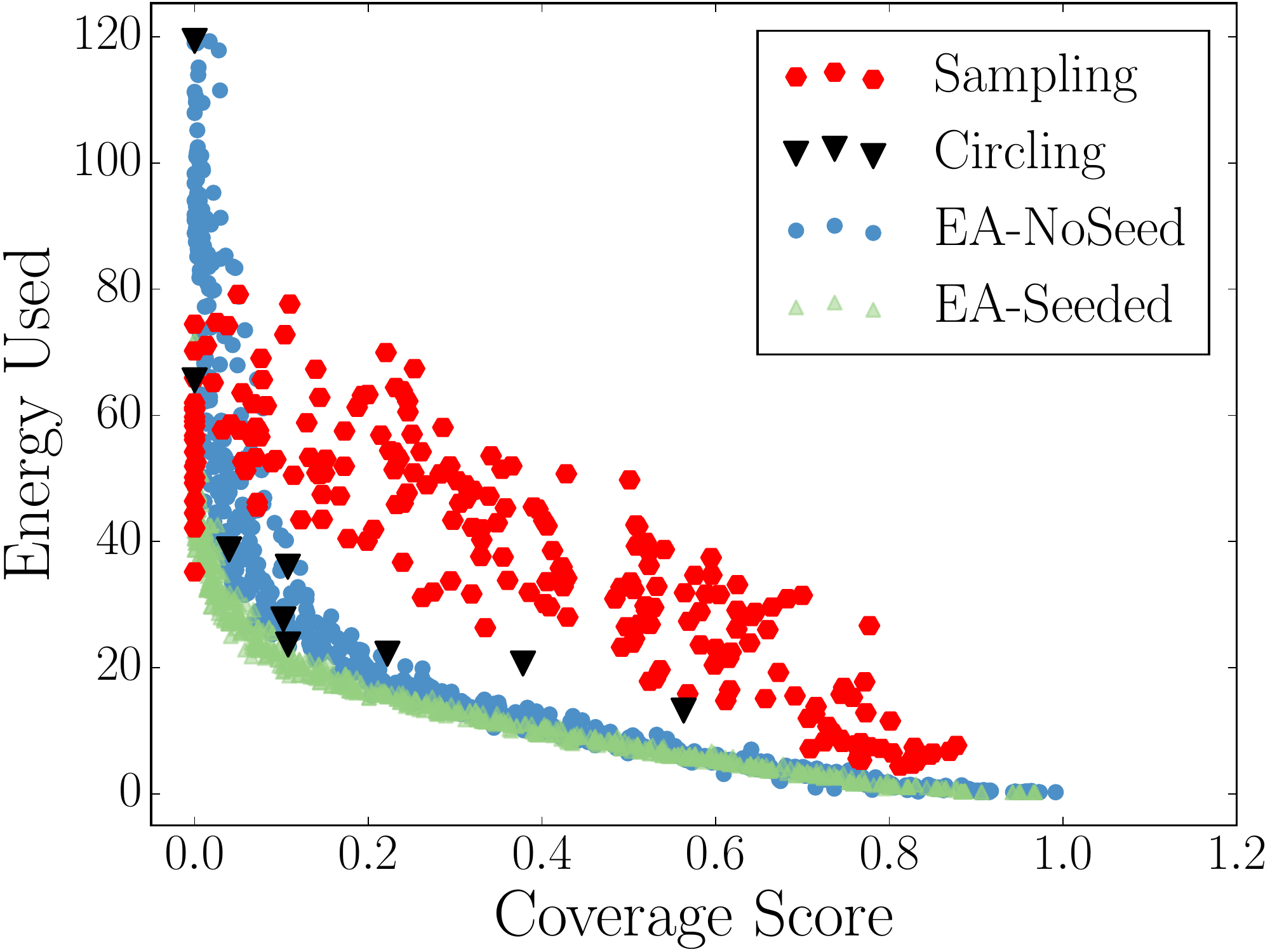}
\caption{Inspection target: Sphere}
\label{fig:scores_sphere}
\end{subfigure}

    \begin{subfigure}[b]{0.55\textwidth}
\includegraphics[width=\textwidth]{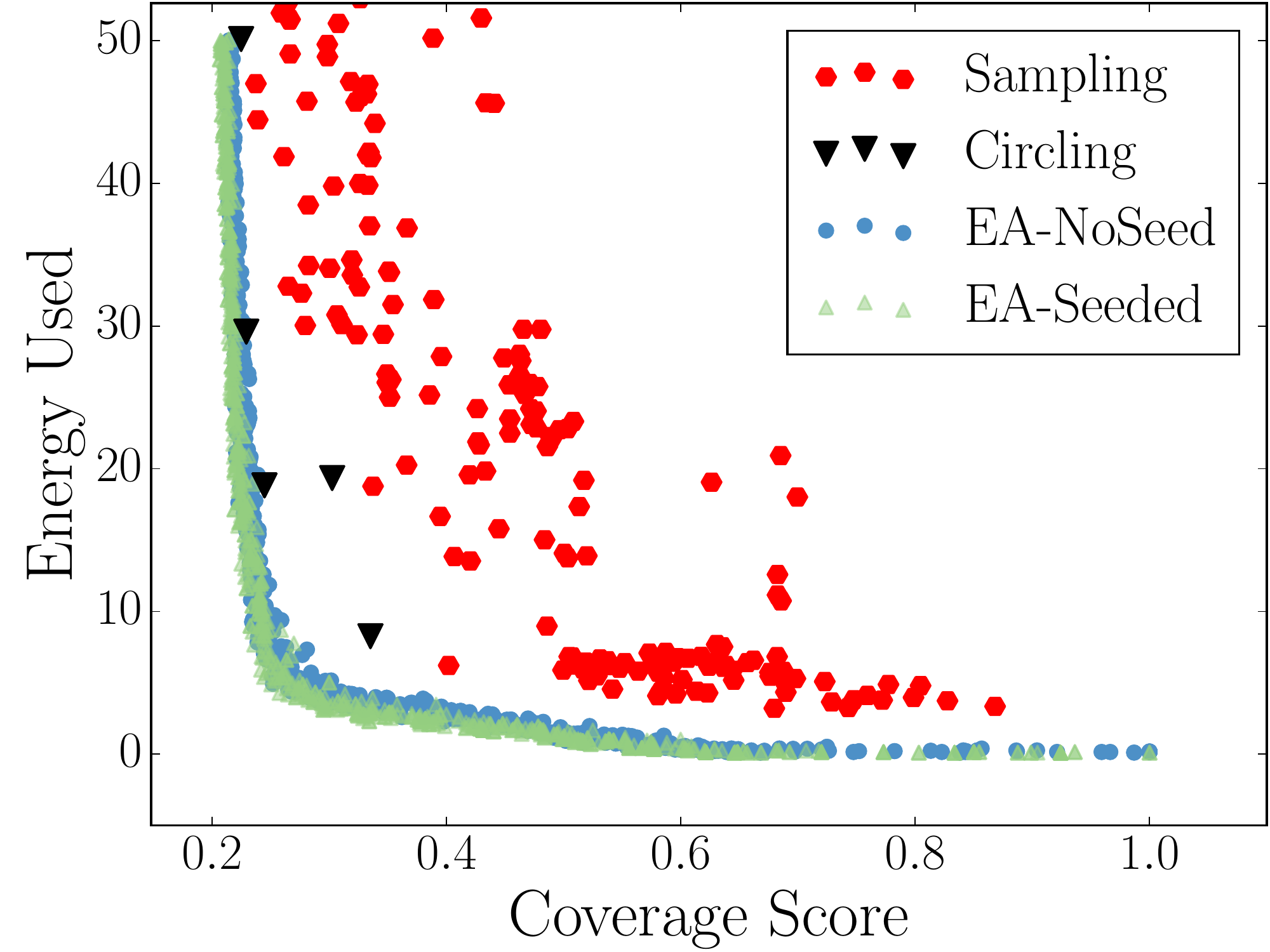}
\caption{Inspection target: SSIV}
\label{fig:scores_pump}
\end{subfigure}

    \begin{subfigure}[b]{0.55\textwidth}
\includegraphics[width=\textwidth]{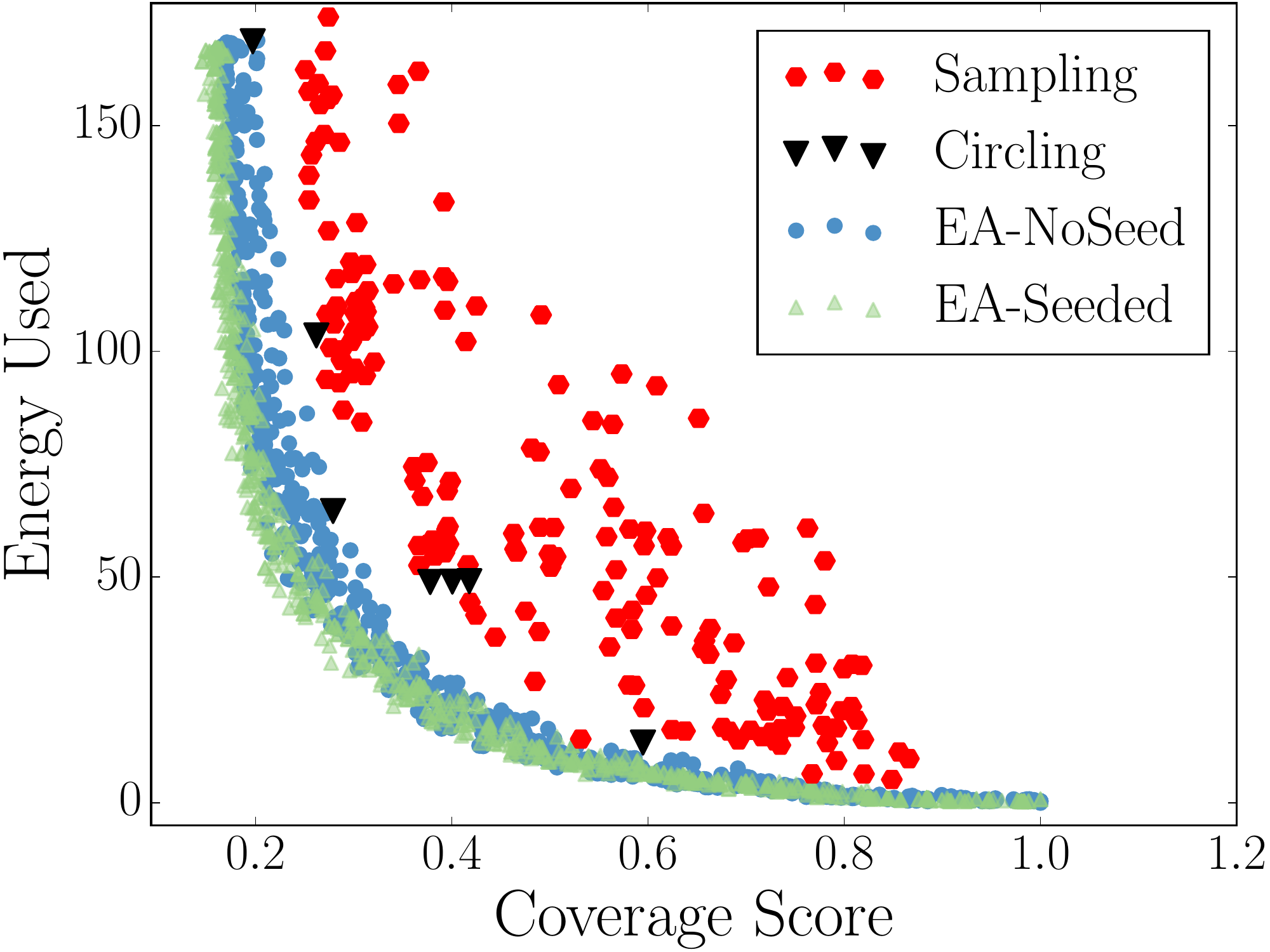}
\caption{Inspection target: Large manifold}
\label{fig:scores_manifold}
\end{subfigure}
\caption{\textbf{Scores of all inspection plans resulting from evolutionary optimization (NSGA-II), sampling and circling on the three inspection targets.} We remind the reader that the optimum for both objectives is 0. To better compare the results, Figures b and c are cropped to exclude the unreasonably long plans sometimes generated by sampling-based planning. Complete versions can be seen in the Supplementary Material.}
\label{fig:all_pareto}
\end{figure}

\begin{figure}
    \centering
    \begin{subfigure}[b]{\textwidth}
        \includegraphics[width=\textwidth]{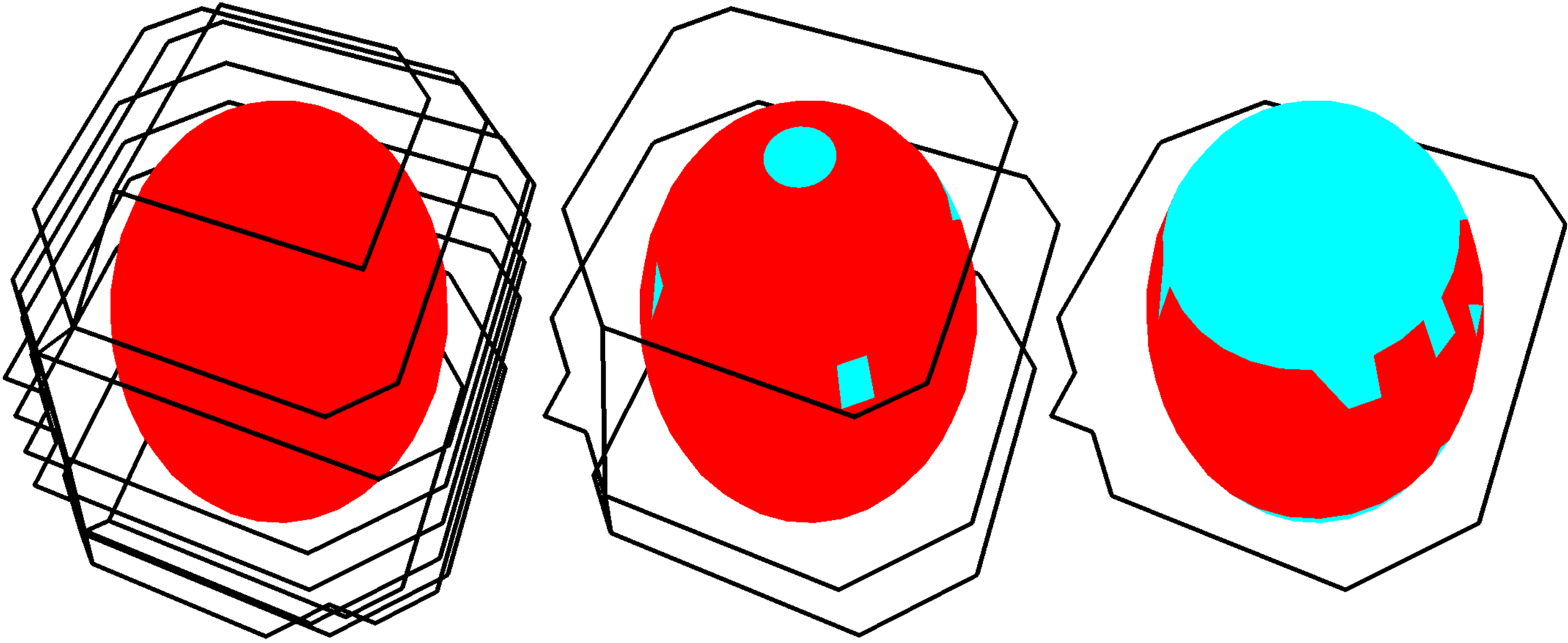}
        \caption{Circling plans -- scores (0.0, 119.3), (0.04, 38.7) and (0.56, 13.2), plans resulting from $\Delta z$ = 1, 3 and 9, respectively.}
        \label{fig:sphere-circling}
    \end{subfigure}
    ~ 
    \begin{subfigure}[b]{\textwidth}
        \includegraphics[width=\textwidth]{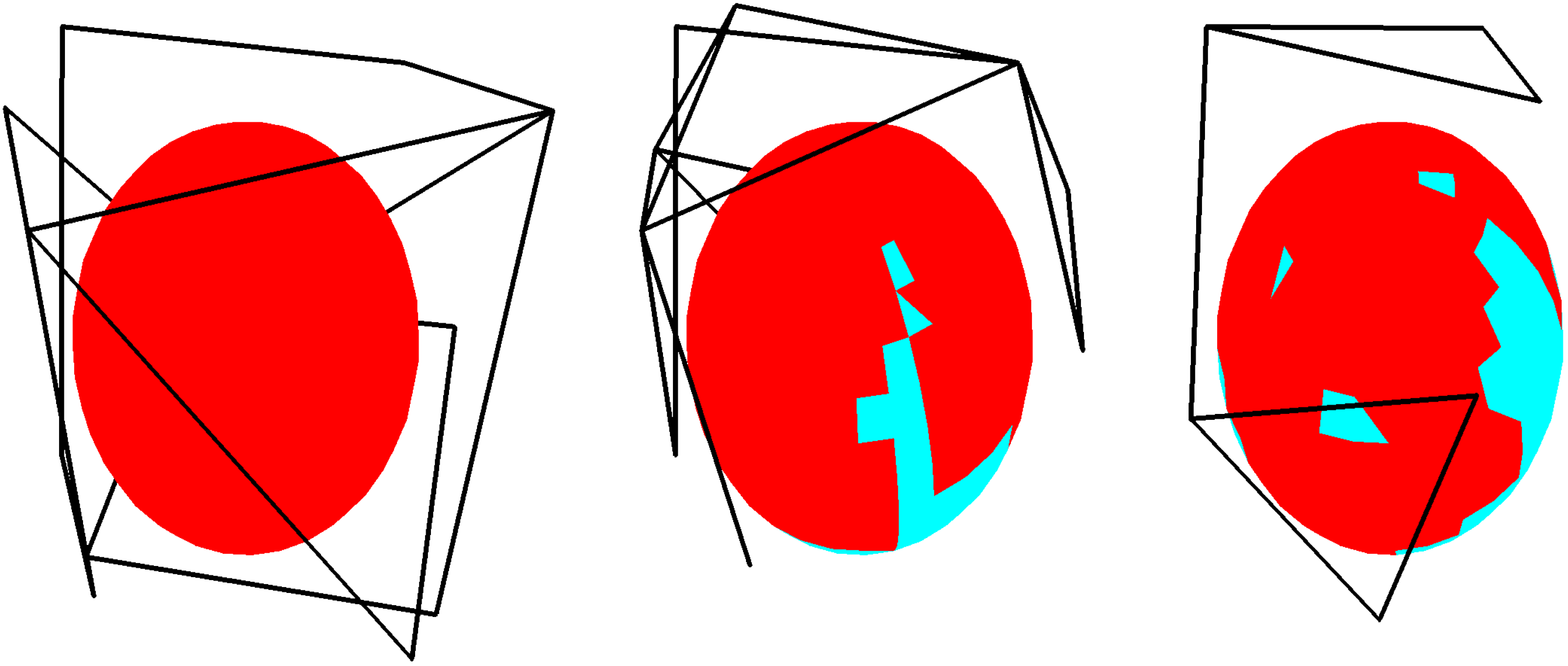}
        \caption{Sampling plans -- scores (0.0, 45.1), (0.17, 49.3) and (0.45, 31.2), plans resulting from $f$ = 1.0, 0.8 and 0.4, respectively.}
        \label{fig:sphere-sampling}
    \end{subfigure}
    ~ 
    \begin{subfigure}[b]{\textwidth}
        \includegraphics[width=\textwidth]{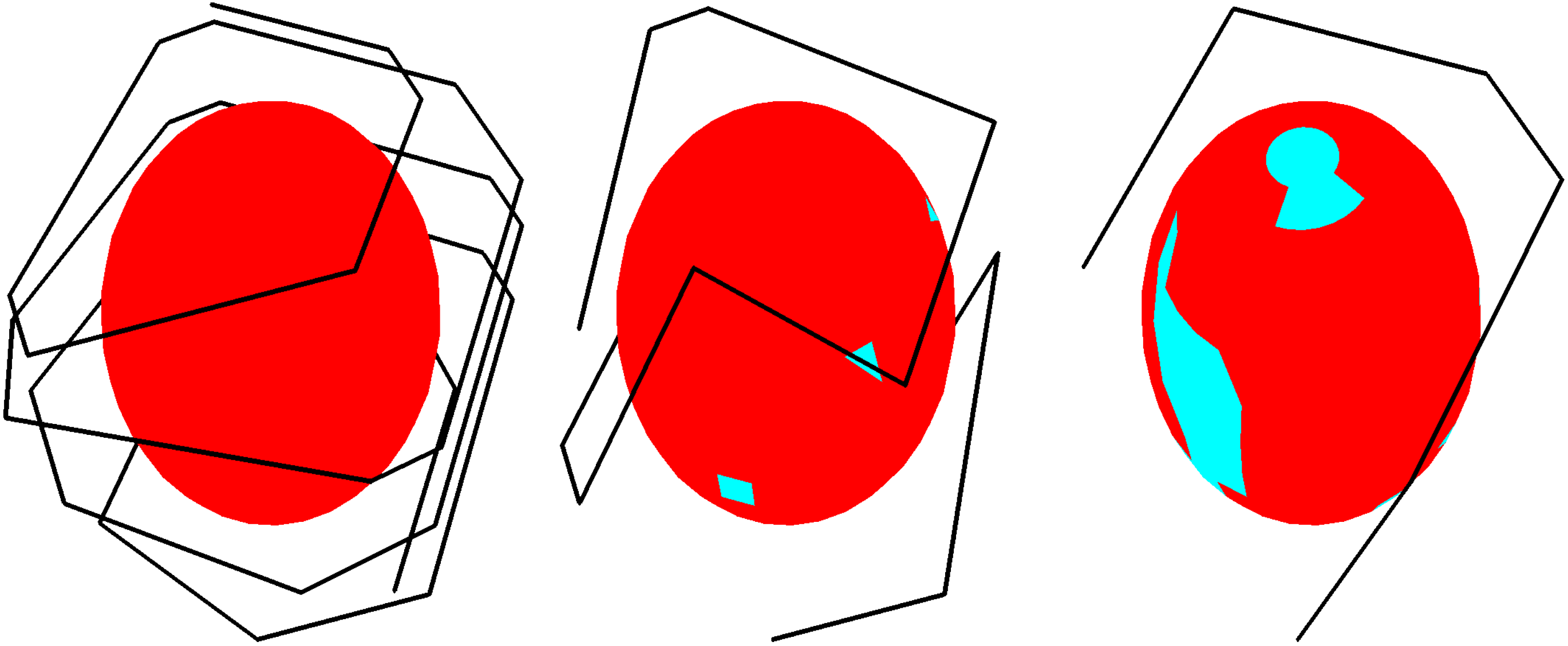}
        \caption{Seeded MOEA plans -- scores (0.0, 47.9), (0.04, 25.0) and (0.39, 9.8).}
        \label{fig:sphere-moea}
    \end{subfigure}
    \caption{\textbf{Sample plans for inspecting the sphere.} The entire set of plans from a single run of each algorithm is given in the Supplementary Material.}
    \label{fig:sphere_plans}
\end{figure}

\begin{figure}
    \centering
    \begin{subfigure}[b]{\textwidth}
        \includegraphics[width=\textwidth]{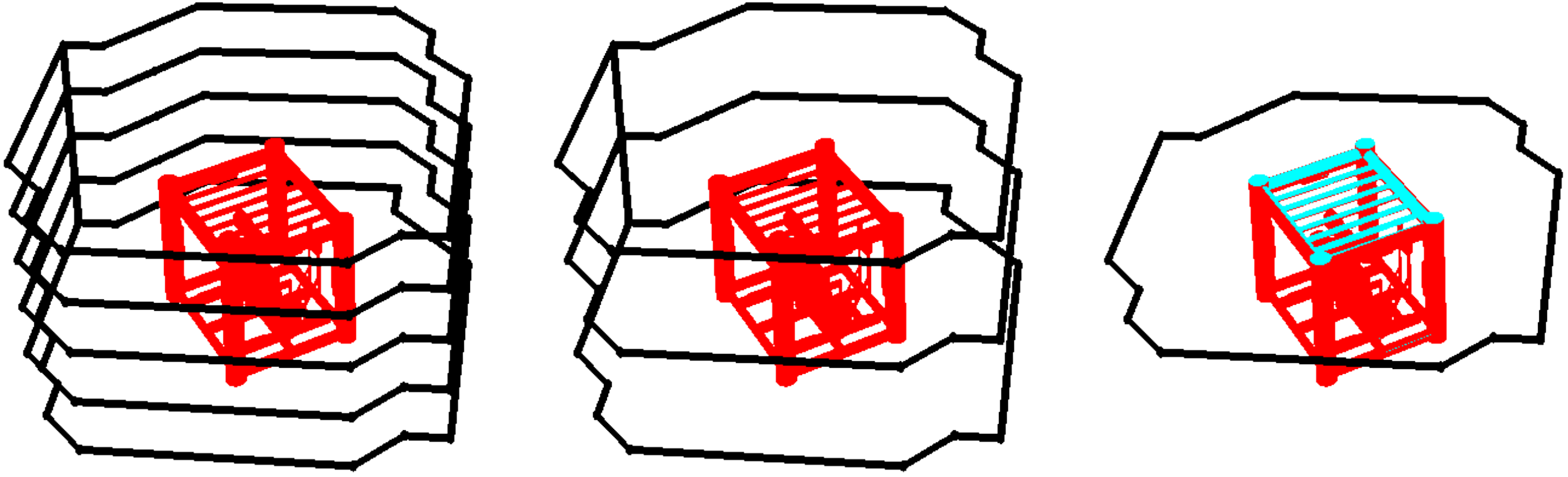}
        \caption{Circling plans -- scores (0.22, 50.1), (0.23, 29.6), (0.34, 8.3), plans resulting from $\Delta z$ = 1, 2 and 5, respectively.}
        \label{fig:pump-circling}
    \end{subfigure}
    ~ 
    \begin{subfigure}[b]{\textwidth}
        \includegraphics[width=\textwidth]{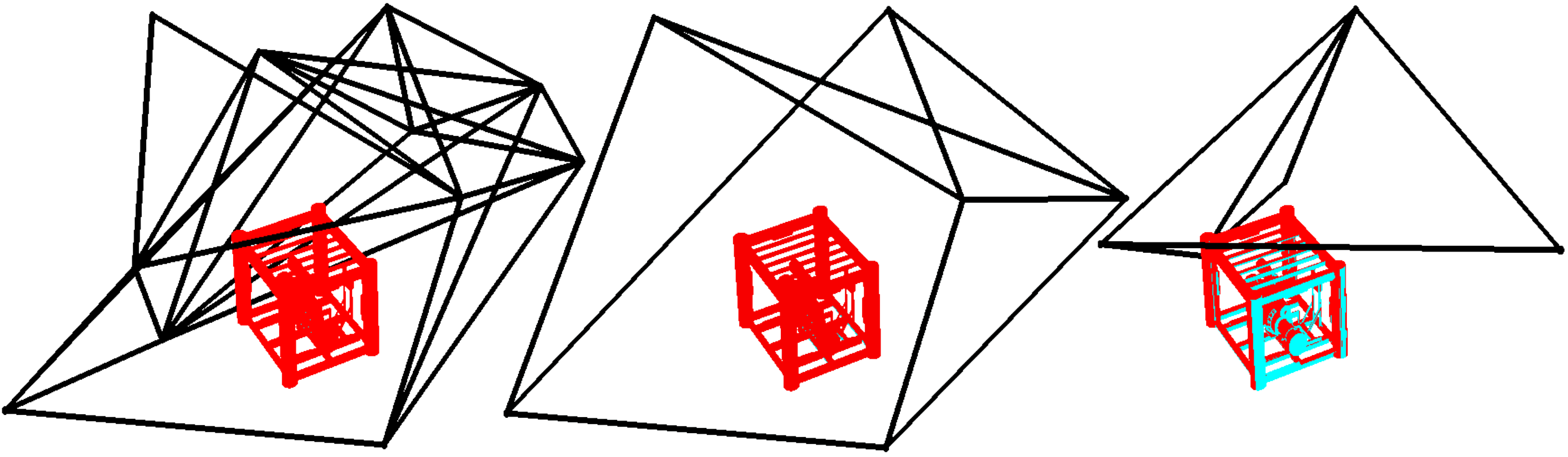}
        \caption{Sampling plans -- scores (0.20, 71.3), (0.31, 30.8), (0.39, 31.9), plans resulting from $f$ = 0.78, 0.64 and 0.38, respectively.}
        \label{fig:pump-sampling}
    \end{subfigure}
    ~ 
    \begin{subfigure}[b]{\textwidth}
        \includegraphics[width=\textwidth]{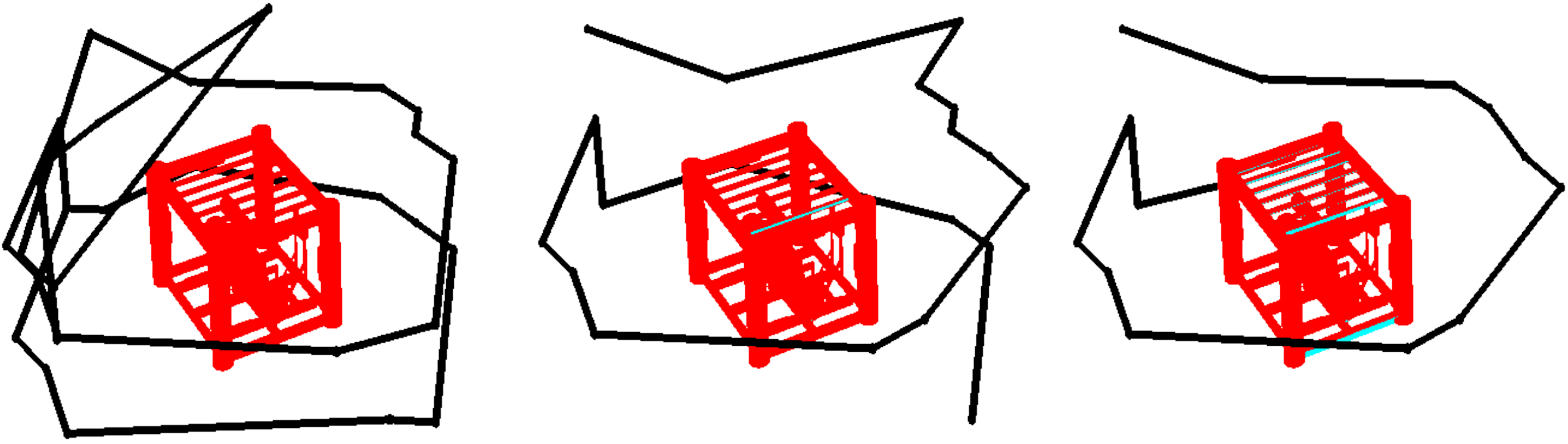}
        \caption{Seeded MOEA plans -- scores (0.22, 23.6), (0.23, 14.4), (0.25, 8.6)}
        \label{fig:pump-moea}
    \end{subfigure}
    \caption{\textbf{Sample plans for inspecting the SSIV.} The entire set of plans from a single run of each algorithm is presented in the Supplementary Material.}\label{fig:oil_pump_plans}
\end{figure}

\begin{figure}
    \centering
    \begin{subfigure}[b]{0.8\textwidth}
        \includegraphics[width=\textwidth]{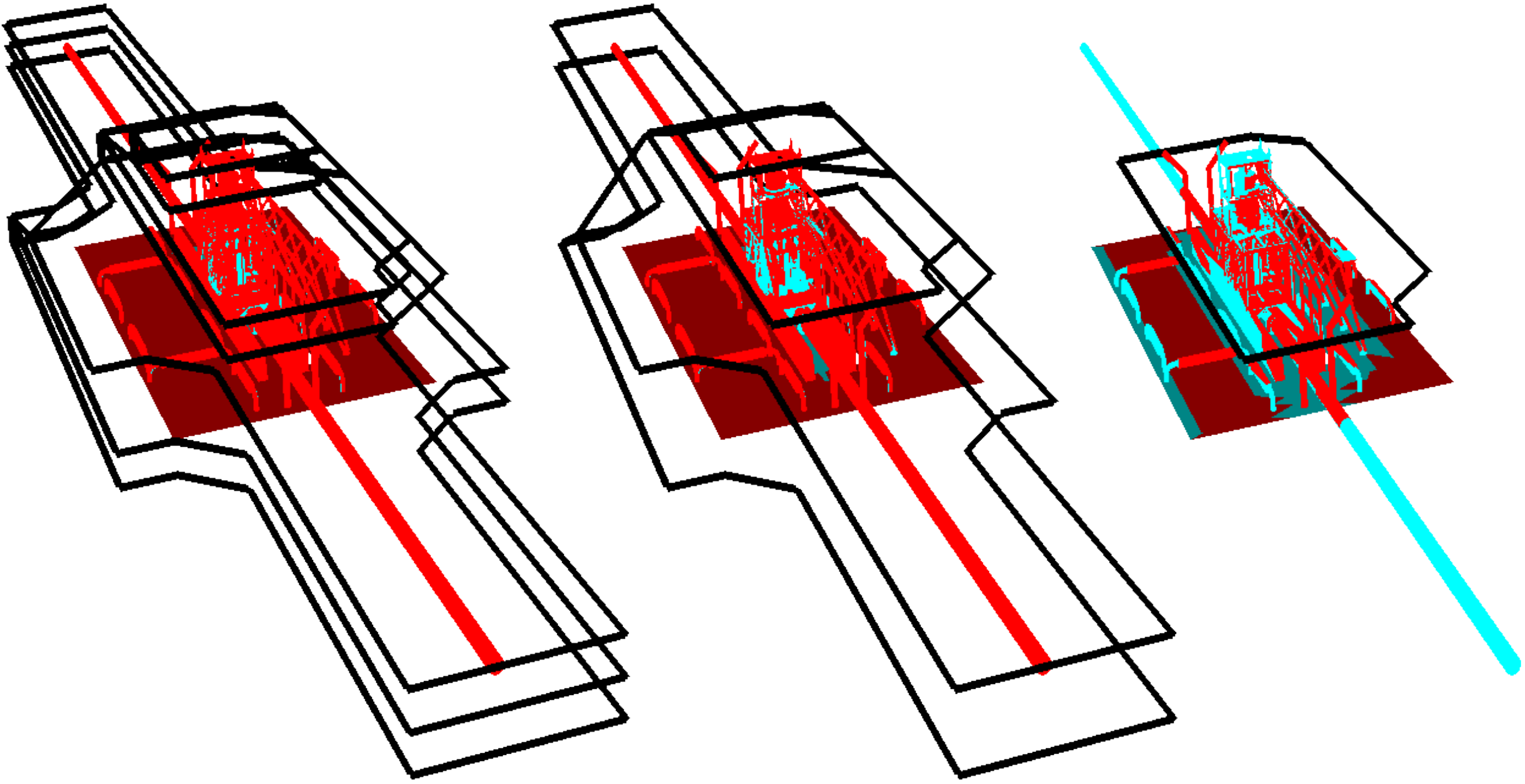}
        \caption{Circling plans -- scores (0.20, 168.8), (0.26, 103.5), (0.59, 13.3), plans resulting from $\Delta z$ = 1, 2 and 7, respectively.}
        \label{fig:manifold-circling}
    \end{subfigure}
    ~ 
    \begin{subfigure}[b]{0.8\textwidth}
        \includegraphics[width=\textwidth]{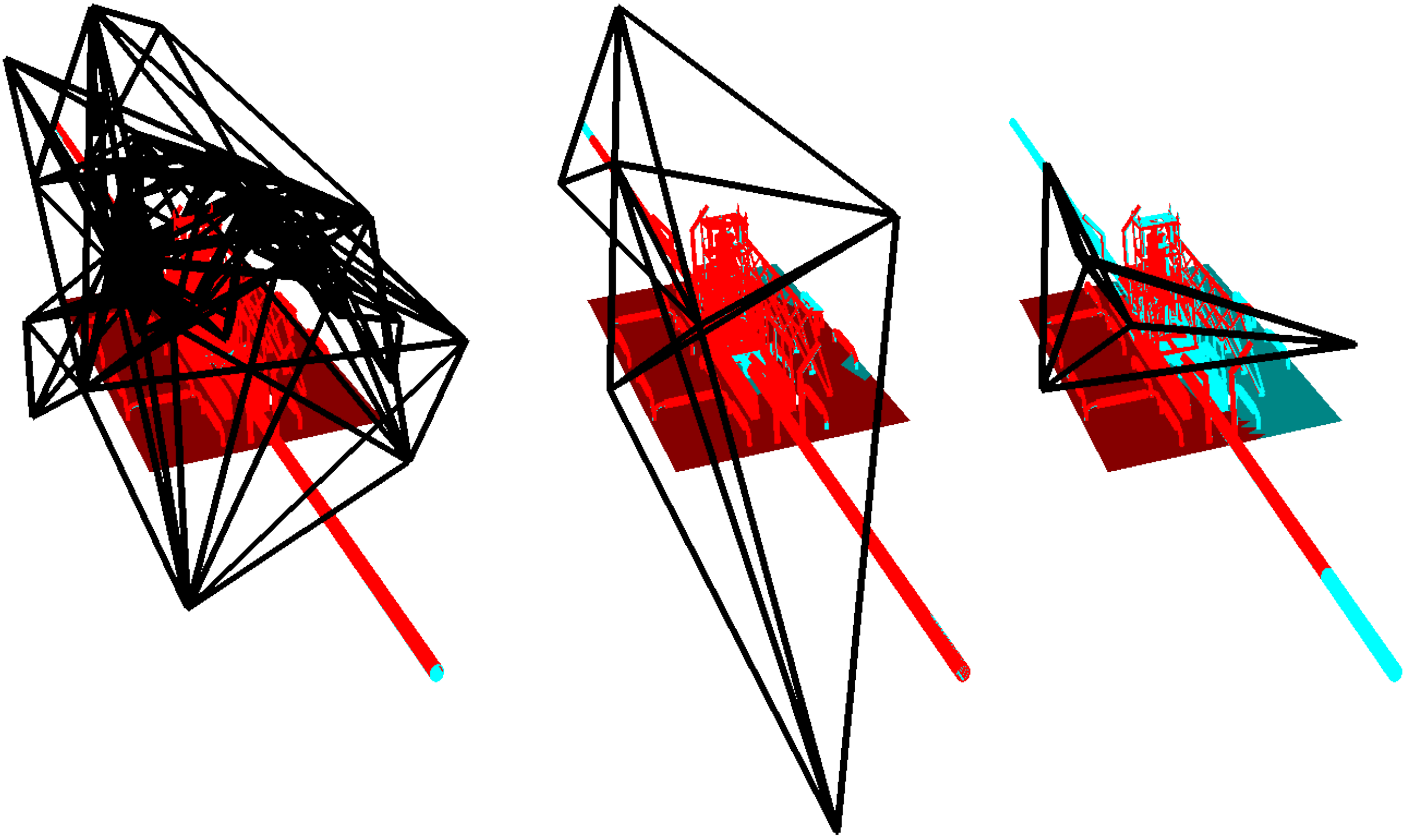}
        \caption{Sampling plans -- scores (0.15, 657), (0.28, 106), (0.60, 60.2), plans resulting from $f$ = 0.84, 0.69 and 0.32, respectively.}
        \label{fig:manifold-sampling}
    \end{subfigure}
    ~ 
    \begin{subfigure}[b]{0.8\textwidth}
        \includegraphics[width=\textwidth]{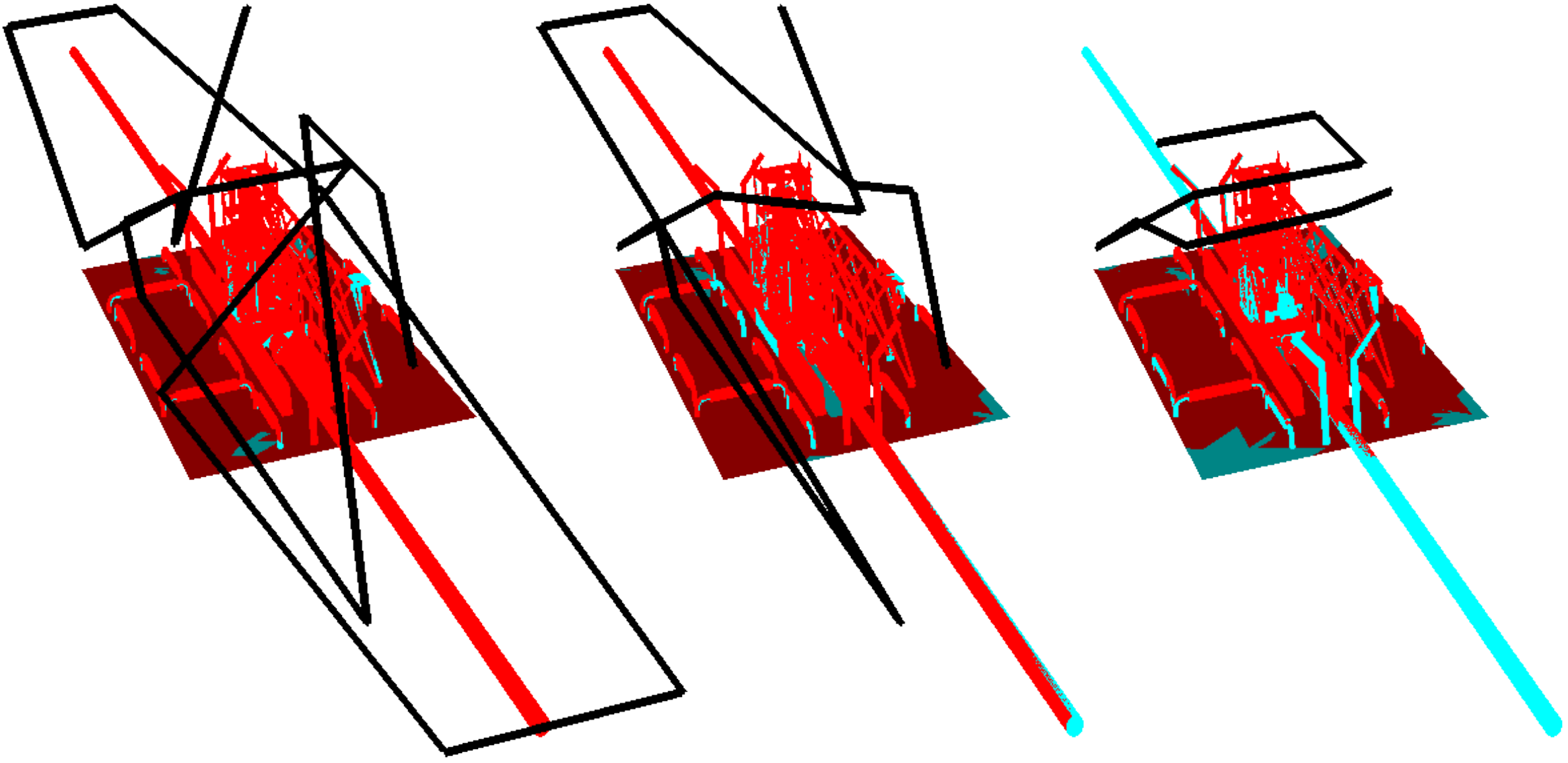}
        \caption{Seeded MOEA plans -- scores (0.20, 71.5), (0.27, 44.0), (0.41, 18.2).}
        \label{fig:manifold-moea}
    \end{subfigure}
    \caption{\textbf{Sample plans for inspecting the oil manifold.} The entire set of plans from a single run of each algorithm is presented in the Supplementary Material.}\label{fig:manifold_plans}
\end{figure}

However, we are in the unusual situation of wanting to compare the performance of traditional single-objective optimization methods with our multiobjective inspection path planner. We found the most suitable way to analyze the results to be visualizing the energy usage and coverage of \emph{all generated plans} in a single plot, and further analyzing the differences between the planners by visually inspecting the generated plans. Figure~\ref{fig:all_pareto} summarizes the results of all experiments, while Figures~\ref{fig:sphere_plans},~\ref{fig:oil_pump_plans} and~\ref{fig:manifold_plans} show representative example plans, along with the parameters used to generate them. Note that the different MOEA-plans were all generated by the same parameters (as outlined in Section~\ref{sec:multiobjective_inspection}), whereas the other planners require changing parameters to adjust the coverage degree. 

The sphere is the only of the three inspection targets that allows for \emph{complete coverage} inspection planning, and all three methods generated complete coverage inspection plans for this structure (leftmost plans in Figure~\ref{fig:sphere_plans}). The sampling-based planner produced the most efficient complete coverage plan, with an energy usage of 35.2, with the most efficient MOEA plan following close behind with a best complete coverage plan at 38.7. The best complete coverage plan generated by the circling planner was far less efficient, with an energy usage of 65.5. These scores all fall in the very leftmost column of Figure~\ref{fig:scores_sphere}, which plots the scores of all generated inspection plans for the sphere.


Studying the \emph{incomplete coverage} plans, we can observe the MOEA generating a well-diversified set of plans for different balances between energy usage and coverage (Figure~\ref{fig:all_pareto}). By optimizing energy and coverage together the evolutionary algorithm has reached solutions achieving \emph{more coverage} for their energy effort than the two other planning methods. In particular, the performance of the sampling-based planner suffers when dealing with incomplete coverage. It typically generates plans with far higher energy usage than the other methods when dealing with incomplete coverage (Figure~\ref{fig:all_pareto}).

For practical inspection missions, we are often interested in slightly reducing coverage if that can yield large reductions in the energy cost. A strength of the multiobjective inspection planner is that it, by generating a large and diversified population of plans, allows just this. Looking at Figures~\ref{fig:sphere-moea} and~\ref{fig:pump-moea}, we see how evolved plans can achieve large energy savings with only small reductions in coverage.  Circling-based plans (Figures~\ref{fig:sphere-circling} and Figures~\ref{fig:pump-circling}) also allow such energy savings to some degree, but their strict shape makes them unable cover the structures as energy-efficiently as the MOEA-plans (Figure~\ref{fig:all_pareto}).

With regards to the practical execution of plans, vehicle constraints will naturally affect the applicability of each planning method -- for instance, the cost of turns will affect how important it is for a plan to be smooth. In general, the circling plans will probably be the simplest to carry out due to their highly regular structures (Figures~\ref{fig:sphere-circling},~\ref{fig:pump-circling} and~\ref{fig:manifold-circling}). The MOEA-plans (in particular the \emph{seeded} ones) also generally demonstrate quite clean and smooth paths (Figures~\ref{fig:sphere-moea},~\ref{fig:pump-moea} and~\ref{fig:manifold-moea}). The sampling-based plans, however, have sharp turns and very irregular structures (Figures~\ref{fig:sphere-sampling},~\ref{fig:pump-sampling} and ~\ref{fig:manifold-sampling}).

\subsection{The Effect of Seeding}
\label{sec:seeding_result}

\begin{figure}

    \centering
    \begin{subfigure}[b]{0.55\textwidth}
\includegraphics[width=\textwidth]{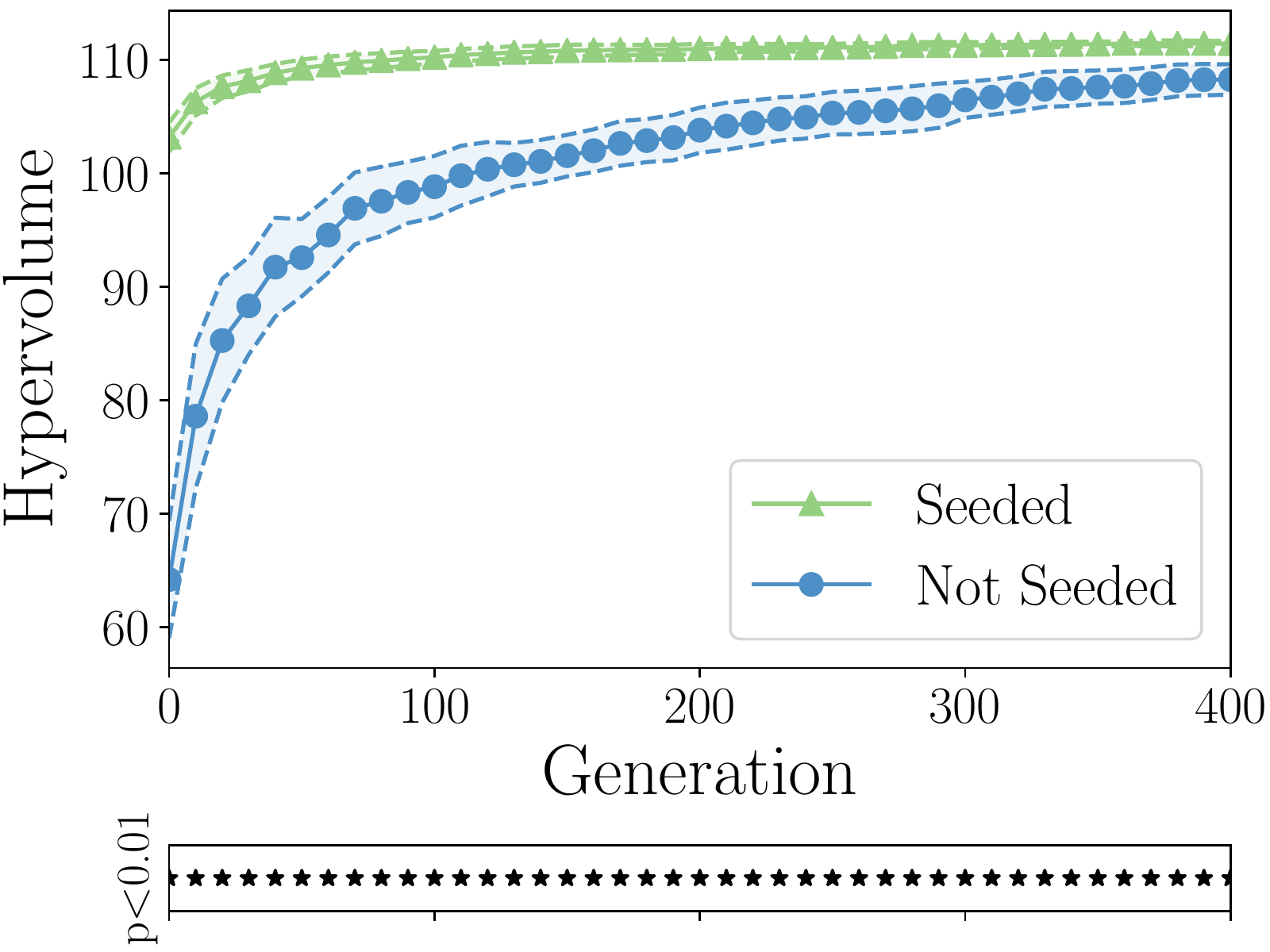}
        \caption{Inspection target: Sphere.}
        \label{fig:hypervol_sphere}
\end{subfigure}

\begin{subfigure}[b]{0.55\textwidth}
\includegraphics[width=\textwidth]{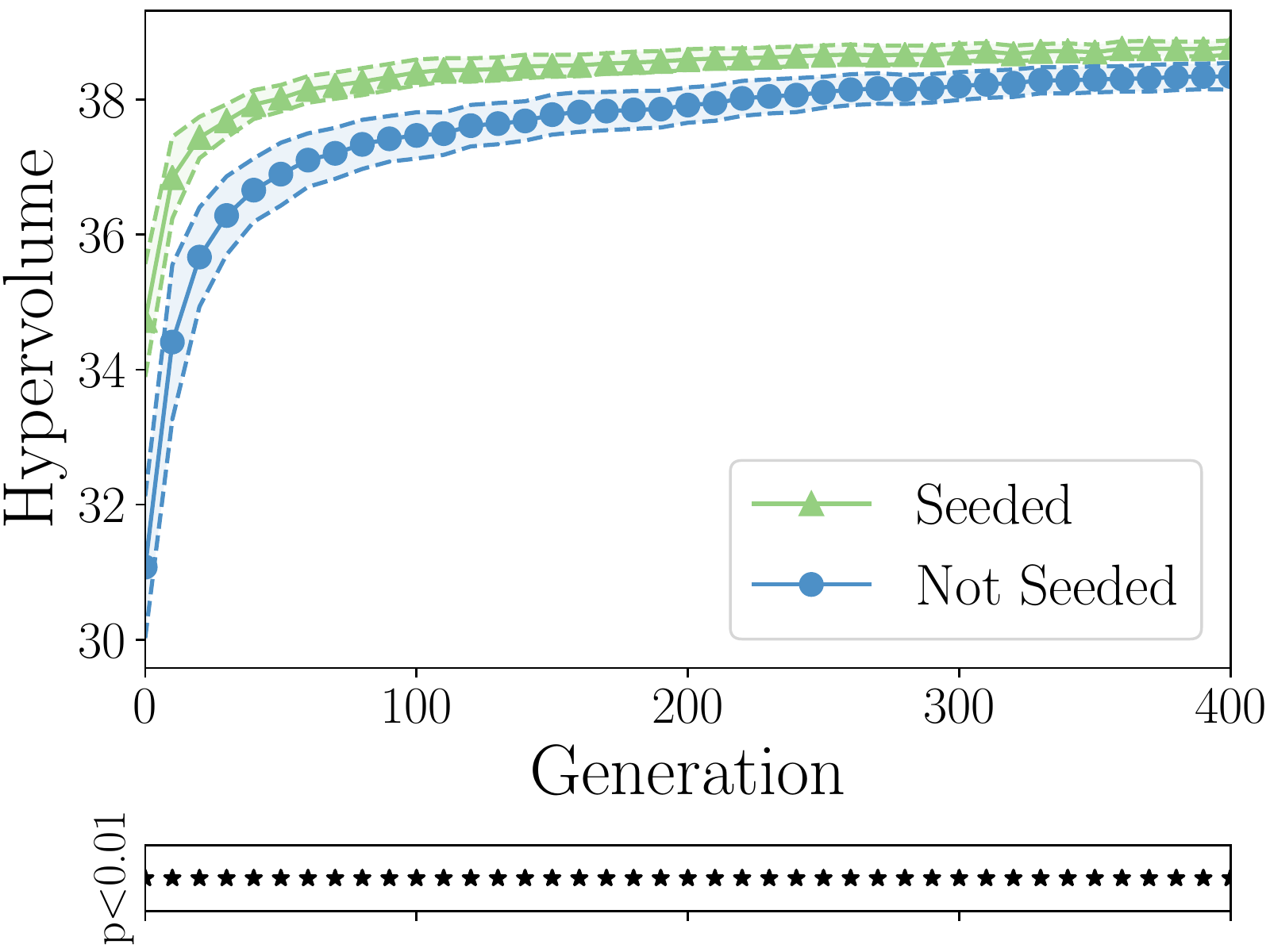}
        \caption{Inspection target: SSIV.}
        \label{fig:hypervol_pump}
\end{subfigure}

\begin{subfigure}[b]{0.55\textwidth}
\includegraphics[width=\textwidth]{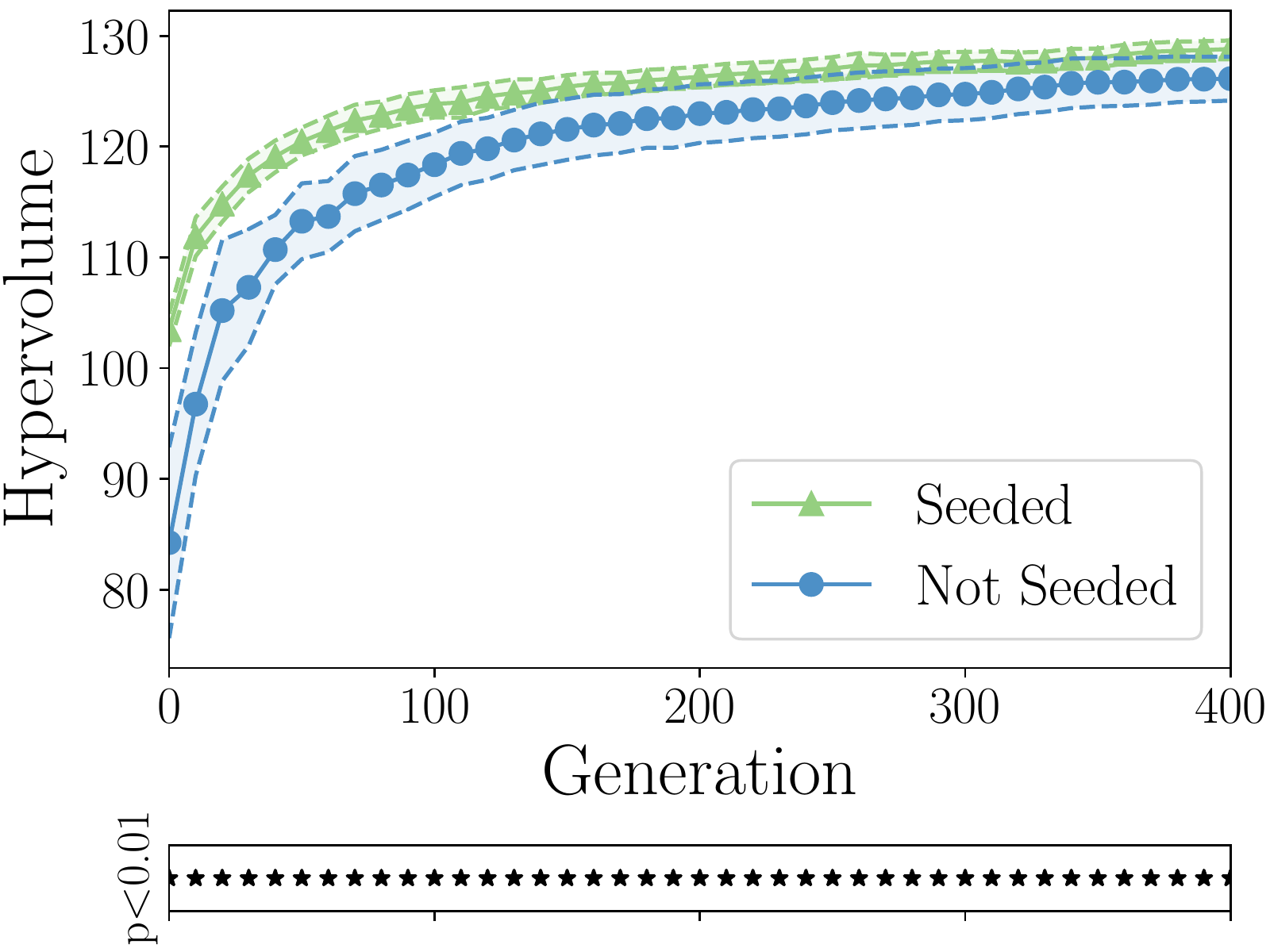}
        \caption{Inspection target: Large manifold.}
        \label{fig:hypervol_manifold}
\end{subfigure}
\caption{\textbf{Hypervolume Comparisons plans generated by NSGA-II with and without seeding.}  Plots show median values over 20 runs, along with standard deviations. Stars under each plot indicate significant differences between the two treatments.}
\label{fig:hypervolumes}
\end{figure}

The effect of seeding on the optimization of inspection path plans was studied by running the NSGA-II algorithm 20 times with parts of its population initiated from the seeds, and 20 times with only random initial individuals. Throughout each evolutionary run, we gathered the energy and coverage scores of all individuals, so that we could plot their hypervolume score as evolution proceeded. Figure~\ref{fig:hypervolumes} shows median hypervolumes throughout the evolutionary optimization for each of the three inspection targets.

The seeded evolutionary runs have a significantly higher hypervolume score throughout evolution -- indicating that seeding produces plans that display a more optimal balance between energy and coverage. We can also see this in the final evolved populations in Figure~\ref{fig:all_pareto}: Seeded runs result in scores closer to the optimum. The improvement produced by seeding is naturally greatest in early generations, and non-seeded runs seem to be close to catching up towards the end of the evolutionary run. One of the beneficial effects of seeding is therefore to produce \emph{good plans faster}.

\begin{figure}
\centering

    \begin{subfigure}[b]{0.7\textwidth}
\includegraphics[width=\textwidth]{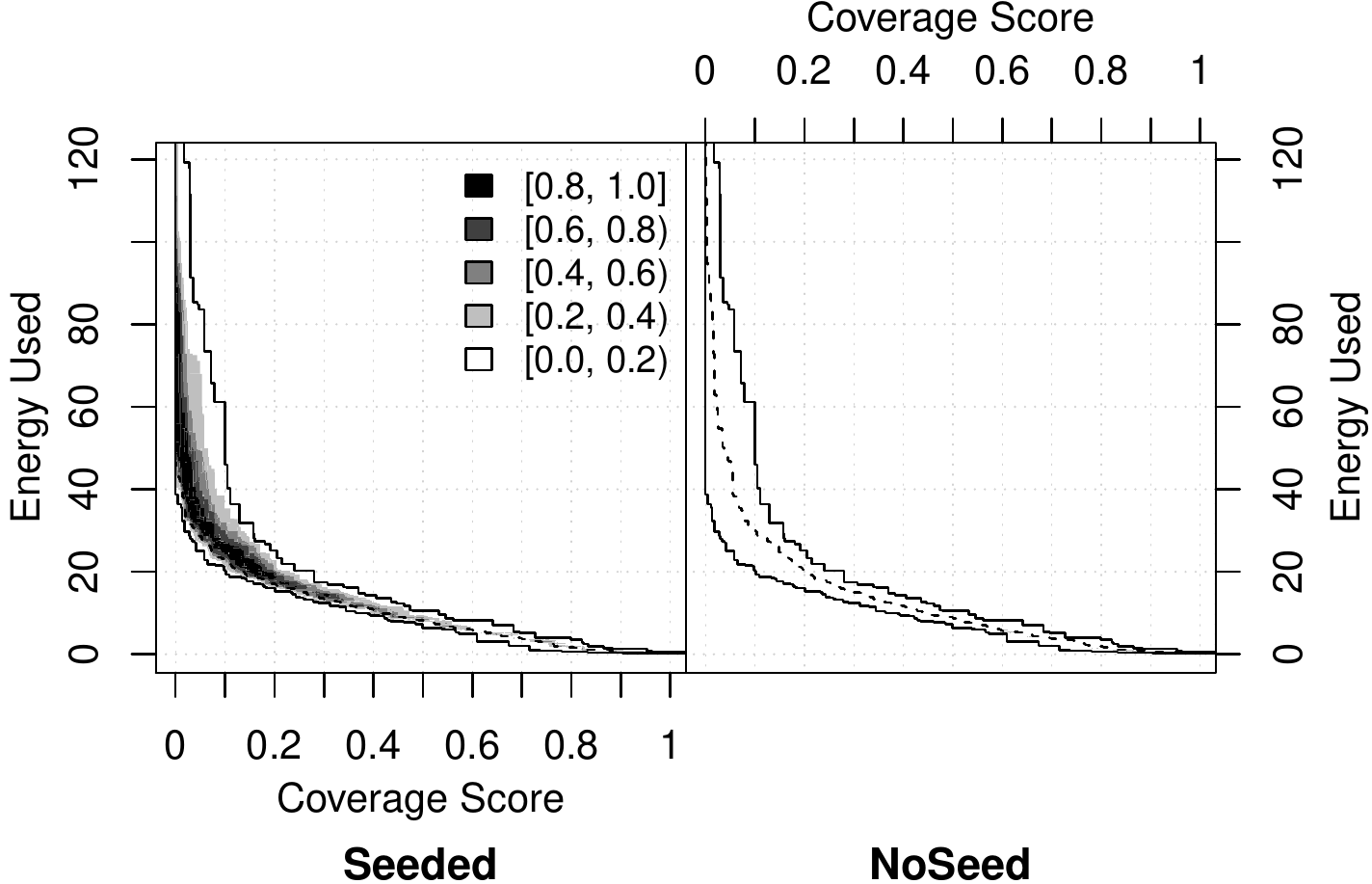}
\caption{Inspection target: Sphere}
\label{fig:eaf_sphere}
\end{subfigure}

    \begin{subfigure}[b]{0.7\textwidth}
\includegraphics[width=\textwidth]{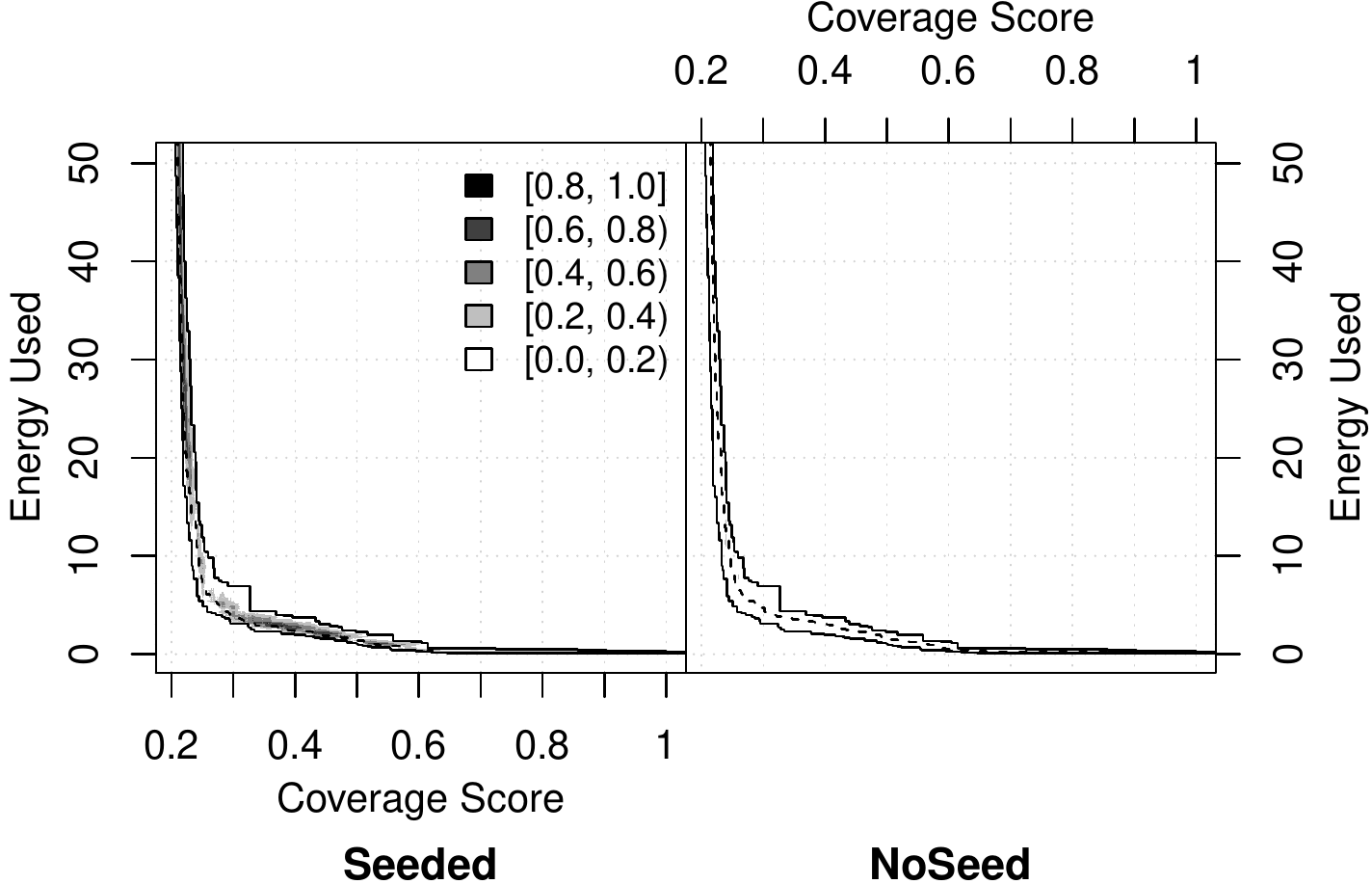}
\caption{Inspection target: SSIV}
\label{fig:eaf_pump}
\end{subfigure}

    \begin{subfigure}[b]{0.7\textwidth}
\includegraphics[width=\textwidth]{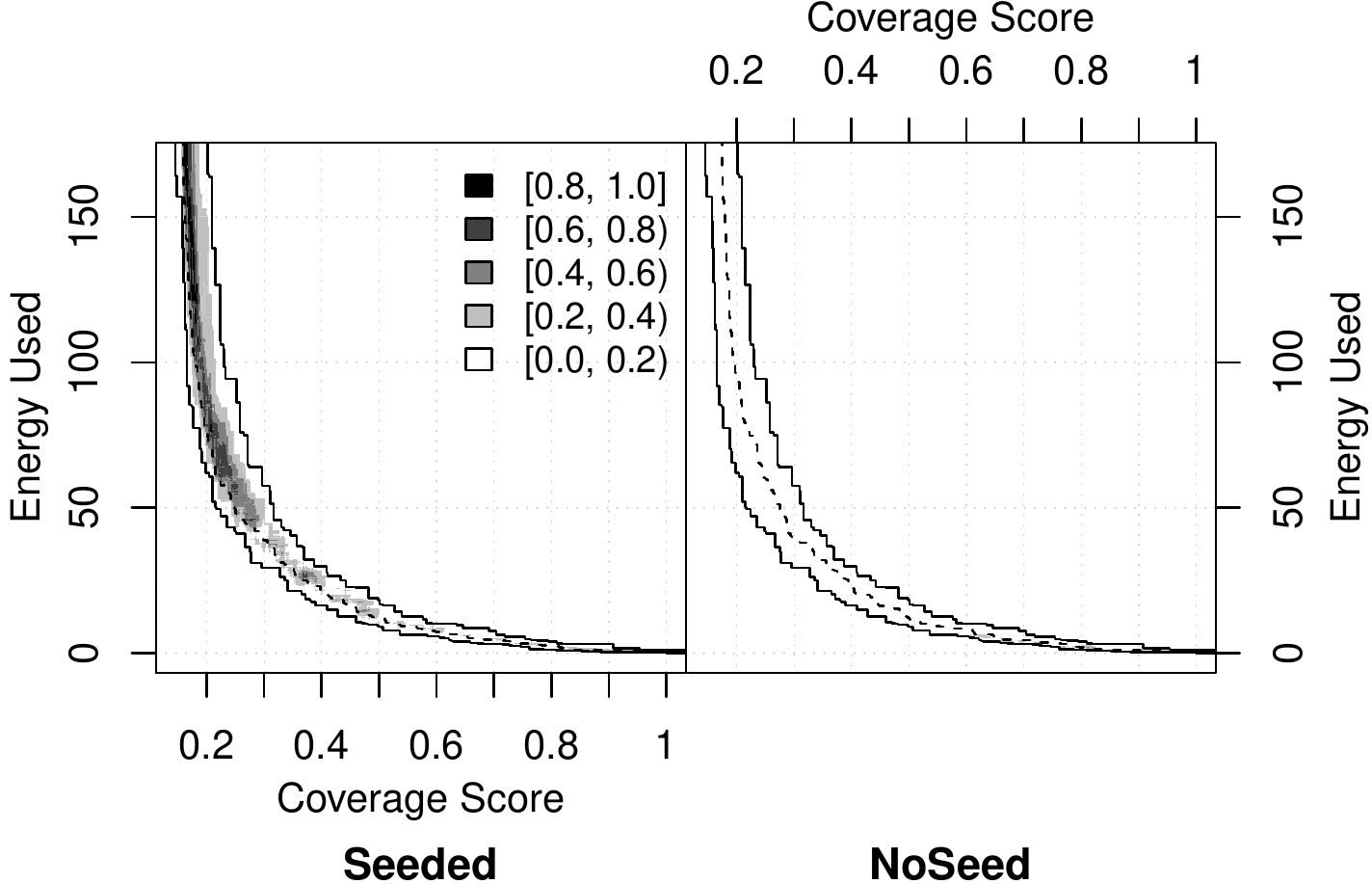}
\caption{Inspection target: Large manifold}
\label{fig:eaf_manifold}
\end{subfigure}
\caption{\textbf{Differences between empirical attainment functions (EAFs) from seeded and non-seeded evolutionary optimization.} Dashed lines show median attainment surface, solid lines show overall best and worst attainment surfaces (over seeded \emph{and} un-seeded runs). Gray shading indicates magnitude of difference in attainment function. See text for details. Plots made with the EAF Graphical Tools Package~\cite{Lopez-Ibanez2010b}}
\label{fig:attainment}
\end{figure}

Another way to analyze the performance difference between seeded and non-seeded evolutionary runs is to study their difference in \emph{empirical attainment functions}~\cite{Fonseca1996, GrunertDaFonseca2001}. Figure~\ref{fig:attainment} shows the attainment surfaces of seeded and non-seeded runs of the evolutionary algorithm on the three structures. Upper and lower attainment surfaces indicate the best and worst areas attained in objective space by any evolutionary run, whereas the dotted line shows the median attainment surface for each treatment (seeded or non-seeded). The shaded areas indicate where there is a difference between the attainment functions of the two treatments. The left figures show differences in favor of seeded runs, that is, areas in objective space that were attained more often by seeded than by non-seeded runs. Right figures show differences in favor of non-seeded runs. These plots confirm the tendency we saw in Figure~\ref{fig:all_pareto}: Seeded runs reach solutions of a higher quality than those reached by non-seeded runs. We can also see another interesting trend here: The advantage of seeded runs is highest for the longest plans (those with a high energy usage). The seeds contain a good selection of long plans circling the structure several times, which may be the reason why seeded runs are especially good at finding long plans.

\begin{figure}
    \centering
    \begin{subfigure}[b]{0.48\textwidth}
        \includegraphics[width=\textwidth]{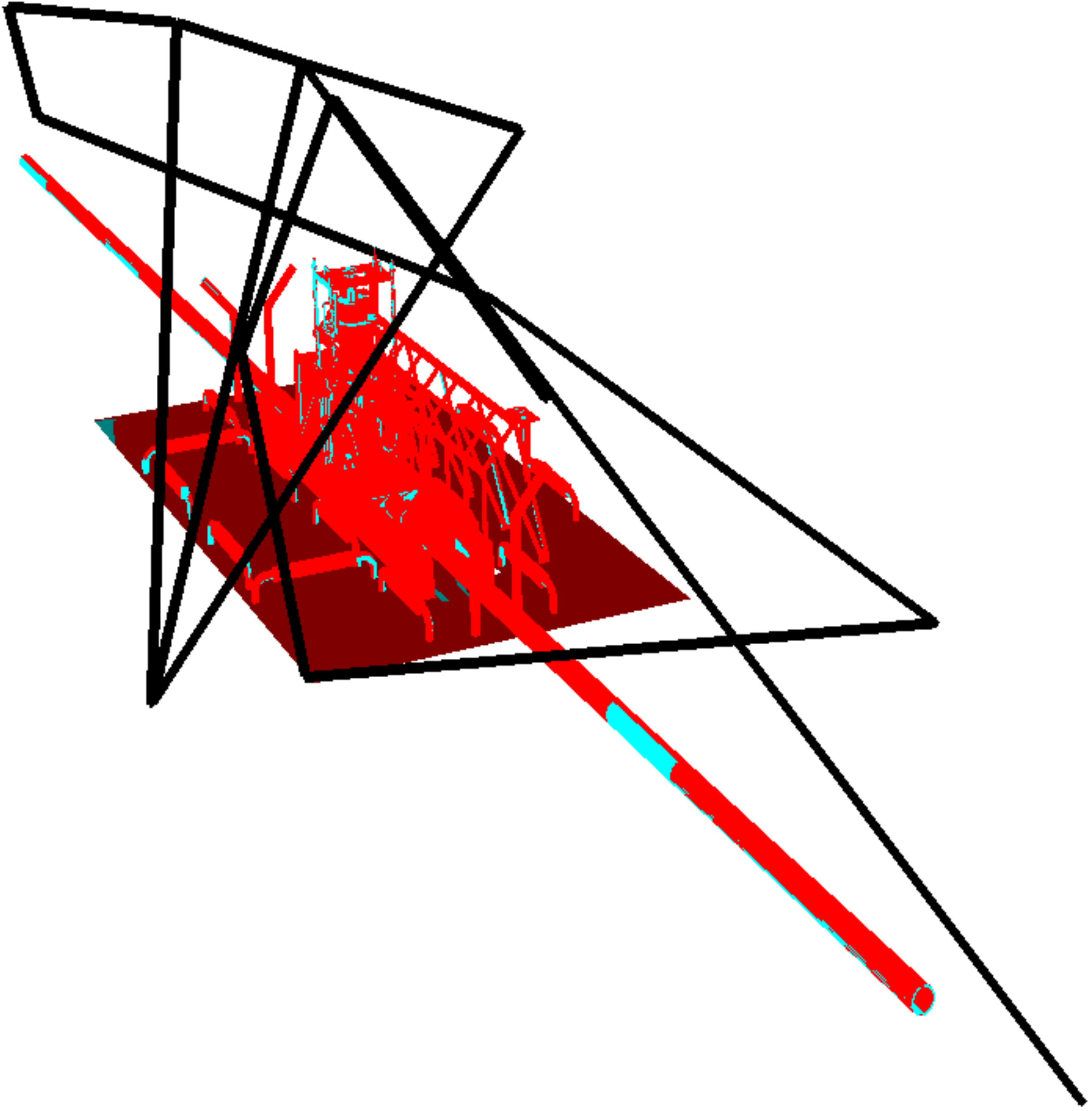}
        \caption{Plan evolved without seeding -- scores (0.23, 71.9).}
    \end{subfigure}
    ~ 
    \begin{subfigure}[b]{0.48\textwidth}
        \includegraphics[width=\textwidth]{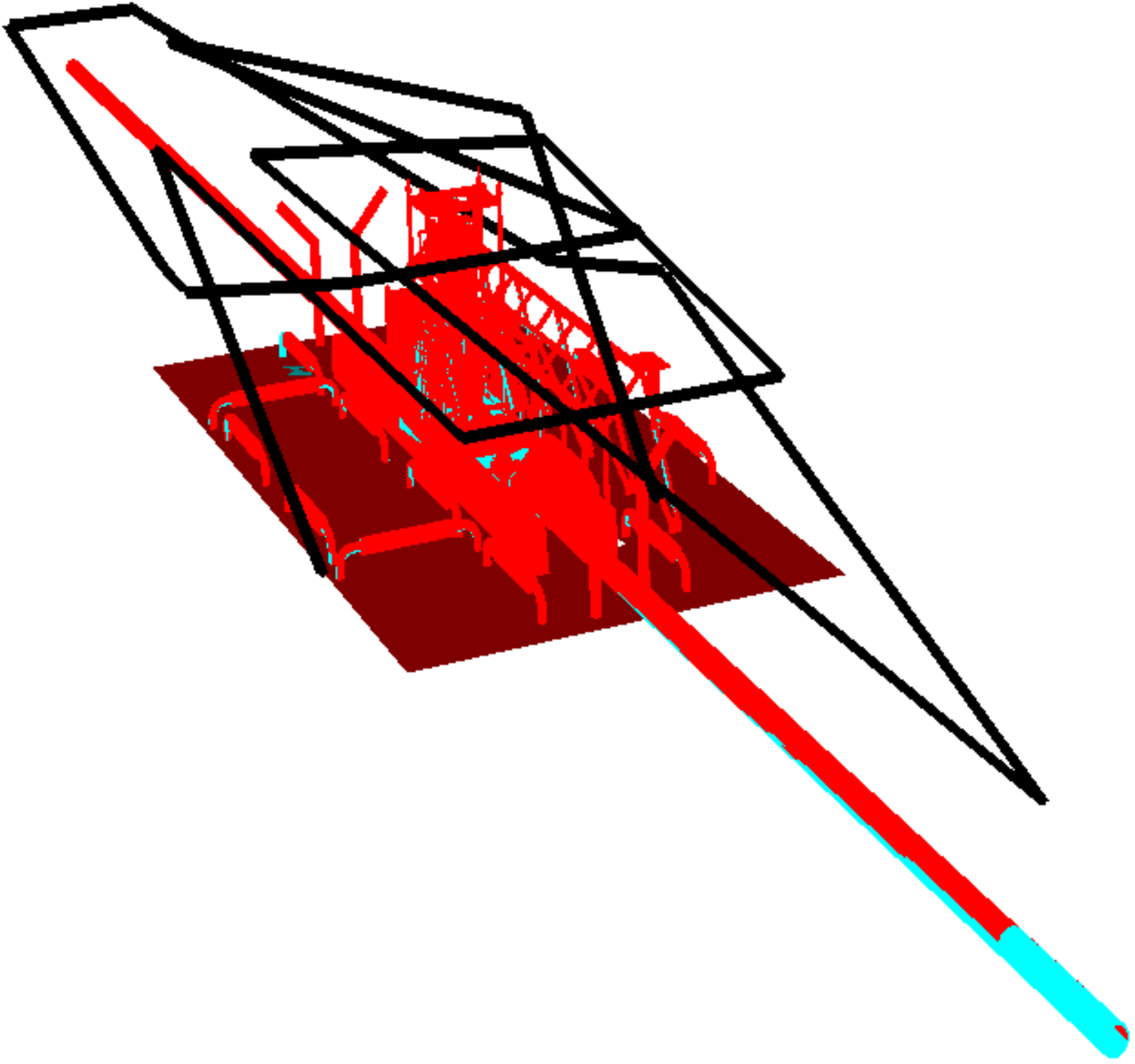}
        \caption{Plan evolved with seeding -- scores (0.21, 73.2).}
    \end{subfigure}
    
    \caption{\textbf{The results on evolved plan structure of seeding.} When seeding (right), evolved plans retain much of the clean structures of the seed plans, whereas evolutionary runs without seeding (left) have a more irregular structure.}
    \label{fig:seed_vs_no_seed}
\end{figure}

A final benefit of seeding is that it produces plans with a more regular, more easily interpretable structure. Even when plans have the same performance, those resulting from seeded runs have a very different structure (Figure~\ref{fig:seed_vs_no_seed}). The cleaner structure of seeded plans is likely to be very useful in practice, as it makes it easier to interpret the plan, and thus to select among candidate plans.

\subsection{Runtime}

The circling planner naturally performs best with regards to runtime, since it performs no optimization. As outlined in Section~\ref{sec:circling_sweeps}, all circling paths for a single inspection target are \emph{generated simultaneously}. All circling plans were generated in 56, 49 and 81 seconds for the sphere, SSIV and manifold, respectively.

\begin{figure}
\centering

    \begin{subfigure}[b]{0.55\textwidth}
\includegraphics[width=\textwidth]{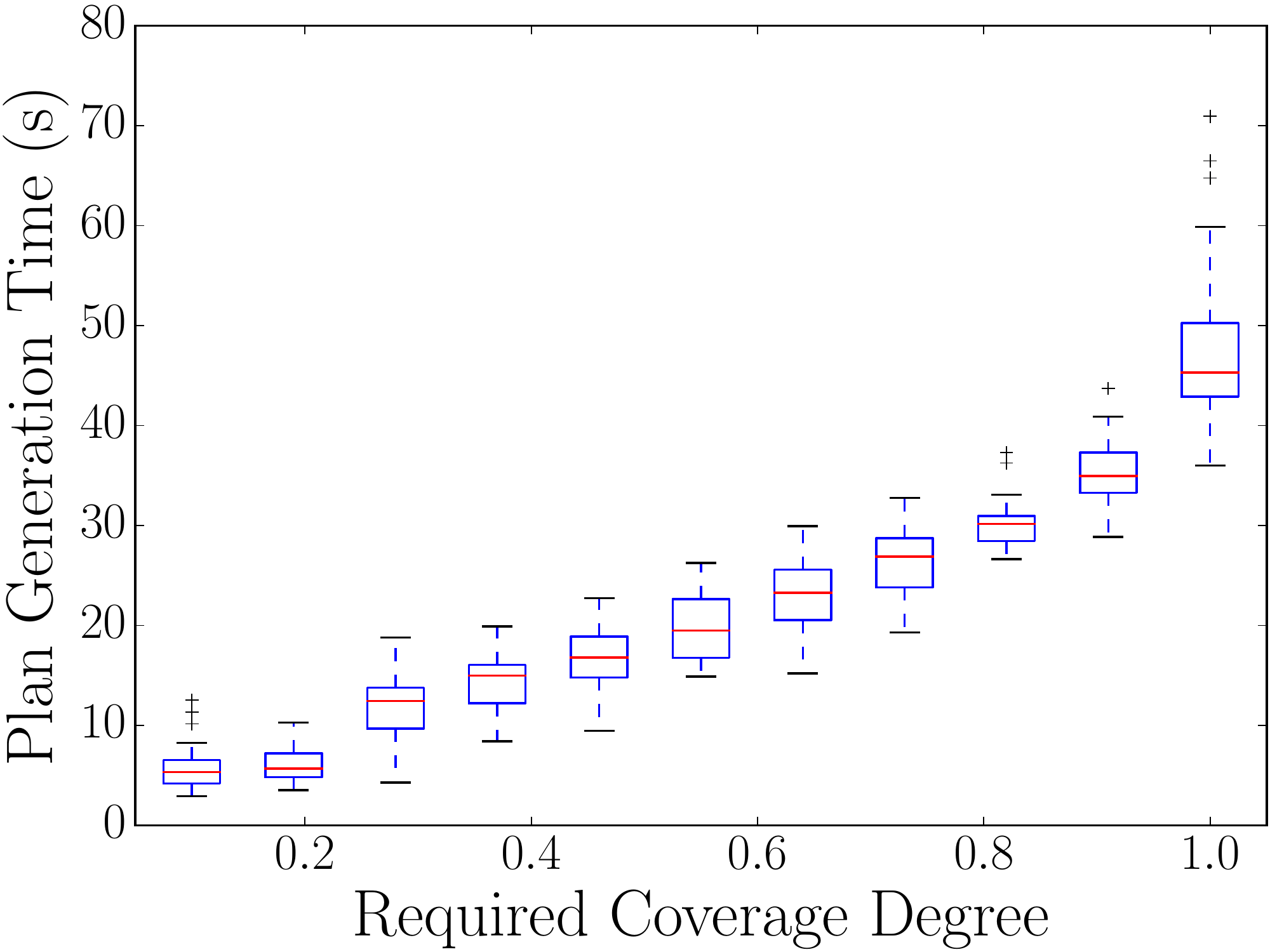}
\caption{Inspection target: Sphere}
\label{fig:timing_sphere}
\end{subfigure}

    \begin{subfigure}[b]{0.55\textwidth}
\includegraphics[width=\textwidth]{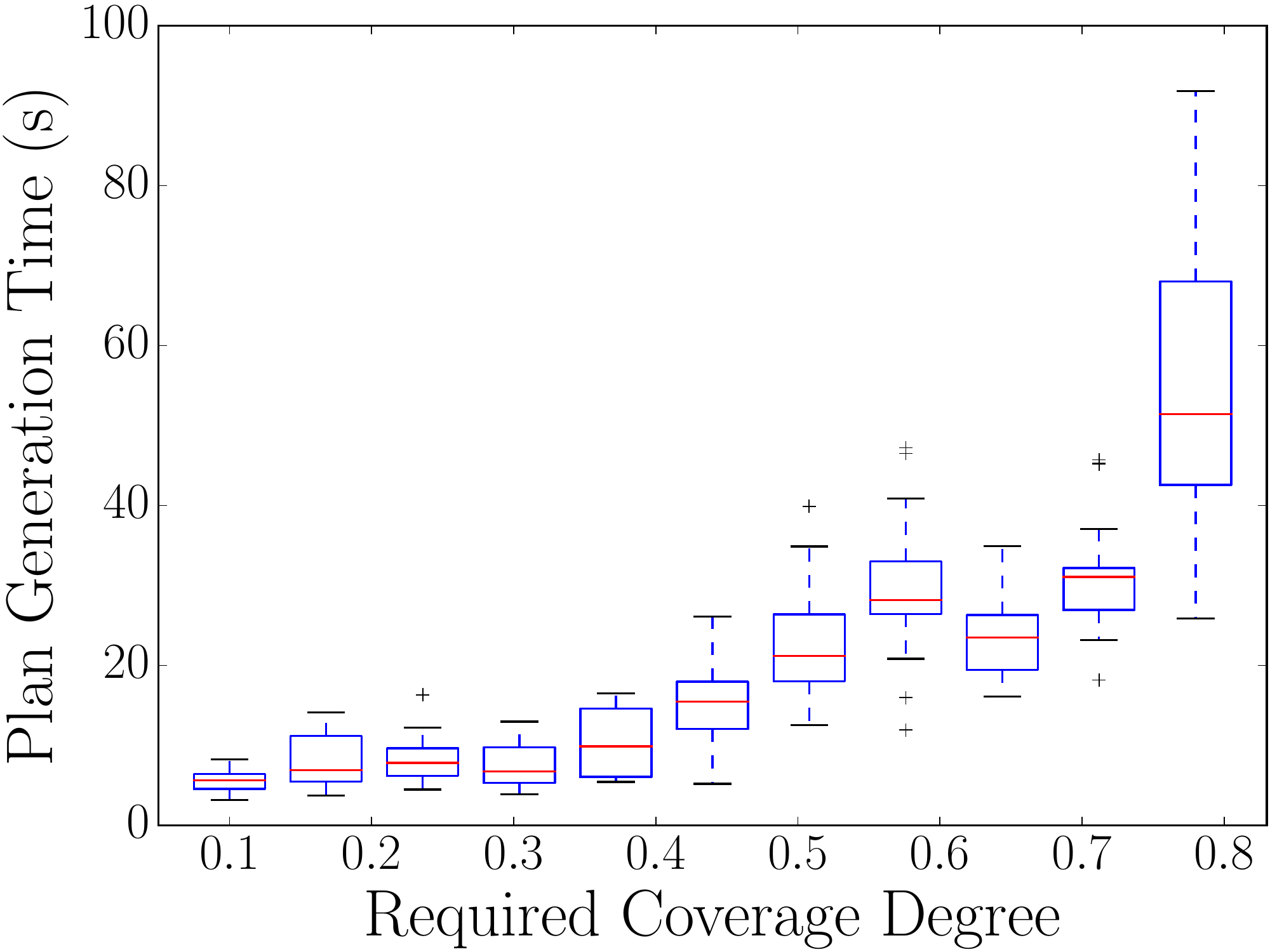}
\caption{Inspection target: SSIV}
\label{fig:timing_pump}
\end{subfigure}

    \begin{subfigure}[b]{0.55\textwidth}
\centering
\includegraphics[width=\textwidth]{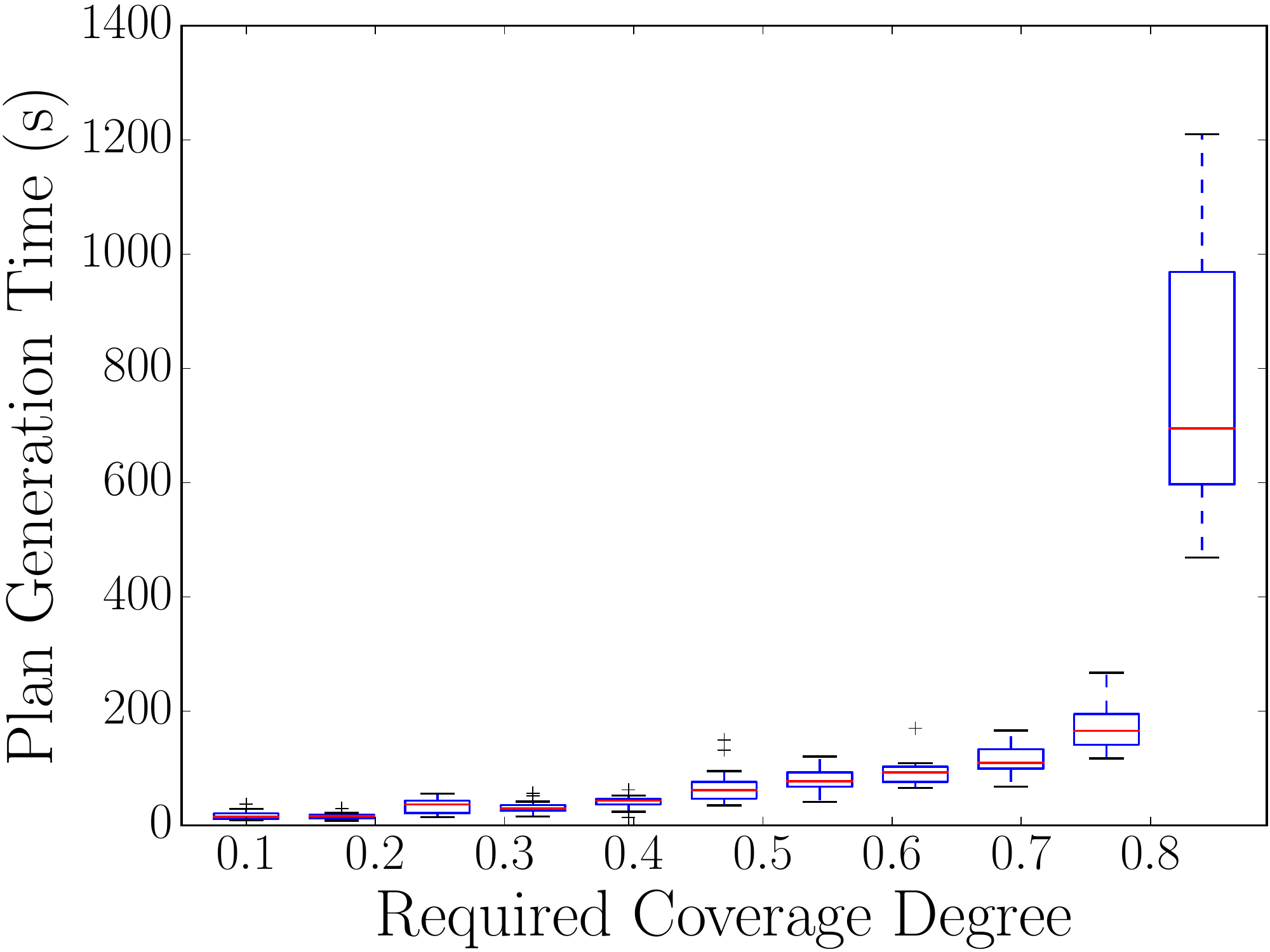}
\caption{Inspection target: Large manifold}
\label{fig:timing_manifold}
\end{subfigure}
\caption{\textbf{Runtime of the generalized sampling-based coverage planning as function of required coverage degree.} Boxes display median and first and third quartiles across 20 runs, whiskers indicate $1.5\times IQR$ (interquartile range).}
\label{fig:timing_info}
\end{figure}

The runtime of the sampling-based planner depends on the desired ~\emph{coverage degree}, $f$. Naturally, a higher-coverage plan takes longer to generate, as it requires the sampling of more edges, and the solution of a larger integer programming problem. Figure~\ref{fig:timing_info} shows the runtime of the sampling-based planner as a function of $f$ on the three inspection targets. For the circle, where complete coverage is possible, the increase in runtime for higher-coverage plans is not too dramatic, but for the two more complex structures (and especially the large manifold), we see extreme increases in runtime when nearing the highest possible levels of coverage.

The evolutionary algorithm generally has a higher runtime than the two other methods (Table~\ref{table:ea_runtime}). The plots of hypervolume, however (Figure ~\ref{fig:hypervolumes}), indicates that high-quality plans were available also early in the evolutionary run (especially for \emph{seeded} runs), suggesting that when necessary, the runtime of this algorithm can be reduced significantly for only small reductions in plan quality. The seeded EA-runs tend to have a higher runtime than un-seeded ones, but the difference is only significant ($p<0.05$ according to the Mann-Whitney U test) on the sphere. Probably this is because the seeded runs start with a population of longer plans, and therefore have to spend more time in total evaluating plan coverage.

\begin{table}
\centering
\begin{tabular}{ l c r}
 \hline
Structure & Seeded EA & Un-seeded EA\\
 \hline
 \textbf{Sphere}\\
 Runtime & 701.3 [681.13, 797.9] & 595 [577.9, 649.0] \\
 Evaluations & 3043.5 [3007.0, 3080.0] & 3072.5 [3035.0, 3122.5] \\
 \textbf{Manifold}\\
 Runtime   & 1053.5 [1013.5, 1073.9] & 999.2 [901.4, 1087.9]\\
 Evaluations   & 3156.5 [3115.0, 3174.0] & 3194.0 [3168.0, 3248.5]\\
 \textbf{SSIV}\\
 Runtime   & 479.3 [307.2, 491.4] & 481.4 [472.9, 487.6]\\
 Evaluations   & 2964.0 [2929.5, 3009.5] & 2966.0 [2959.0, 3017.0]\\
 \hline
\end{tabular}
\caption{\textbf{Runtimes and evaluation counts for the evolutionary algorithm.} Median runtimes (in seconds) and number of fitness evaluations for the seeded and un-seeded NSGA-II runs on all three structures. Brackets show 95\% bootstrapped confidence intervals of the median.}
\label{table:ea_runtime}
\end{table}

All runtime measurements were made on an LG A410 laptop with 8 GB DDR3 RAM, Intel Core i5 CPU, and a GeForce 310M GPU, Running 64-bit Ubuntu 14.04

\subsection{MOEA/D}
\label{sec:moead_results}

\begin{figure}
\centering
    \begin{subfigure}[b]{0.55\textwidth}
\includegraphics[width=\textwidth]{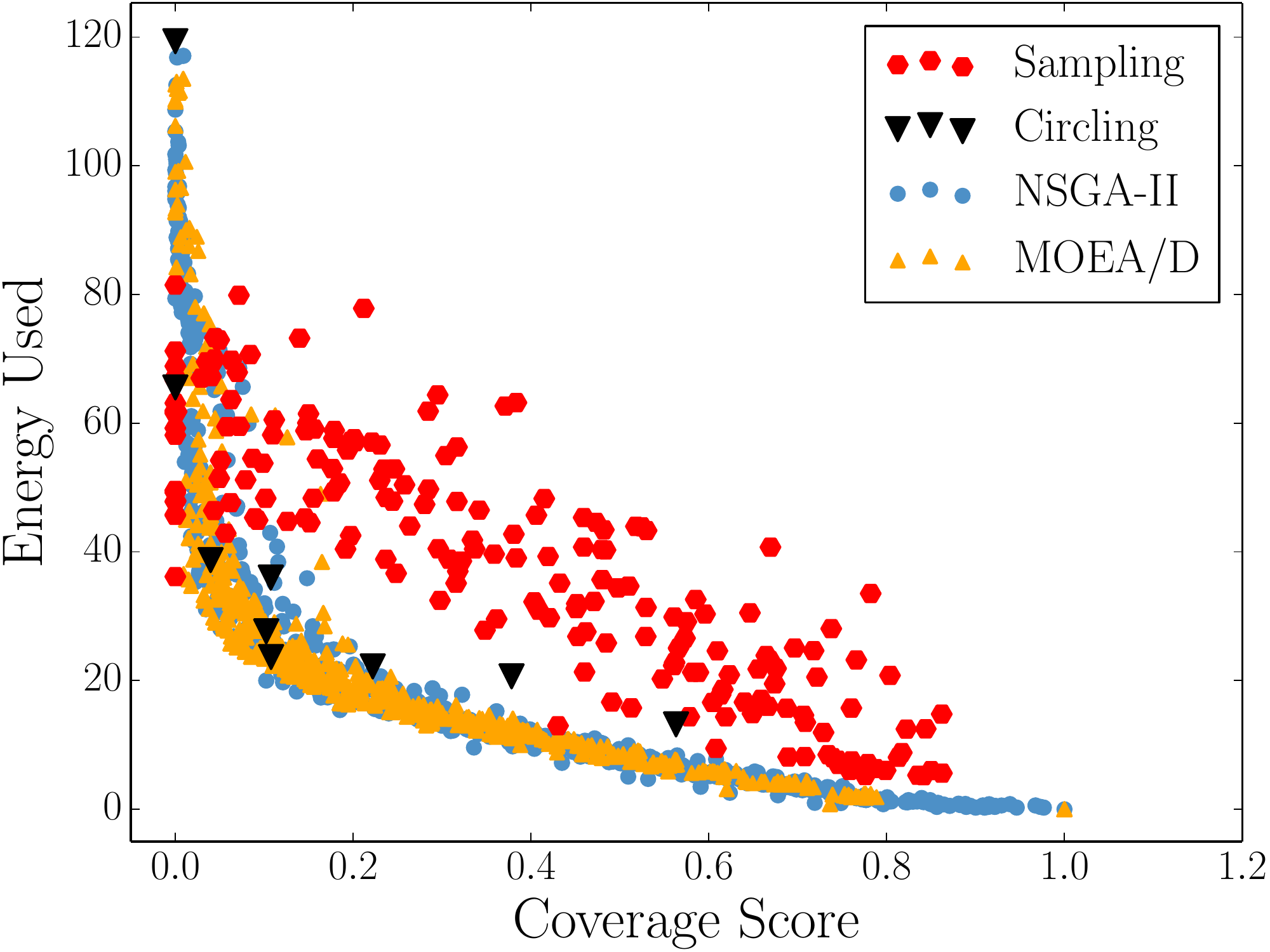}
\caption{Inspection target: Sphere}
\label{fig:scores_sphere_moead}
\end{subfigure}

    \begin{subfigure}[b]{0.55\textwidth}
\includegraphics[width=\textwidth]{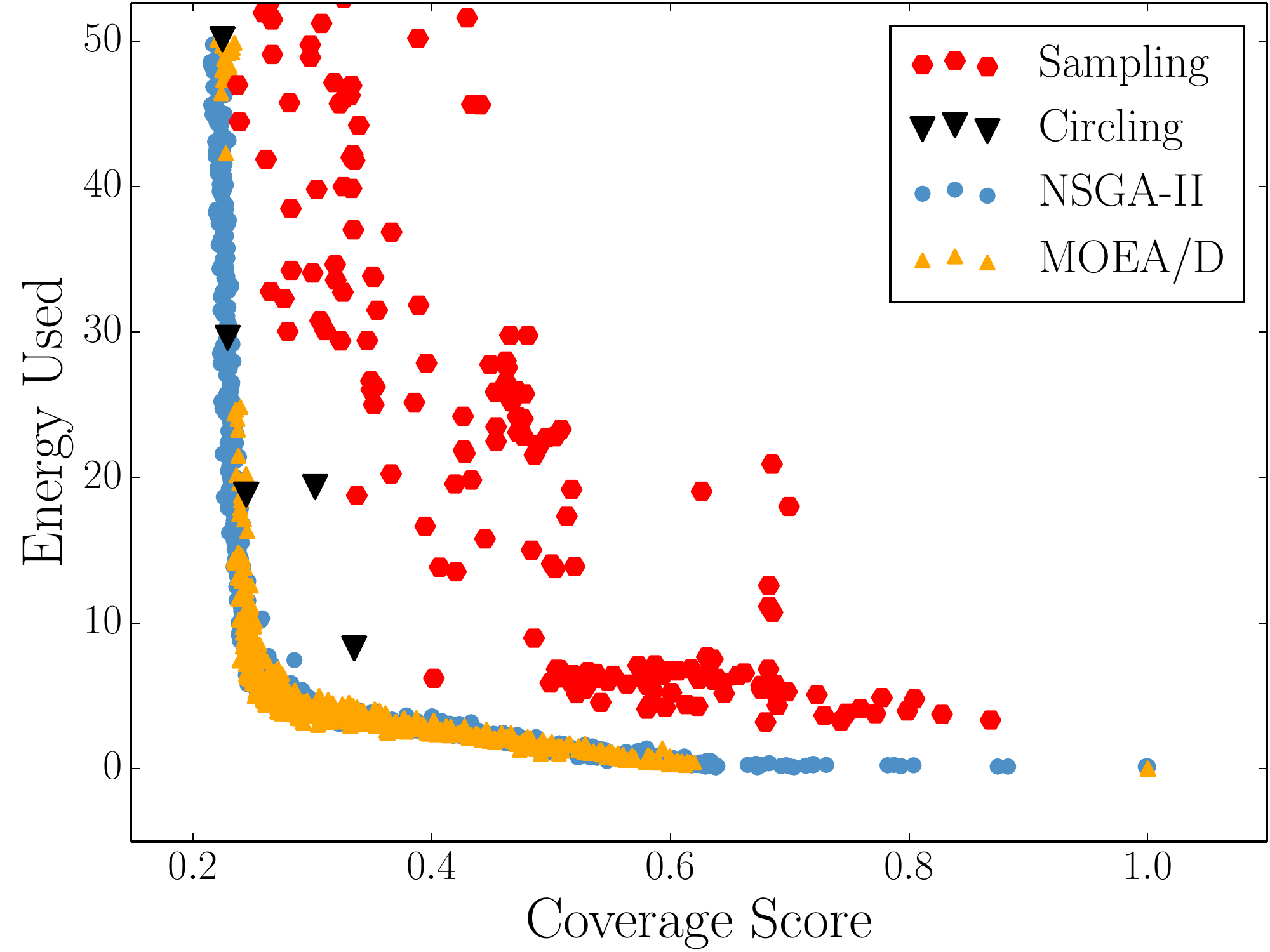}
\caption{Inspection target: SSIV}
\label{fig:scores_pump_moead}
\end{subfigure}

\begin{subfigure}[b]{0.55\textwidth}
\includegraphics[width=\textwidth]{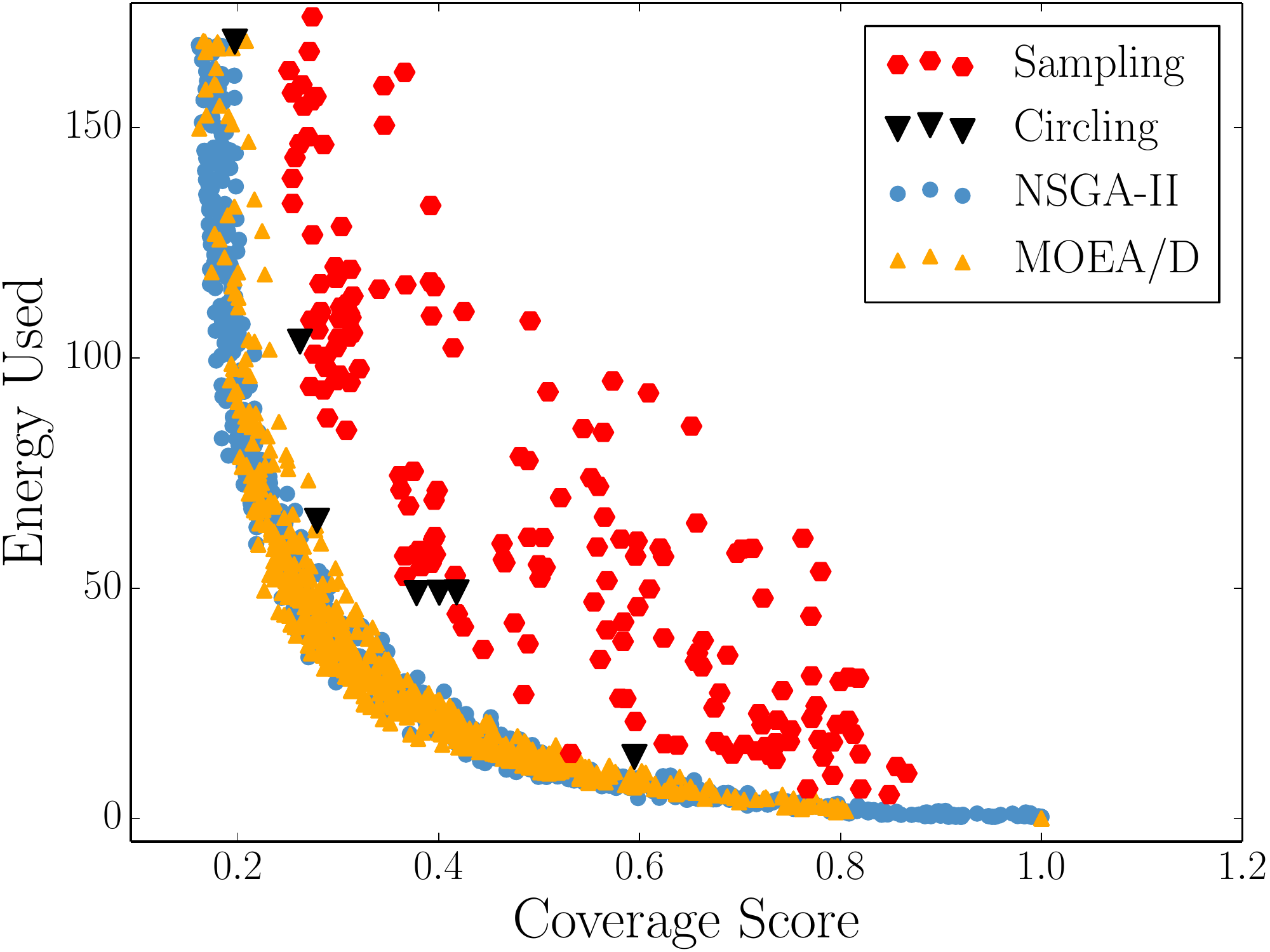}
\caption{Inspection target: Large manifold}
\label{fig:scores_manifold_moead}
\end{subfigure}
\caption{\textbf{Comparison of the two multi-objective evolutionary algorithms NSGA-II and MOEA/D.} Both were run 20 times to generate these results. Results from the sampling and circling-based planners are repeated from Figure~\ref{fig:all_pareto} for comparison. All plans in this figure were produced without seeding.}
\label{fig:pareto_moead}
\end{figure}

The experiments on evolving inspection plans described above applied the NSGA-II algorithm. However, to increase our confidence that the findings are valid for MOEAs \emph{in general}, and not specific to that algorithm, we performed additional tests with the MOEA/D algorithm, which is from a different class of MOEA (Section~\ref{sec:nsga_moead}). The results (Figure~\ref{fig:pareto_moead}) demonstrate that also MOEA/D finds good balances between coverage and energy, and outcompetes the traditional methods on complex structures. NSGA-II is however better at distributing plans evenly along the Pareto front for these inspection targets. This has resulted in a significant difference in the hypervolume spanned by the two algorithms on the real-world inspection targets (Table~\ref{table:nsga_moead_comparison}).

\begin{table}
\centering
\begin{tabular}{l c c r}
 \hline
Structure & NSGA-II Hypervolume & MOEA/D Hypervolume & p-value\\
 \hline
 Sphere   & 108.4 [107.3, 109.3] & 108.0 [107.2, 108.4] & 0.11 \\
 SSIV   & 38.2 [38.1, 38.3] & 37.8 [37.7, 37.9] & $< 0.001$ \\
 Manifold   & 126.0 [125.4, 126.7] & 124.8 [124.2, 125.6] & 0.006\\
 \hline
\end{tabular}
\caption{\textbf{The hypervolumes spanned by NSGA-II and MOEA/D.} Median and bootstrapped confidence intervals from 20 runs are shown for each treatment. p-values indicate significant differences between the spanned hypervolumes, calculated by the Mann-Whitney U algorithm.}
\label{table:nsga_moead_comparison}
\end{table}

\begin{figure}
\centering
    \begin{subfigure}[b]{0.55\textwidth}
\includegraphics[width=\textwidth]{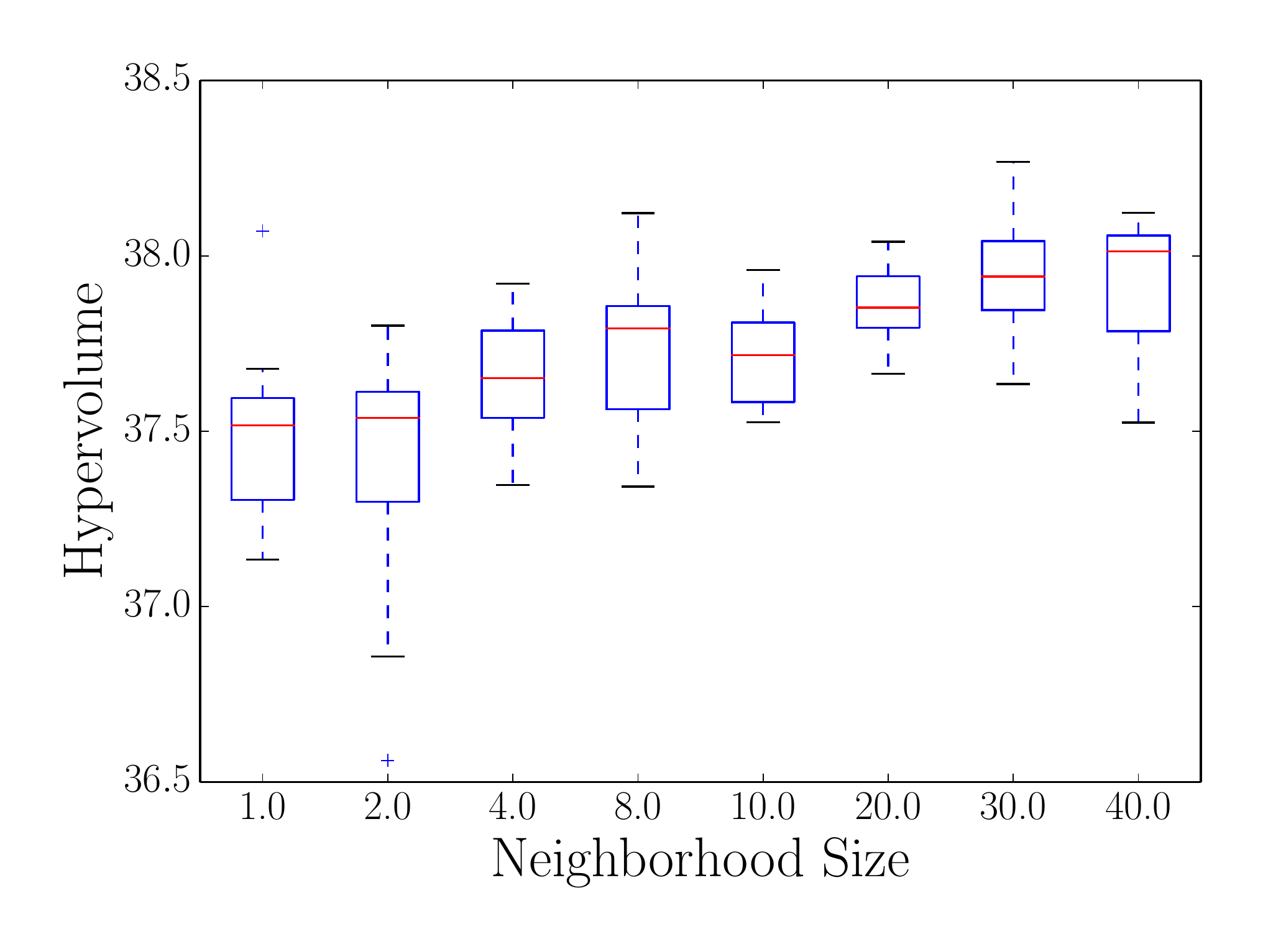}
\caption{The effect of varying the neighborhood size.}
\label{fig:sweep_neighborhood}
\end{subfigure}

    \begin{subfigure}[b]{0.55\textwidth}
\includegraphics[width=\textwidth]{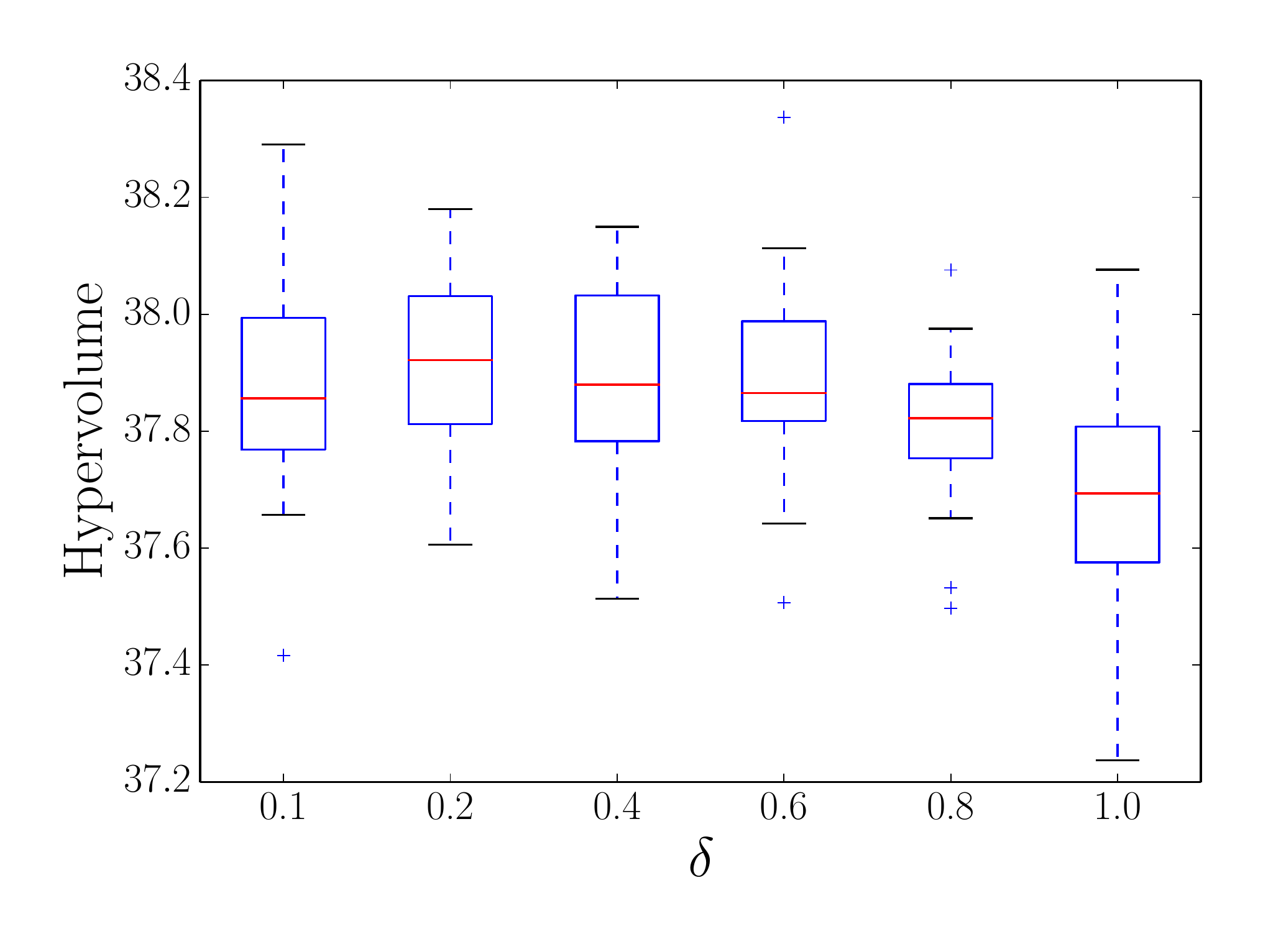}
\caption{The effect of varying $\delta$.}
\label{fig:sweep_delta}
\end{subfigure}
\caption{\textbf{The effect of the MOEA/D specific neighborhood-parameters on performance.} In MOEA/D, mating is restricted to an individual's neighborhood with probability $\delta$ -- otherwise all other individuals are considered for mating. All parameter values were tested on the SSIV inspection target, with 20 independent repetitions.}
\label{fig:sweep}
\end{figure}

As explained in Section~\ref{sec:moead_parameters}, two key parameters affecting the way MOEA/D searches for inspection plans is the \emph{neighborhood size} and $\delta$, the probability of selecting individuals for mating only from an individual's neighborhood. Exploring how different values for these parameters affect the plans generated for the SSIV inspection target demonstrates that for this problem, expanding the neighborhood considered during mating can increase performance (Figure~\ref{fig:sweep}). The paper originally proposing MOEA/D already demonstrated that the algorithm performs worse with very small neighborhood sizes~\cite{Zhang2007}. However, while that paper found a neighborhood size of 10 to be large enough, we here see the largest possible neighborhood sizes (and correspondingly smallest $\delta$ values) result in the best performance. This may indicate that for the inspection planning problem, also very different solutions (in the sense that they are solving very different MOEA/D subproblems) may benefit from sharing information.

\begin{figure}
\centering
   \includegraphics[width=0.55\textwidth]{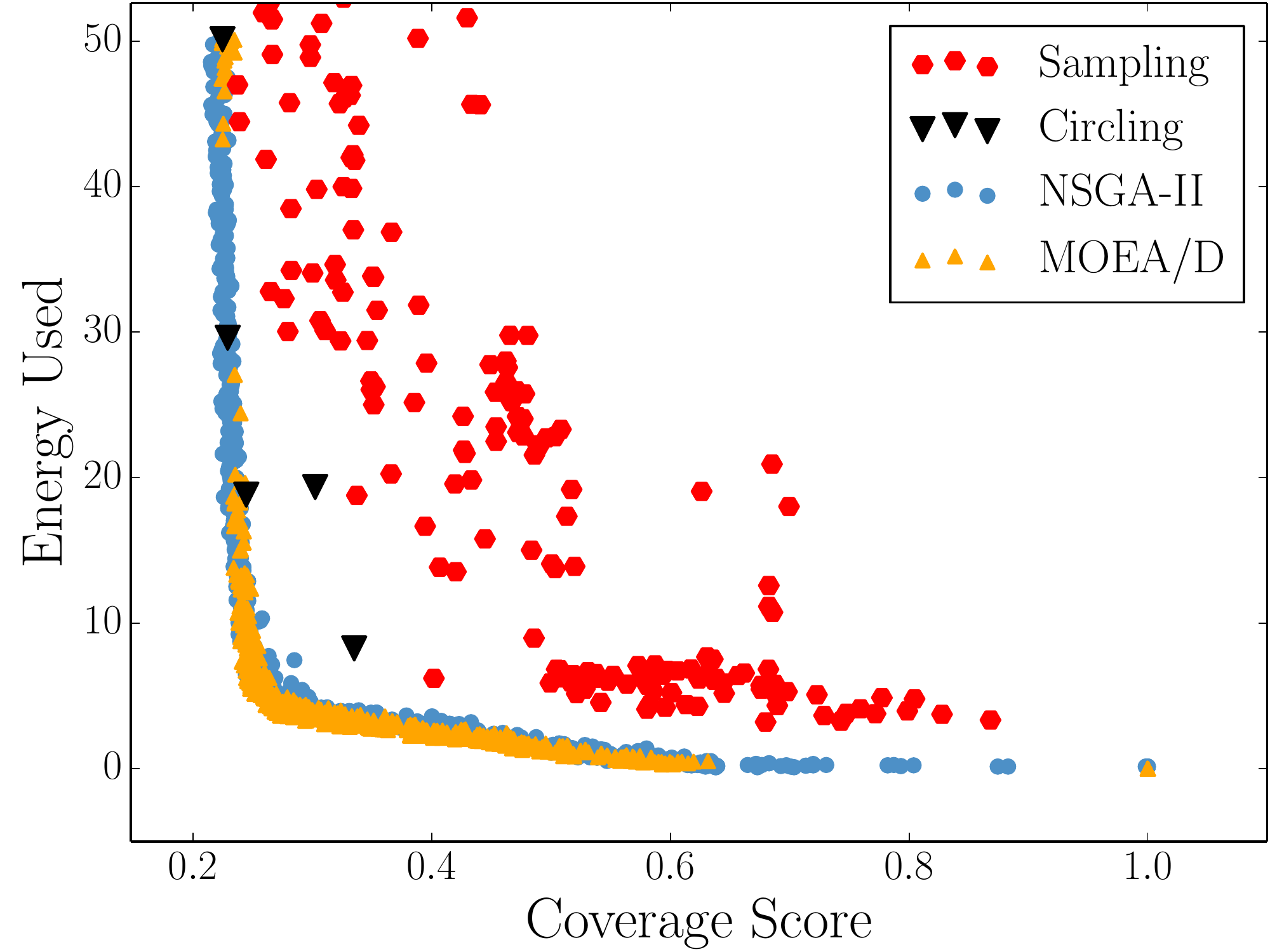}
\caption{\textbf{MOEA/D performance with the maximal neighborhood size of 40.} Results are from the SSIV inspection target, the other algorithm scores are repeated from figure~\ref{fig:scores_pump_moead}.}
\label{fig:neighborhood_40_comparison}
\end{figure}

A closer analysis of the MOEA/D results with the maximal neighborhood size (Figure~\ref{fig:neighborhood_40_comparison}), reveals that the increased neighborhood size has not resulted in a better distribution of solutions along the Pareto front. The increased hypervolume is instead explained by an increased performance of solutions in an intermediate region along the two objectives. This indicates that NSGA-II and MOEA/D have different strengths with regards to optimizing inspection plans: While NSGA-II produces a more diverse selection of plans, MOEA/D generates plans performing very well within a more restricted region of the objective space. Further exploring the strengths and weaknesses of different classes of multiobjective optimization algorithms in generating inspection plans is an important direction for future work.

\subsection{Discussion}

We want to emphasize that this comparison of coverage path planning techniques does not prove that multiobjective inspection planning outperforms the \emph{concepts} of sampling-based or circling-based planning. There exist extensions and specific implementations of both concepts~\cite{Bibuli2007, Englot2012, EnglotHover2012} that may perform better. We could however not compare directly with those, since they were not made available by the authors, and since we needed generalized versions of these algorithms to make a complete comparison. Still, the results presented above give an indication of the strengths and weaknesses of each coverage path planning concept. These are summarized in Table~\ref{table:strengths_weaknesses}, and discussed in more detail below.

\begin{table}
  \centering
  \begin{tabular}{p{1.5cm}p{5.5cm}p{5.5cm}}
    \toprule
    Method & Strengths & Weaknesses \\
    \midrule
    \textbf{Circling}  & \noindent{\begin{itemize}[leftmargin=0.3cm] \itemsep0em  \item Fast and simple method \item Most regular plan structure \end{itemize}} & \noindent{\begin{itemize}[leftmargin=0.3cm] \itemsep0em \item Least flexible \item No optimization \end{itemize}}\\
    
     \textbf{Sampling} & \noindent{\begin{itemize}[leftmargin=0.3cm] \itemsep0em \item Offers coverage guarantees \item Efficient complete coverage planner \end{itemize}} & \noindent{\begin{itemize}[leftmargin=0.3cm] \itemsep0em \item Irregular plan structure \item Does not handle structures with hidden parts well \end{itemize}}\\
     
     \textbf{MOEA} & \noindent{\begin{itemize}[leftmargin=0.3cm] \itemsep0em \item Good performance on structures with hidden parts \item Automatically balances coverage and energy \end{itemize}} & \noindent{\begin{itemize}[leftmargin=0.3cm] \item The most time consuming method \itemsep0em \item No coverage guarantees \end{itemize}}\\
    \bottomrule
\end{tabular}
\caption{\textbf{The strengths and weaknesses of each of the compared coverage path planning methods.} See text for further discussion.}
\label{table:strengths_weaknesses}
\end{table}

The circling-based planner has the advantages of being fast, easy to implement and understand, and producing the most regular plan structures, which may lead to simpler and more reliable mission execution. However, its simplicity also makes it \emph{inflexible}: The method cannot optimize plans to fit a specific inspection target. This may be a problem for complex structures, where circling behaviors do not produce a sufficient coverage, for instance due to occluding elements blocking the robot's view of important elements.

The advantage of the sampling-based method is seen when 100\% coverage is \emph{required}: Of the three methods tested, this one found the most energy efficient way to cover 100\% of the sphere. Further, it is the only one offering a \emph{coverage guarantee}: When run with a required coverage degree of $f$, it does not stop until a fraction $f$ of the structure is covered. Of course, this guarantee only works if a coverage of $f$ is actually achievable for that structure. The disadvantages of the method are that it does not produce efficient plans for the complex structures where complete coverage is not possible, and that the generated plans have a very complex and irregular structure. This can be a challenge for robotic control, mission execution and data interpretation. However, the researchers who developed this procedure have since developed procedures to make sampling-based plans more regular~\cite{EnglotHover2012}, which may mitigate this disadvantage.

The \emph{evolved inspection plans} on the other hand, retain much of the regular circling structure when using circling plans as seeds, while additionally improving the plans to be more energy efficient and more adapted to the inspection target. They also demonstrate good performance structures with hidden or occluded parts, and automatically balance coverage and energy -- since the two objectives are optimized \emph{together}. The main disadvantage of evolving inspection plans is the longer runtime. Evolving inspection plans clearly needs to happen before deployment -- and a different process is required to adapt the plans online, if unexpected situations occur.

\section{Conclusion}

Through a comparison with two state-of-the-art coverage path planning techniques, we have validated the ability of our recently suggested multiobjective coverage path planning method to generate good inspection path plans. While our method is also found to generate good complete coverage path plans for a simple structure, its advantage is seen most clearly when generating plans for \emph{complex real-world structures}. For such structures, a completely covering inspection path is often impossible, and we demonstrated how traditional \emph{complete coverage} path planning methods struggle to generate good plans for these -- also when we relax the complete coverage constraint on these planners.

Multiobjective coverage path planning is inherently \emph{different} from previous methods, allowing plans to gracefully adapt to the amount of occlusion and complexity in the inspection target. Optimizing energy usage and coverage together ensures a good balance between the two both when 100\% coverage is available, and when large parts of the object are hidden. The performance of the algorithm on the three very different inspection targets increases our confidence that multiobjective optimization is a general and flexible approach -- enabling inspection path planning on many structures that traditional methods do not support, since they constrain search to \emph{complete coverage} plans. 

Our technique offers a starting point for automatic inspection path planning on complex real-world structures, and there is much interesting work to be done in developing this approach for the autonomous inspection missions of the future. While the applied methods prove that multiobjective optimization can generate good inspection plans for complex structures, further studies are required to systematically assess the best way to apply such techniques. These studies include testing different multiobjective optimization algorithms, as well as studying the impact of key parameters. Such systematic studies have recently illuminated the performance of evolutionary algorithms in underwater path planning~\cite{Zamuda2014,Zamuda2016}, and we consider them an important step in further studies of \emph{inspection path planning}.

The method is so far only used offline, and future work will also need to study how to best update evolved plans on-line. The evolved population of inspection paths may provide a good starting point for online adjustments, as it maintains a diverse set of solutions with different ways to handle the problem. Another interesting direction for future work is introducing more knowledge in the evolutionary algorithm (for instance by \emph{hybridizing} it with a local search). This paper demonstrated the benefits of inserting knowledge in the form of initial seed solutions -- suggesting that an even more knowledge-intensive search may give additional performance benefits.


\section{Acknowledgments}

Work by Kai Olav Ellefsen was funded by Brazilian National Research Council (CNPq) through a “Young Talent Attraction” scholarship [grant number 314886/2014-1], and also partially supported by The Research Council of Norway as a part of the Engineering Predictability with Embodied Cognition (EPEC) project, under grant agreement 240862. The ongoing development of the FlatFish AUV is financed by Shell, ANP and Embrapii. We also thank the members of the Brazilian Institute of Robotics for their support, Brendan Englot for enlightening email discussions and Joost Huizinga for helpful feedback.

\section{Vitae}

\includegraphics[width=0.3\columnwidth]{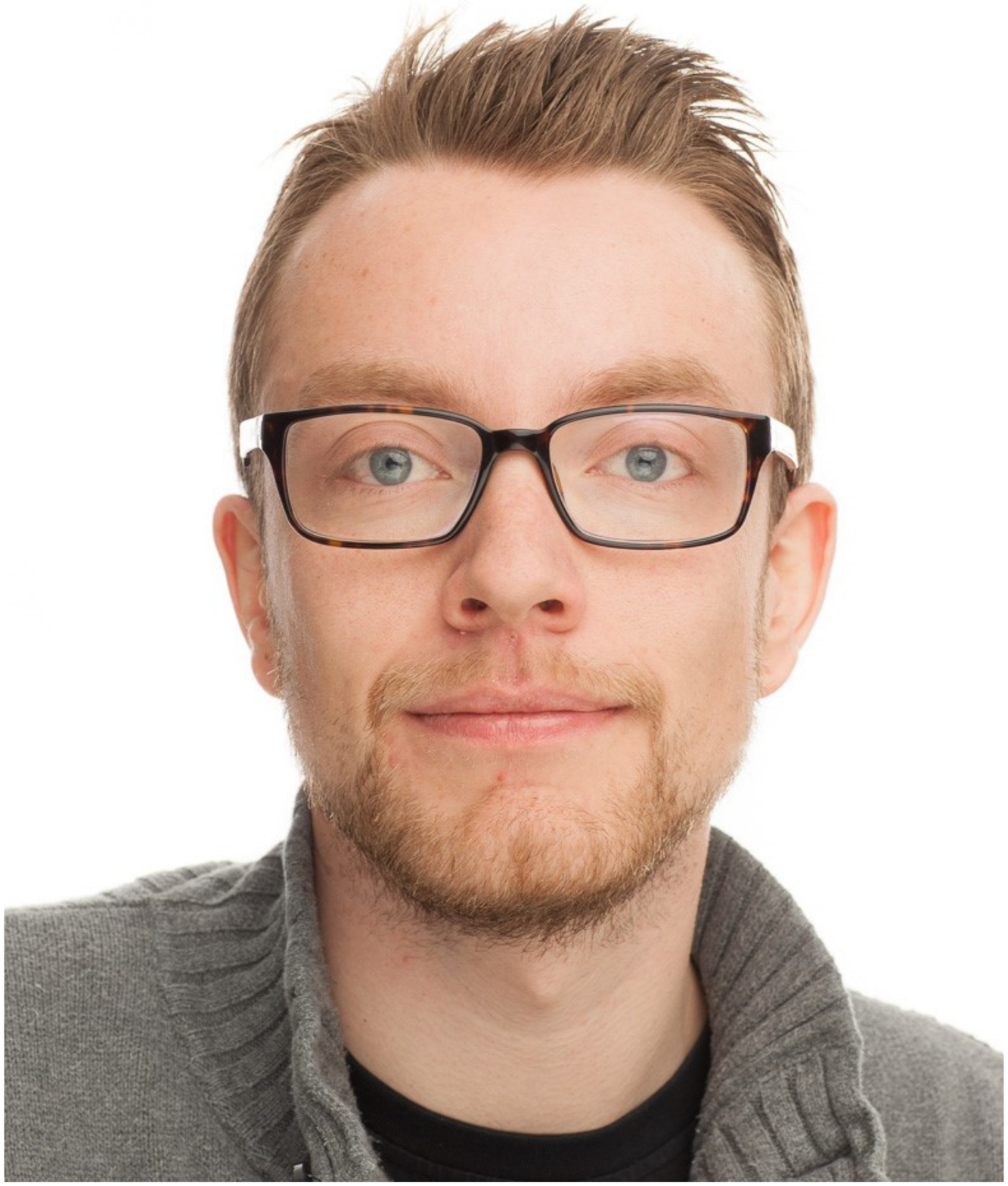}

Kai Olav Ellefsen is a postdoctoral researcher at the Brazilian Institute of Robotics at SENAI CIMATEC (Salvador, Brazil) where he has been part of the FlatFish AUV project since 2014. He holds a Masters degree (2010, winner of the Norwegian Artificial Intelligence Society Best Master Thesis Award) and Ph. D. (2014) from the Norwegian University of Science and Technology. His research interests cover many topics in Artificial Intelligence, including evolutionary algorithms, artificial neural networks, adaptation and learning.

\includegraphics[width=0.3\columnwidth]{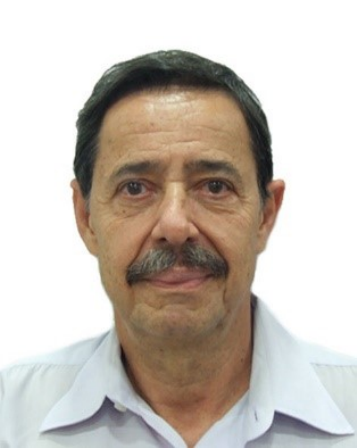}

Herman Lepikson is director of the SENAI Institute for Innovation in Automation and Associate Professor at UFBA (Federal University, Bahia State). He holds a Doctor degree in Engineering from the Federal University of Santa Catarina in Manufacturing Systems; MBA in Engineering Economics from PUC-MG, and Mechanical Engineering from UFBA. He is a DT2 researcher of CNPq (Brazil's National Research Council). His research work is in the areas of manufacturing integration, advanced manufacturing and mechatronic systems design. He is a permanent Professor of the Graduate Programs in Mechatronics and Industrial Engineering at UFBA, in which he advises master and doctorate students.

\includegraphics[width=0.3\columnwidth]{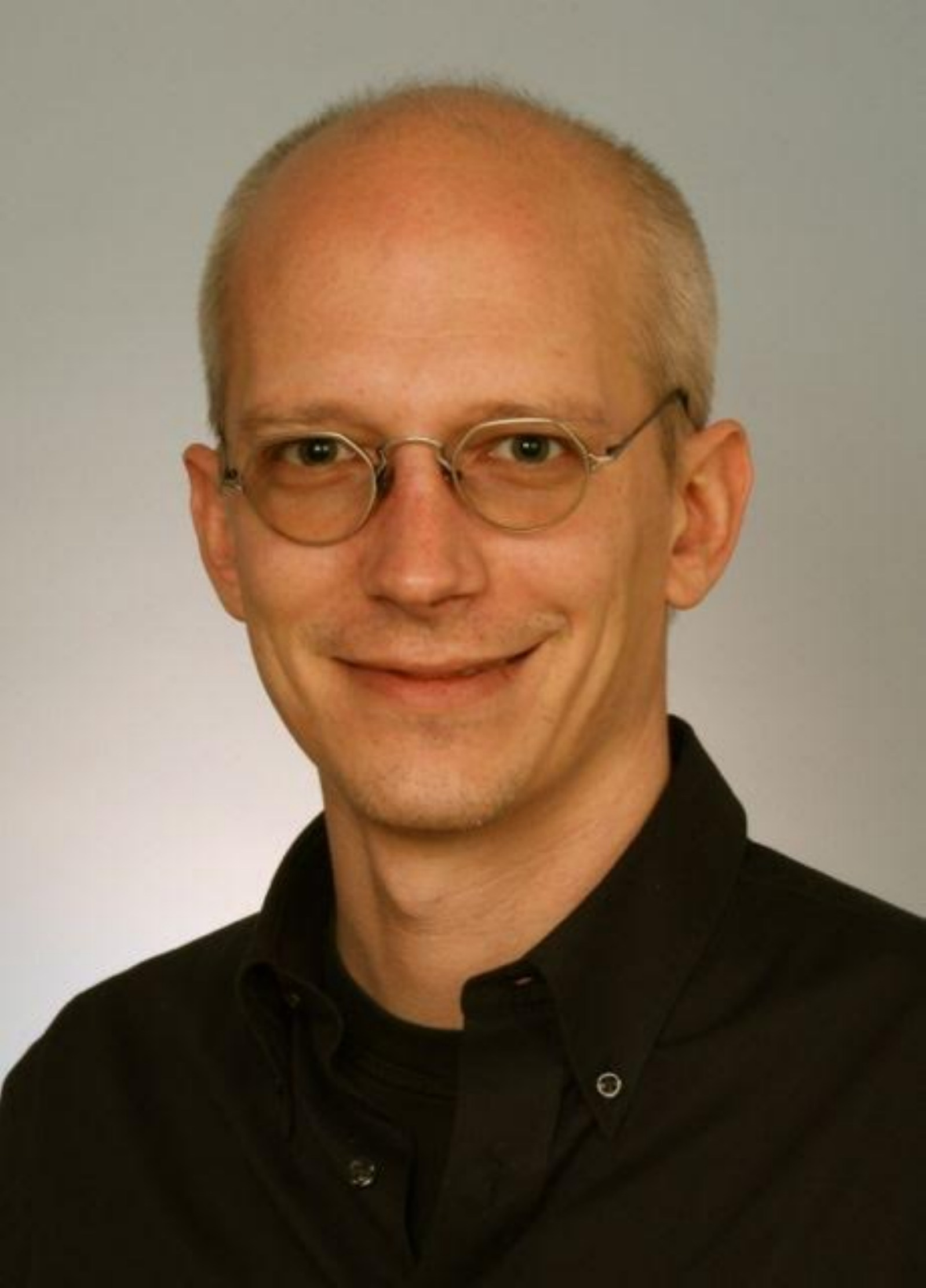}

Jan Albiez received his PhD 2007 in the area of biological inspired walking machines, after working for 6 years as researcher in Robotics at FZI in Karlsruhe, Germany. In 2006 he started at the newly founded robotic branch of DFKI in Bremen, Germany, where he was responsible for developing the area of underwater robotics, acted as project leader for major research projects and set up all of RIC's testing environments. In 2014 he went to SENAI CIMATEC in Salvador, Brasil where he started as technical project leader of the FlatFish AUV project and trained a 20 person team in robotics.

\section{Bibliography}

\bibliographystyle{plain}
\bibliography{library}

\end{document}